\documentclass[a4paper,fleqn]{cas-sc}

\usepackage{caption}
\usepackage{subcaption}
\usepackage[numbers]{natbib}
\usepackage{algorithm}
\usepackage{mathtools}
\usepackage[noend]{algpseudocode}
\usepackage{multirow}

\usepackage{wasysym}
\usepackage{ulem}
\usepackage{siunitx}
\usepackage{booktabs}
\usepackage{comment}
\usepackage{lineno}

\newcommand{\argmax}{\mathop{\rm arg~max}\limits}

\usepackage[labelformat=simple]{subcaption}

\def\tsc#1{\csdef{#1}{\textsc{\lowercase{#1}}\xspace}}
\tsc{WGM}
\tsc{QE}

\begin{document}
\let\WriteBookmarks\relax
\def\floatpagepagefraction{1}
\def\textpagefraction{.001}

\shorttitle{Discrete-Choice Model with GAIUNet}    

\shortauthors{T. Nishi and Y. Hara}  

\title [mode = title]{Discrete-Choice Model with Generalized Additive Utility Network}  



%

\author[1]{Tomoki Nishi}[
        orcid=0000-0001-8418-4633
]

\cormark[1]


\ead{nishi@mosk.tytlabs.co.jp}


\credit{Conceptualization of this study, Methodology, Software}

\affiliation[1]{organization={Toyota Central R\&D Labs., Inc.},
            addressline={41-1, Yokomichi}, 
            city={Nagakute},
            postcode={480-1192}, 
            state={Aichi},
            country={Japan}}

\author[2,3]{Yusuke Hara}[
        orcid=0000-0003-0268-9908
]


\ead{hara@tohoku.ac.jp}


\credit{Conceptualization of this study, Methodology}

\affiliation[2]{organization={Tohoku University},
            addressline={6-6-10 Aramaki Aza Aoba, Aoba-ku}, 
            city={Sendai},
            postcode={980-8579}, 
            state={Miyagi},
            country={Japan}}

\affiliation[3]{organization={University of Tokyo},
            addressline={7-3-1 Hongo}, 
            city={Bunkyo-ku},
            postcode={113-0033}, 
            state={Tokyo},
            country={Japan}}

\cortext[1]{Corresponding author}



\begin{abstract}
Discrete-choice models are powerful frameworks for analyzing decision-making behavior to provide valuable insights for policymakers and businesses. Multinomial logit models~(MNLs) with linear utility functions~(MNL-Linear) have been used in practice because they are easy to use and interpretable. Recently, MNLs with neural networks (e.g., ASU-DNN) have been developed, and they have achieved higher prediction accuracy in behavior choice than classical MNLs such as MNL-Linear. However, these models often make it difficult to interpret the impact of each factor individually because of their complex structures. We developed utility functions with a novel neural-network architecture based on generalized additive models, called the generalized additive utility network ~(GAUNet), for discrete-choice models. We evaluated the performance of the MNL with GAUNet using synthetic data for transport mode choice and trip survey data collected in Tokyo. The results indicated that our model improved the policy evaluation and interpretability compared with previous models while maintaining prediction accuracy on par with ASU-DNN.
\end{abstract}


\begin{highlights}
    \item Neural network-based generalized additive utility function, GAIUNet, is developed.
    \item Multinomial logit model via GAIUNet provides intuitive interpretation for user preference.
    \item Our model excelled in policy evaluation via synthetic data with hidden constraints.
    \item We assessed prediction accuracy and interpretation via actual probe-person data.
\end{highlights}

\begin{keywords}
 Discrete-Choice Model\sep Multinomial Logit Model\sep Neural Network\sep Generalized Additive Model\sep Interpretability
\end{keywords}

\maketitle

\section{Introduction}\label{sec:intro}

Discrete-choice models (DCMs) are powerful frameworks for analyzing and predicting decision-making behavior in various contexts to provide valuable insights for policymakers and businesses~\cite{train2009discrete}. DCMs have supported policymakers and businesses for many years in practice while having been developed as theory-driven models on the basis of random utility theory after the random utility maximization model~\cite{mcfadden1973conditional} was developed. These models are commonly used in marketing, transportation, urban planning, and other fields. DCMs can be used to analyze the effects of attributes on choice behavior, predict choices, and evaluate the impact on different policy interventions or changes in the features of alternatives. The model is typically formulated with utility and choice probability functions. 

The typical objective of the analysis using DCMs is policy evaluation, which measures the impact of novel interventions. A naive straightforward approach to realizing the objective is to measure the impact of the intervention by implementing the policy. However, in some domains, it is often difficult to measure such impacts directly for various reasons such as costs. Therefore, approaches to measure the impact indirectly are desired. DCMs have also been used for indirect evaluation by shifting the values of explanatory variables while fixing the model parameters learned from prior collected survey data. 

The prediction accuracy of the models measured from collected data does not always reflect the accuracy of  the policy evaluation because the model needs to be extrapolated rather than interpolated for policy evaluation when some areas have nonoverlapping areas between the distribution of data obtained from the intervention and the distribution of collected data. These challenges have been mitigated by expert interpretation of the model. That is, interpretability is also essential for the models to be used for policy evaluation. The classical DCMs, such as multinomial logit models with linear utility, have been used for the analysis because of interpretability with respect to the contribution of explanatory variables. Conversely, applying a suitable utility function (e.g., linear or logarithmic) with domain-specific expert knowledge is required to accurately predict choice behavior, but this is often difficult, for example, when the utility function becomes nonlinear because of unknown and unobservable constraints (e.g., budget). 

With the rise of machine learning and AI fields, DCMs using machine learning have been developed since 2017~\cite{van2022choice}.  Many early studies mainly aimed at proposing methods with higher prediction accuracy than standard multinomial logit (MNL) models. For example, Wang et al. \cite{wang2020deep} proposed ASU-DNN, which learns the utility function for each alternative using a neural network (NN). The model achieved more accurate predictions than previous classical methods because of the nonlinear utility function with higher-order interactions between explanatory variables for each alternative. In ASU-DNN, sensitivity analysis, which quantifies the effect of a variable by changing the value of the variable and fixing the one of the other variables,  is required to analyze the effect of each variable owing to its network structure. That is, the quantified effect of a variable depends on the fixed values of the other variables. Some hybrid approaches have recently been developed, that combine classical DCMs and machine learning models to improve predictability and interpretability. Sifringer et al.~\cite{sifringer2020enhancing} proposed a linear MNL (L-MNL), which combines linear utility with MNL for the main factors and an NN to handle nonlinear utility for other additional factors. Han et al.~\cite{han2020neural} proposed TasteNet-MNL, which also learns the utility functions for the main factors for each alternative with the MNL-Linear and for individual-specific taste parameters with an NN. They reported that the model improved the accuracy of behavior prediction. The approach maintains interpretability for key factors by assuming linear utility functions and achieves higher accuracy via the NNs. However, the approaches still have concerns about the interpretability on the part of the NNs, similar to other studies. Although predictability has been evaluated  through various experiments in previous studies for neural network-based approaches, the performance of policy evaluation has not yet been evaluated. 

In this study, we developed utility functions using a novel NN architecture based on generalized additive models called the generalized additive utility network~(GAUNet). GAUNet represents utility functions via generalized additive models~\cite{hastie1990generalized}, whose shape function of each variable uses the separated NNs with the NNs of the other variables for each alternative, whereas ASU-DNN represents utility functions via the fully connected DNN for each alternative. Our approach maintains interpretability for the effect of each explanatory variable by learning the effect of each variable as a different function, and achieves high accuracy when the NNs are used as the utility function. The utility function with GAUNet can be applied to other DCMs on the basis of random utility theory because our approach is a natural generalization of classical MNL. We also extend GAUNet to a generalized additive and interactive utility network~(GAIUNet), which adds the interaction terms between two explanatory variables represented by fully connected NNs to GAUNet to capture the effect of the interaction between two variables. 

We demonstrate the multinomial logit model via GAUNet~(MNL-GAUNet) and GAIUNet~(MNL-GAIUNet). We evaluate our model with two types of data for transport mode choice: synthetic data and a probe-person data. The synthetic data were generated on the basis of the transport mode choice of users, who have unobservable constraints on travel costs and access time. In the synthetic data, we evaluate how accurately our models can measure the impact of policy intervention as the policy evaluation after the prediction accuracy is measured because it is difficult to accurately compare the performance of the policy evaluation across the models in the real world. The other dataset includes data collected in Tokyo, Japan, from 2018 through 2021. The models learn to make decisions about the choice of transportation mode (e.g., train, bus, or car). We confirm that our models have interpretability while maintaining prediction accuracy comparable to that of ASU-DNN.

The contributions of this study are as follows:
\begin{enumerate}[a)]
    \item We propose utility functions for DCMs via a novel NN architecture that is based on the generalized additive models GAUNet and GAIUNet, which add 2D interaction terms in addition to GAUNet (Section~\ref{subsec:GAIUNet}).
    \item We developed MNL-GAUNet and MNL-GAIUNet, which constitute MNL employing GAUNet and GAIUNet (Section~\ref{subsec:MNL-GAUNet}, \ref{subsec:MNL-GAIUNet}). 
    \item We compare our method to ASU-DNN and MNL with linear utility functions on the synthetic data concerning transport mode choice with unobservable user constraints on travel costs and access time (Section~\ref{sec:experiment-synthetic-data}). We evaluate the prediction accuracy and performance of the policy evaluation across the models. We reveal that our method is promising for policy evaluation both in cases where utility is linearly affected by variables and when sudden, significant changes in utilities occur due to unobservable constraints. We also show that the learned utilities provide intuitive interpretations of user preferences, such as where the constraints lie.
    \item We compare our models to probe-person data collected in Tokyo, Japan (Section~\ref{sec:experiment-toyosu}). We confirm that the properties of our models are consistent with the real-world datasets in terms of prediction accuracy and interpretability.  
\end{enumerate}

\section{Literature Review} \label{sec:related-work}

DCMs have been actively studied for many years since McFadden proposed random utility maximization\cite{mcfadden1973conditional}. DCMs, grounded in statistical theory, assume that decision-makers make the choices sampled from the probability distribution calculated by their utility, which is the degree of preference for each alternative. The approaches focus on specifying the mechanism to understand how decision-makers make choices rather than predicting the choices\cite{train2009discrete}.  Many variations of DCMs have been developed: MNL~\cite{mcfadden1973conditional}, probit~\cite{thurstone1927}, mixed logit (MXL)~\cite{mcfadden2000mixed}, cross-nested logit (CNL)~\cite{vovsha1997application}, and latent-class forms (LCM)~\cite{greene2003latent}. The models differ in their assumptions of decision-making processes or probability distributions. After selecting a DCM (e.g., MNL), the analyst must prepare explanatory variables (e.g., travel time). Moreover, the variables must be handcrafted via variable transform (e.g., $\exp(x)$ and $\ln(x)$ ), where nonlinear characteristics are necessary when domain-specific expert knowledge is used. In conventional DCMs such as MNL, the utility is often estimated via a weighted linear combination of the handcrafted explanatory variables. 

The transformation of variables is not always easy for even experts because the nonlinearity of utility is often derived from the unobservable constraints of users. Constrained MNL~(CMNL) has been developed to handle such constraints~\cite{martinez2009constrained, castro2013estimation}. They explicitly formulate the constraints within the MNL framework. Previous studies have shown that CMNL can also estimate thresholds of variables in addition to learning parameters of explanatory variables that are the same as those of standard MNL. However, in CMNL, we face a priori question of which variables are subject to thresholds. Although our method does not explicitly represent constraints like thresholds, it can implicitly handle user constraints by learning nonlinear features of variables.
 
MNL, whose utility functions are automatically designed by NNs, has recently been developed: ASU-DNN\cite{wang2020deep}, L-MNL\cite{sifringer2020enhancing}, and TasteNet-MNL\cite{han2020neural}.  ASU-DNN separately estimates the utility of each alternative via a fully connected NN. These neural network-based approaches could mitigate the issues (e.g., variable transform and unobservable constraints) of the classical MNL methods because of the high representation of the NNs.  Previous studies have reported that these methods achieve higher prediction accuracy than conventional DCMs and ML methods in practice. Moreover, the method must be applied post-processing, such as sensitivity analysis, to interpret the effect of each explanatory variable. L-MNL learns the utility of the main factors via MNL-Linear and one of the other factors via a fully connected NN. The model learned via L-MNL provides the understanding of the main factors due to the MNL-Linear but makes us struggle to understand the other factors without post-processing because of the NN. TasteNet-MNL learns utility through MNL-Linear via hand-crafted explanatory variables for the main factor and NN-learned variables for the decision-maker attributes. Our MNL-GAUNet learns the utility for each explanatory variable in each alternative separately via an NN. We can easily interpret the preference for each variable because the utility of each variable can be computed independently of other variables. MNL-GAIUNet also learns the utility for 2D interactions between the explanatory variables via NNs in addition to MNL-GAIUNet. Moreover, although predictability has been evaluated  through various experiments in previous studies for neural network-based approaches, the performance of policy evaluation has not yet been evaluated. Therefore, in Section~\ref{sec:experiment-synthetic-data}, we also compare these methods in terms of policy evaluation via the synthetic data.

We can also analyze MNL-GAIUNet in the context of generalized additive models (GAMs)\cite{hastie1990generalized}. GAMs are a flexible and powerful class of statistical models that extend the concept of generalized linear models (GLMs)\cite{nelder1972generalized}. Whereas GLMs assume that the relationship between explanatory variables and the outcomes is linear, GAMs, which are modeled by the weighted sum of nonlinear shape functions, allow for nonlinear relationships. Lou et al. proposed $\rm{GA^2M}$, which introduces an interaction term into GAM.  $\rm{GA^2Ms}$ using modern shape functions have been developed. The explainable boosting machines (EBMs)~\cite{nori2019interpretml} and GAMI-Net~\cite{yang2021gami} are $\rm{GA^2M}$ with gradient boosting trees and NNs as shape functions, respectively. Our MNL-GAUNet (or MNL-GAIUNet) can be regarded as the MNL using the GAM (or $\rm{GA^2M}$), whose shape functions are NNs, as the utility function for each alternative. 

\section{Preliminaries} \label{sec:preliminaries}

This section introduces MNLs and GAMs. 

\subsection{Multinomial Logit Model}

DCMs are statistical models that analyze decision-making behavior in situations where people choose an alternative from the set of available alternatives $\mathcal{A}$. DCMs assume that individuals make choices based on their preferences, called utility, for the alternatives themselves and their characteristics. Using statistical models, DCMs aim to represent the relationships between characteristics and individual choices. Random utility maximization (RUM) is the fundamental concept in DCMs, which assumes that individuals make choices by maximizing their utility.  

Let $\mathcal{A} = \{ 1,\dots,K\}$ be set of the alternative index. DCMs treat two types of inputs: alternative-specific variables (e.g., travel time) $\mathbf{x}_{i\in \mathcal{A}}\coloneqq [x_{i1},\dots,x_{iN}]^\top$ and individual-specific taste parameters (e.g., income) $\mathbf{z}\coloneqq [z_{1},\dots,z_{M}]^\top$, where $N$ and $M$ are the number of variables.  In DCMs, the utility of each alternative $U_k$ is formulated as the sum of the deterministic utility $V_k$ and random utility $\varepsilon_k$ as follows:
\begin{linenomath}
\begin{align}
    U_k(\mathbf{x}_k,\mathbf{z}) &\coloneqq V_k(\mathbf{x}_k,\mathbf{z}) + \varepsilon_k,\quad \forall k \in \mathcal{A}.
\end{align}  
\end{linenomath}
For simplicity, we omit the taste parameters $\mathbf{z}$ from our formulation hereafter because we can easily extend the following formulation to the one with the taste parameters.  According to RUM, individuals select the alternative with the maximum utility: $i^* = \argmax_{i \in \mathcal{A}}{U_i}$. The probability that the individual selects alternative $i$ is 
\begin{linenomath}
\begin{align}
P_i &= \mathrm{Prob}(U_i > U_j), \quad i,\forall j\neq i \in \mathcal{A} \nonumber \\
    &= \mathrm{Prob}(V_i + \varepsilon_i > V_j + \varepsilon_j) \nonumber \\
    &= \mathrm{Prob}(V_j - V_i > \varepsilon_j - \varepsilon_i)
\end{align}
\end{linenomath}

The choice model, in which $\varepsilon_{i\in \mathcal{A}}$ follows the Gumbel distribution, is known as the MNL. The choice probability maximizing the utility can be represented as the following softmax function:
\begin{linenomath}
\begin{align}
    \varepsilon_i &\sim \mathrm{Gumbel}(\varepsilon_i;0,\mu),\quad i \in \mathcal{A}, \\
                  &\coloneqq \frac{1}{\mu}\exp\left(-\frac{\varepsilon_i}{\mu}\right)\exp\left[\exp\left(-\frac{\varepsilon_i}{\mu}\right)\right],\\
    P(i|\mathbf{x}) &= \frac{\exp(V_i(\mathbf{x}_i))}{\sum_{j\in \mathcal{A}}\exp(V_j(\mathbf{x}_j))},\quad i \in \mathcal{A},\label{eq:pix}
\end{align}
\end{linenomath}
where $\mu$ is a parameter of the Gumbel distribution and $\mathbf{x}$ is the set of the variables (i.e., $\mathbf{x}\coloneqq \{\mathbf{x}_1,\dots, \mathbf{x}_K\}$). The proof is available in the book\cite{train2009discrete}.

The deterministic utility function can be approximated via any function. For example, MNL with a linear utility function (MNL-Linear) applies a weighted sum of explanatory variables (hereafter used as a generic term for alternative- and individual-specific variables). The choice probability of MNL-Linear can be estimated as follows:
\begin{linenomath}
\begin{align}
    V_i(\mathbf{x}_i;\boldsymbol{\omega}_i) &\coloneqq \boldsymbol{\omega}_{i}^\top \mathbf{x}_{i},\quad i \in \mathcal{A},\label{eq:v_linear}\\
    P(i|\mathbf{x};\boldsymbol{\omega}) &= \frac{\exp(V_i(\mathbf{x}_i;\boldsymbol{\omega}_i))}{\sum_{i\in \mathcal{A}}\exp(V_i(\mathbf{x}_i;\boldsymbol{\omega}_i))},\quad i \in \mathcal{A},\\
    \mathbf{x} &\coloneqq \{\mathbf{x}_i|i\in \mathcal{A}\},\\
    \boldsymbol{\omega}&\coloneqq\{\boldsymbol{\omega}_i|i\in \mathcal{A}\},
\end{align}
\end{linenomath}
where $\boldsymbol{\omega}_i$ is the weight vector of the explanatory variables for alternative $i$. Fukuda and Yai approximated utility using smoothing cubic spline\cite{fukuda2010semiparametric}. ASU-DNN\cite{wang2020deep} adopts NNs as deterministic utilities. Both L-MNL\cite{sifringer2020enhancing} and TasteNet-MNL\cite{han2020neural} use the linear function mainly and NNs as supplements.

The weights can be estimated by maximizing the log-likelihood for data $\mathcal{D} \coloneqq \{(y_1,\mathbf{x}_1),\dots,(y_D,\mathbf{x}_D)\}$, where $(y_d,\mathbf{x}_d)$ is the $d$-th data and $y_d\in \mathcal{A}$ is the chosen alternative index in the $d$-th data:
\begin{linenomath}
\begin{align}
    \boldsymbol{\tilde{\omega}} = \argmax_{\boldsymbol{\omega}} \sum_{(y_d,\mathbf{x}_d) \in \mathcal{D}} \ln P(y_d|\mathbf{x}_d;\boldsymbol{\omega}),
\end{align}
\end{linenomath}
where $\boldsymbol{\tilde{\omega}}$ are the estimated weights. Various gradient methods can be used as optimization methods: BFGS~\cite{liu1989limited} in MNL-Linear and stochastic gradient descent approaches (e.g., Adam~\cite{kingma2014adam}) in MNL with NNs.

In conventional MNL analyses that use MNL-Linear, weights for certain variables, such as travel time and distance, are frequently pooled across all alternatives to comprehend their universal characteristics. This shared weighting allows for the estimation of travel time utility that is agnostic to specific transportation modes. Conversely, weights that are unique to each mode facilitate the estimation of mode-specific utilities. The selection between these approaches is contingent upon the objectives of the analysis.  When we apply some MNLs with NN approaches~(e.g., ASU-DNN), we cannot analyze such universal characteristics because the utility for each variable cannot be calculated independently of the other variables owing to the network architectures.

\subsection{Generalized Additive Models}

Linear regression models (LRMs) have been widely used to model the linear relationship between an outcome $y$ and the explanatory variables $\mathbf{x}=\{x_1,\dots,x_N\}$:
\begin{linenomath}
\begin{align}
y = \beta_0 + \sum_j^N \beta_j x_j + \varepsilon,
\end{align}
\end{linenomath}
where N is the number of explanatory variables. $\beta_0$ and $\{\beta_1,\dots,\beta_N\}$ are the coefficients associated with the intercept and slope, respectively; $\varepsilon$ represents the error term.  The models implicitly assume that the error term is drawn from the normal distribution. LRMs cannot work well when assumptions of the linear relationship and the error term distribution are violated.

GLMs are statistical models that relax the assumption of the error term distribution via a link function $g$: 
\begin{linenomath}
\begin{align}
\mathbb{E}(y) = g^{-1}\left(\beta_0 + \sum_j^N \beta_j x_j\right),
\end{align}
\end{linenomath}
where $\mathbb{E}(y)$ denotes the expected value of $y$. By changing the link functions, GLMs can represent various regression models, such as Poisson regression and logistic regression. MNL-Linear can also be derived from GLMs by applying baseline category logit link functions $\ln(y_{i\in\mathcal{A}}/y_0)$, where $y_0$ is the outcome of the baseline category \cite{agresti2015foundations}. Estimating the parameters of GLMs involves maximizing the likelihood function represented by the probability of observing the training data.

GAMs are the statistical models that extend GLMs. Whereas GLMs assume the relationships between the explanatory variables and the outcome projected by the link function, GAMs allow for nonlinear relationships as follows:
\begin{linenomath}
\begin{align}
\mathbb{E}(y) = g^{-1}\left(\beta_0 + \sum_j^N \beta_j h_j(x_j)\right),
\end{align}
\end{linenomath}
where $h_j(x_j)$ is a shape function for the $j$-th explanatory variable. The smooth functions are typically represented using spline functions. Lou et al. developed an extended GAM, called $\textrm{GA}^2\textrm{M}$, which is the model that incorporates the interaction effects between two explanatory variables into GAMs \cite{lou2013accurate}:
\begin{linenomath}
\begin{align}
\mathbb{E}(y) = g^{-1}\left(\beta_0 + \sum_{j}^N \beta_j h_j(x_j) + \sum_{j,k\neq j}^N \beta_{jk} h_{jk}(x_j,x_k)\right).
\end{align}
\end{linenomath}
The explainable boosting machine EBM)\cite{nori2019interpretml}, which is $\textrm{GA}^2\textrm{Ms}$ with gradient boosting trees as the shape functions, was developed. GAMI-Net uses NNs as shape functions \cite{yang2021gami}. EBM and GAMI-Net have been reported to have both high precision accuracy and interpretability.

\subsection{Artificial Neural Network}

Artificial neural networks~(ANNs) are supervised-learning frameworks that represent the relationship between the explanatory variables $\mathbf{x}$ and the outcomes $\mathbf{y}$ by repeatedly mapping via nonlinear functions\cite{lecun2015deep}:
\begin{linenomath}
\begin{align}
\mathbf{y}^{(l)} = h^{(l-1)}\left(\beta_0^{(l-1)}+\boldsymbol{\beta}^{(l-1)\top}\mathbf{y}^{(l-1)}\right),\quad \forall l = 1,\dots,L
\end{align}
\end{linenomath}
where L is the number of layers, $\mathbf{y}^{(0)}=\mathbf{x}$, $y^{(L)}=\mathbf{y}$, and $l$  is the layer index in the NN. Let $\boldsymbol{\beta}^{(l-1)}$ and $\beta_0^{(l-1)}$ be the weight vector and intercept in $(l-1)$-th layer, respectively. $h^{(l-1)}$ is the activation function. We often use several nonlinear functions as the activation functions: hyperbolic tangent $\mathrm{Tanh}(\cdot)$, rectified linear function $\mathrm{ReLU}(\cdot)=\max(0,\cdot)$\cite{nair2010rectified}, and softmax function $\mathrm{Softmax}(\cdot)$. MNL-Linear can be derived by using a single layer ($L=1$) and the softmax function as the activation function $h^{0}$, where $\mathbf{y}$ is the choice probability vector of each alternative and $\mathbf{x}$ is the explanatory variable. When we use categorical cross-entropy\cite{richard1991neural} as the loss function of the NN, the model is equivalent to MNL-Linear maximizing log-likelihood. 

\begin{figure}[ht]
    \centering
    \begin{minipage}[b]{0.28\linewidth}
    \centering
    \includegraphics[width=1.0\hsize]{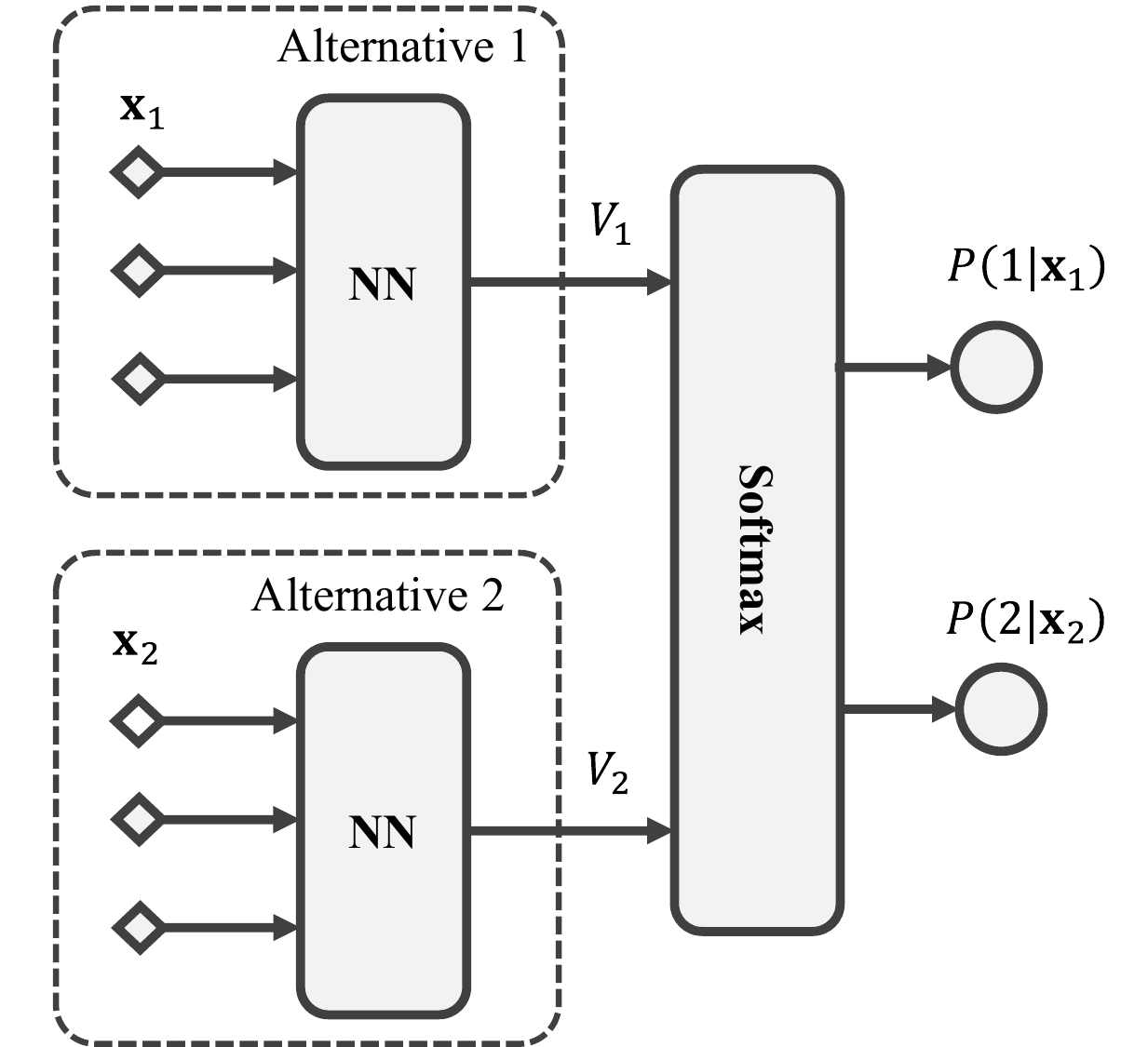}
    \subcaption{ASU-DNN}\label{fig:asu-dnn}
    \end{minipage}
    \begin{minipage}[b]{0.28\linewidth}
    \centering
    \includegraphics[width=1.0\hsize]{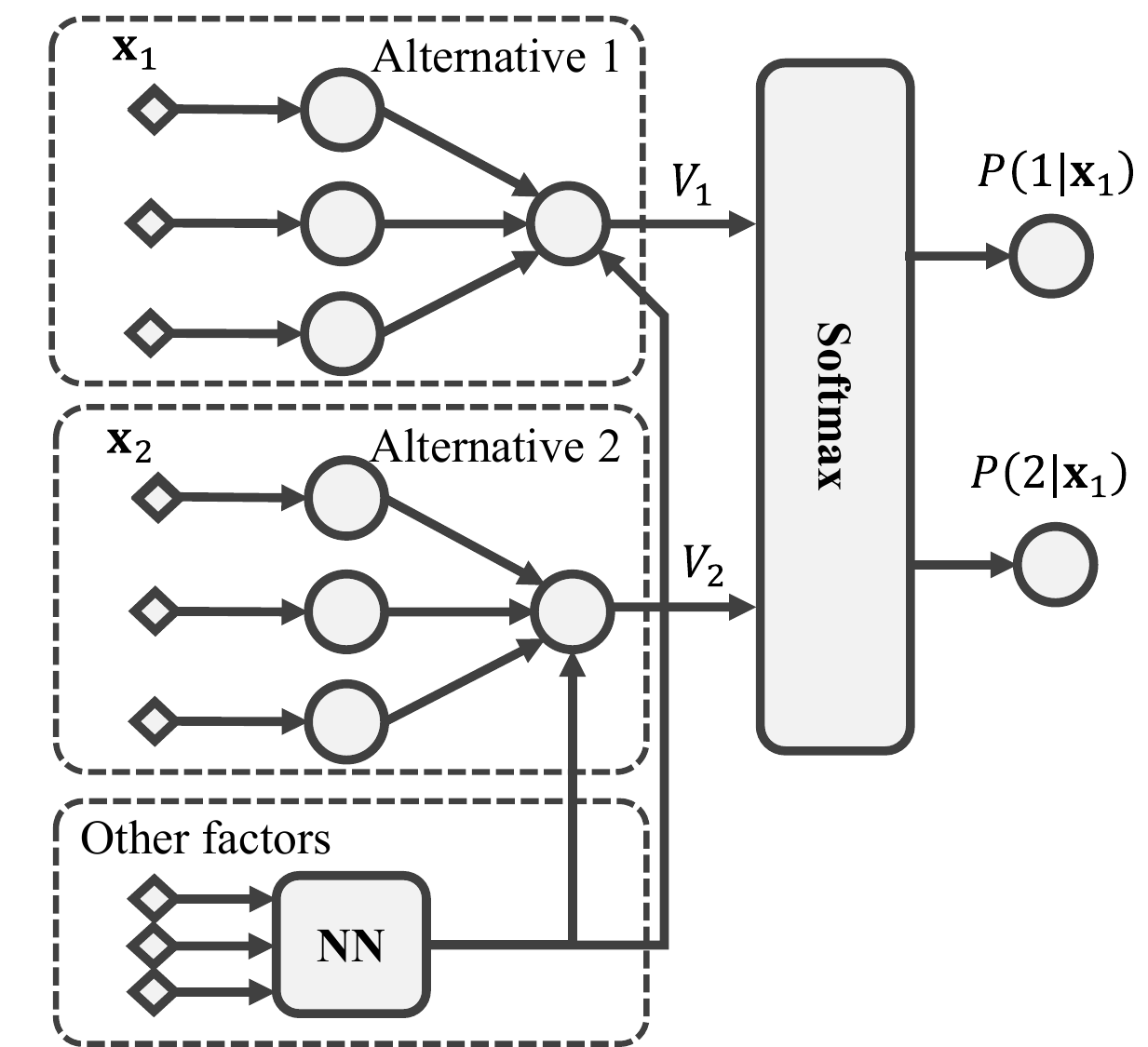}
    \subcaption{L-MNL}\label{fig:l-mnl}
    \end{minipage}
    \begin{minipage}[b]{0.34\linewidth}
    \centering
    \includegraphics[width=1.0\hsize]{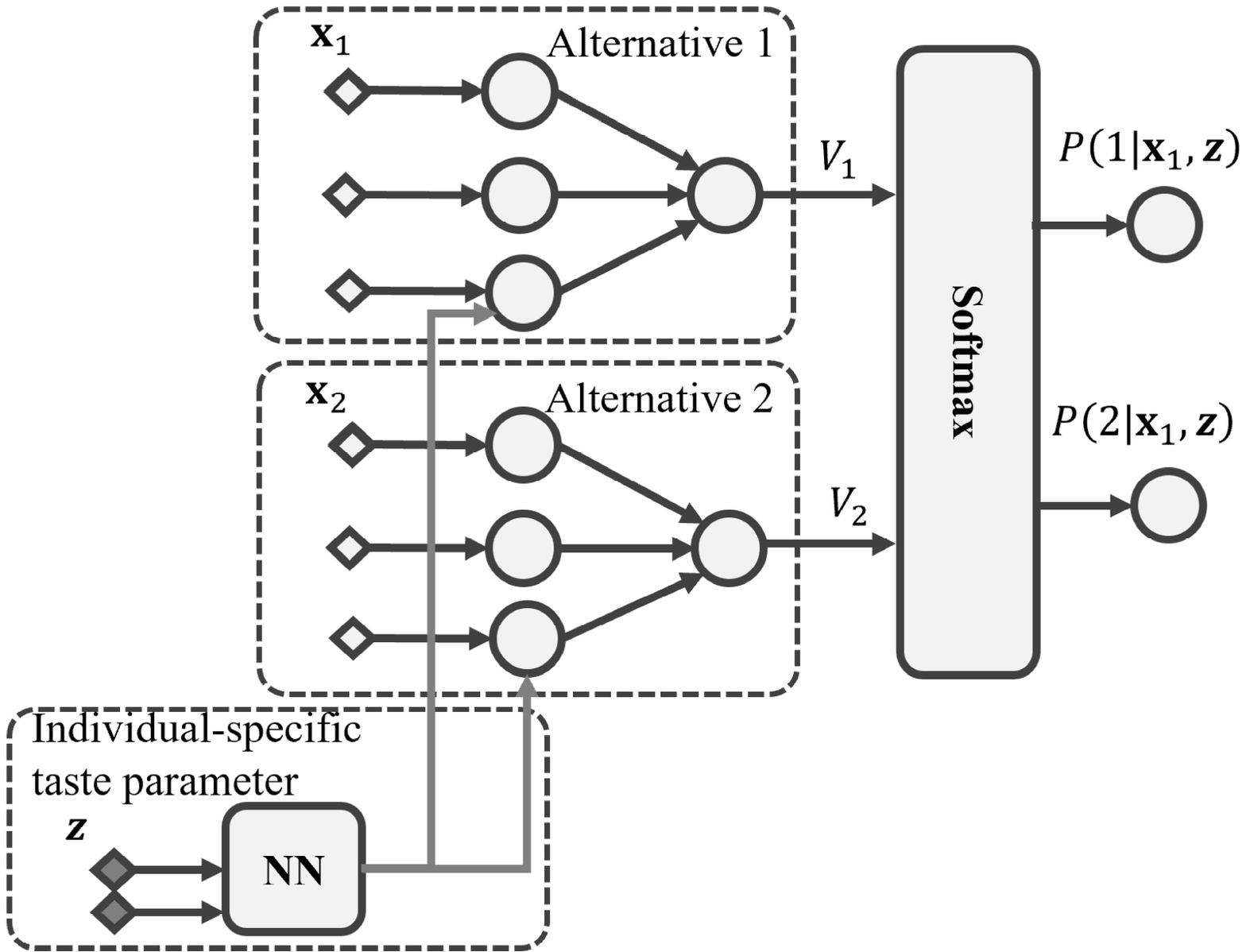}
    \subcaption{TasteNet-MNL}\label{fig:tastenet-mnl}
    \end{minipage}
    \caption{Multinomial Logit Model with Neural Networks.  ASU-DNN learns deterministic utilities of each alternative via individual neural networks. On the other hand, L-MNL learns the linear part of the utilities via the standard linear regression model and the nonlinear part via the neural networks. TasteNet-MNL uses a neural network to learn the nonlinear effects of the individual-specific variables and learns the deterministic utilities with the standard linear regression.}
    \label{fig:mnl-nn}
\end{figure}

Recent MNLs with NNs have different neural-network architectures. Figure \ref{fig:mnl-nn} shows the neural-network architectures of ASU-DNN, L-MNL, and TasteNet-MNL. In ASU-DNN, the deterministic utilities of each alternative are learned via individual NNs. Conversely, L-MNL learns the linear part of the utilities via the standard linear regression model and the nonlinear part via NNs. TasteNet-MNL uses an NN to learn the nonlinear effects of the individual-specific variables and learns the deterministic utilities with standard linear regression.

\begin{figure}[ht]
    \centering
    \begin{minipage}[b]{0.45\linewidth}
    \centering
        \includegraphics[width=1.0\hsize]{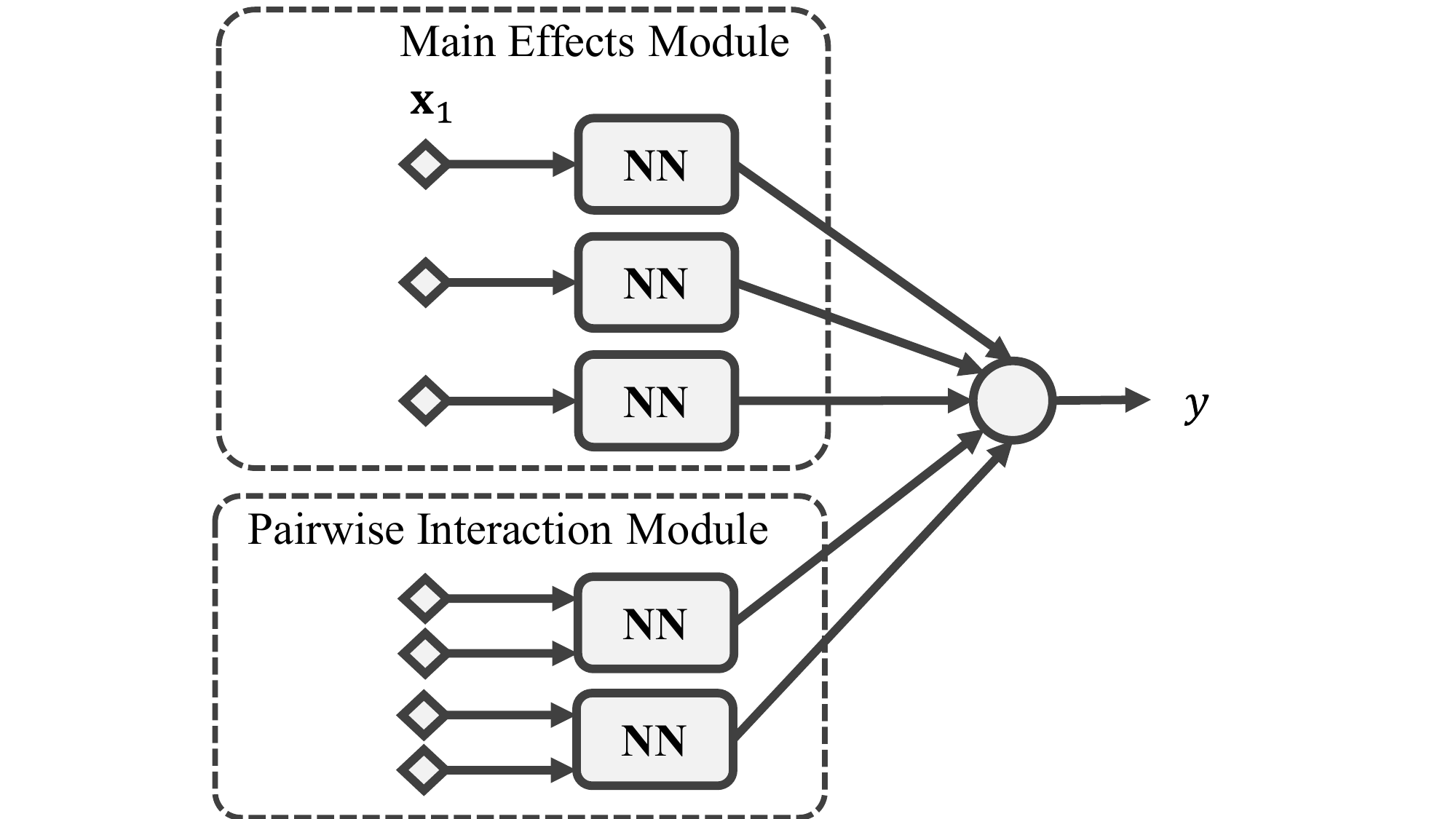}
    \subcaption{GAMI-Net}\label{fig:gaminet}
    \end{minipage}
    \begin{minipage}[b]{0.45\linewidth}
    \centering
        \includegraphics[width=0.7\hsize]{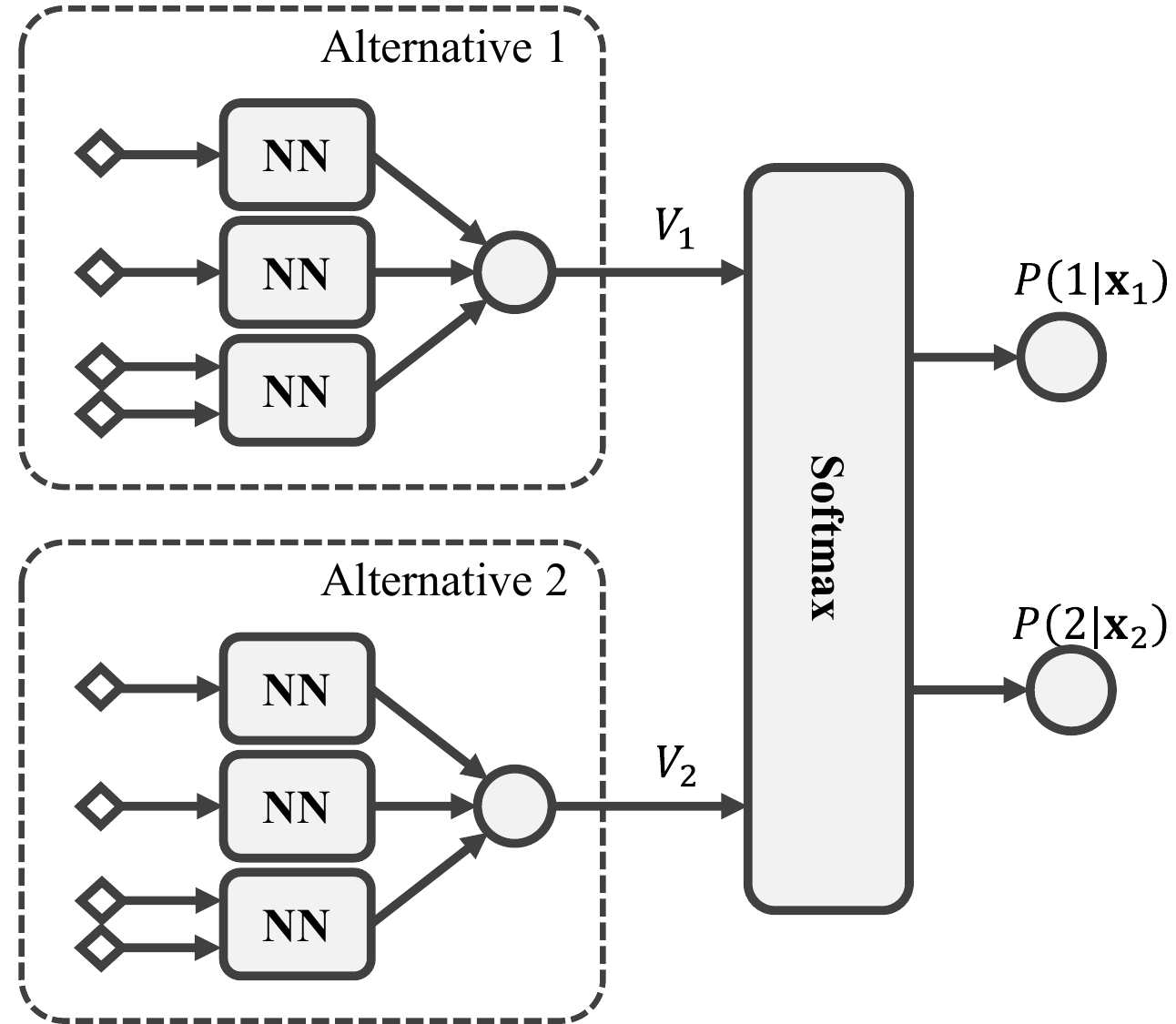}
    \subcaption{MNL-GAIUNet}\label{fig:mnl-gaiunet}
    \end{minipage}

    \caption{GAMI-Net and MNL-GAIUNet}\label{fig:gaminet-mnl-gaiunet}
\end{figure}

We show the network architecture of GAMI-Net, which is the generalized additive model with NNs in Fig. \ref{fig:gaminet}. GAMI-Net uses NNs as the shape functions of $\textrm{GA}^2\textrm{Ms}$. It consists of a main effect module and a pairwise interaction module. The main effect module captures the effects of each alternative-specific variable and the pairwise interaction module captures the effects of the interaction between the two variables. Our MNL-GAIUNet is the MNL that uses GAMI-Net to estimate utilities for each alternative (Fig. \ref{fig:mnl-gaiunet}).

\section{Discrete-Choice Model with Generalized Additive and Interactive Utility Networks} \label{sec:proposed-method}

\subsection{Generalized Additive and Interactive Utility Network} \label{subsec:GAIUNet}

We introduce two GAM-type utility functions using ANNs: GAUNet and GAIUNet. GAUNet is specified as the utility function by summing over each explanatory-specific utility represented by the NNs. GAUNet formulates the utility function for alternative $i \in \mathcal{A}$ as follows:
\begin{linenomath}
\begin{align}
    V_i(\mathbf{x}_i;\boldsymbol{\omega}_i,\boldsymbol{\theta}_i) &\coloneqq ASC_i + \sum_{j\in \mathcal{S}_i} \omega_{ij} \mathrm{NN}_{ij}(x_{ij};\theta_{ij}),\quad i \in \mathcal{A},\label{eq:gaunet-utility}
\end{align}
\end{linenomath}
where $ASC_i$ is an alternative-specific constraint and $\omega_{ij}$ is the parameter of the  $j$-th explanatory variable for alternative $i$. $\mathcal{S}_i$ is the set of explanatory variables for alternative $i$. $\mathrm{NN}_{ij}(x_{ij};\theta_{ij})$ is the NN that estimates the utility of the $j$-th explanatory variable for alternative $i$ with network parameters  $\theta_{ij}$. In other words, the NN $\textrm{NN}_{ij}$ depends only on the $j$-th explanatory variable. Let $\mathbf{x}_i$, $\boldsymbol{\omega}_i$, and  $\boldsymbol{\theta}_i$ be the vectors of the explanatory variables or parameters for alternative $i$.

GAIUNet further considers the pairwise interaction term between explanatory variables in addition to GAUNet as follows:
\begin{linenomath}
\begin{align}
V_i(\mathbf{x}_i;\boldsymbol{\omega}_i,\boldsymbol{\omega}^I_i,\boldsymbol{\theta}_i,\boldsymbol{\theta}^I_i) &\coloneqq ASC_i + \sum_{j\in \mathcal{S}_i} \omega_{ij} \mathrm{NN}_{ij}(x_{ij};\theta_{ij}) + \sum_{(j,k)\in \mathcal{S}_i^I} \omega_{ijk}^I \mathrm{NN}_{ijk}(x_{ij},x_{ik};\theta_{ijk}^I) ,\quad i \in \mathcal{A},\label{eq:gaiunet-utility}
\end{align}
\end{linenomath}
where $\omega^I_{ijk}$ and $\theta^I_{ijk}$ are parameters of the pairwise interaction terms between the $j$-th and $k$-th variables for alternative $i$. $\mathcal{S}_i^I$ is the set of considered pairs of explanatory variables. $\boldsymbol{\omega}^I_i$ and $\boldsymbol{\theta}^I_i$ are the vectors of parameters of the pairwise interaction terms for the alternative $i$. $\mathrm{NN}_{ijk}$ formulates the interaction among the $j$-th and $k$-th explanatory variables.

We compare GAUNet and GAIUNet with the four utility functions of previous studies in Table \ref{tab:utility-functions}; MNL-Linear $ASC_i + \boldsymbol{\omega}_{i}^\top \mathbf{x}_{i}$, Fukuda\&Yai $ASC_i + \mathrm{SCS}_i(\mathbf{x}_i)$, ASU-DNN $ASC_i + \mathrm{NN}_i(\mathbf{x}_{i})$, L-MNL/Taste-Net-MNL $ASC_i + \boldsymbol{\omega}_{i}^\top \mathbf{x}_{i} + \mathrm{NN}(\mathbf{z})$, where $SCS_i(\cdot)$ is the smoothing cubic spline, and $\mathbf{z}$ represents the vector explanatory variables other than $\mathbf{x}_i,\forall i\in \mathcal{A}$. 
\begin{table}[ht]
    \centering
    \rowcolors{2}{}{gray!15}
    \small
    \caption{Comparison of GAUNet and GAIUNet with utility functions of previous studies. $\mathrm{SCS}_i(\cdot)$ is the smoothing cubic spline, and $\mathbf{z}$ represents the vector explanatory variables other than $\mathbf{x}_i,\forall i\in \mathcal{A}$.}
    \label{tab:utility-functions}
    \begin{tabular}{ll}
        \toprule
         Methods & Formulation of Deterministic Utility Function \\
        \midrule
        MNL-Linear & $ASC_i + \boldsymbol{\omega}_{i}^\top \mathbf{x}_{i}$\\
        Fukuda\&Yai\cite{fukuda2010semiparametric} & $ASC_i + \mathrm{SCS}_i(\mathbf{x}_i)$\\
        ASU-DNN\cite{wang2020deep} & $ASC_i + \mathrm{NN}_i(\mathbf{x}_{i})$\\
        L-MNL\cite{sifringer2020enhancing} & $ASC_i + \boldsymbol{\omega}_{i}^\top \mathbf{x}_{i} + \mathrm{NN}(\mathbf{z})$\\
        Taste-Net-MNL \cite{han2020neural} & $ASC_i + \boldsymbol{\omega}_{i}^\top \mathbf{x}^{\mathrm{MNL}}_{i} + \mathrm{NN}^{\top}(\mathbf{z}_n) \mathbf{x}^{\mathrm{TN}}_{in}   $\\
        GAUNet & $ASC_i + \sum_{j\in \mathcal{S}_i} \omega_{ij} \mathrm{NN}_{ij}(x_{ij})$ \\
        GAIUNet & $ASC_i + \sum_{j\in \mathcal{S}_i} \omega_{ij} \mathrm{NN}_{ij}(x_{ij}) + \sum_{(j,k)\in \mathcal{S}_i^I} \omega_{ijk}^I \mathrm{NN}_{ijk}(x_{ij},x_{ik})$\\
        \bottomrule
    \end{tabular}
\end{table}

\subsection{MNL-GAUNet}  \label{subsec:MNL-GAUNet}

We introduce MNL with GAUNet (MNL-GAUNet) as the deterministic utility function represented with Eq.~\eqref{eq:gaunet-utility}. The choice probability for each alternative $i\in\mathcal{A}$ can be calculated as follows:
\begin{linenomath}
\begin{align}
    P(i|\mathbf{x};\boldsymbol{\omega},\boldsymbol{\theta}) &= \frac{\exp(V_i(\mathbf{x}_i;\boldsymbol{\omega},\boldsymbol{\theta}))}{\sum_{i\in \mathcal{A}}\exp(V_i(\mathbf{x}_i;\boldsymbol{\omega},\boldsymbol{\theta}))},\quad i \in \mathcal{A}.
\end{align}
\end{linenomath}
We find the neural-network parameters $\boldsymbol{\omega}$ and $\boldsymbol{\theta}$ by maximizing the following objective function: $O(\mathcal{D};\boldsymbol{\omega},\boldsymbol{\theta})$ with respect to the data $\mathcal{D}$:
\begin{linenomath}
\begin{align}
&\boldsymbol{\omega},\boldsymbol{\theta} = \argmax_{\boldsymbol{\omega},\boldsymbol{\theta}} O(\mathcal{D};\boldsymbol{\omega},\boldsymbol{\theta}),\label{eq:mnl-gaunet}\\
& O(\mathcal{D};\boldsymbol{\omega},\boldsymbol{\theta}) \coloneqq \mathrm{LL}(D;\boldsymbol{\omega},\boldsymbol{\theta})+ \alpha L_1(D;\boldsymbol{\omega}),\\
&\mathrm{LL}(D;\boldsymbol{\omega},\boldsymbol{\theta}) \coloneqq \sum_{(y_d,\mathbf{x}_d)\in D}\log P(y_d|\mathbf{x}_d;\boldsymbol{\omega},\boldsymbol{\theta}), \label{eq:mnl-gaunet-likelihood}\\
&L_1(D;\boldsymbol{\omega}) \coloneqq \sum_{i \in \mathcal{A}}\sum_{j\in \mathcal{S}_i} |\omega_{ij}|, \label{eq:l1-norm}
\end{align}
\end{linenomath}
where $L_1(\cdot)$ denotes the $L_1$ norm for the weights of the NNs, and $\alpha (\leq 0)$ is the regularization parameter for the $L_1$ norm with a negative value. We can find the sparse solution for $\boldsymbol{\omega}$ due to the $L_1$ norm. Stochastic gradient descent approaches such as Adam are often used to find solutions.

We define the importance score for each explanatory variable as follows:
\begin{linenomath}
\begin{align}
&\textrm{Imp}(x_{ij};\boldsymbol{\omega},\boldsymbol{\theta}) \coloneqq \frac{\int_{\min{(x_{ij})}}^{\max{(x_{ij})}}\omega_{ij}v_{ij}(x_{ij};\theta_{ij})dx_{ij}}{(\max{(x_{ij})}-\min{(x_{ij})})} ,\quad i \in \mathcal{A}, j\in \mathcal{S}_i, \label{eq:mnl-gaunet-importance}
\end{align}
\end{linenomath}
We calculate $\mathrm{Imp}(x_{ij};\boldsymbol{\omega},\boldsymbol{\theta})$ using evenly partitioned values in the range of $\min(x_{ij})$ to $\max(x_{ij})$ because it would be difficult to calculate the integral directly. Our MNL-GAUNet is equivalent to MNL-Linear when the number of neural-network layers for all explanatory variables is one, and the activation functions are linear.

Unlike conventional methods such as ASU-DNN, our approach uses NNs to model the utility function of each variable independently, allowing us to analyze universal characteristics by pooling variable weights across all alternatives.

\subsection{MNL-GAIUNet} \label{subsec:MNL-GAIUNet}

We also develop the MNL with GAIUNet, in which the deterministic utility is specified in Eq.~\eqref{eq:gaiunet-utility}. The choice probability is derived as follows:
\begin{linenomath}
\begin{align}    
P(i|\mathbf{x};\boldsymbol{\omega},\boldsymbol{\omega}^I,\boldsymbol{\theta},\boldsymbol{\theta}^I) &= \frac{\exp(V_i(\mathbf{x}_i;\boldsymbol{\omega},\boldsymbol{\omega}^I,\boldsymbol{\theta},\boldsymbol{\theta}^I))}{\sum_{i\in \mathcal{A}}\exp(V_i(\mathbf{x}_i;\boldsymbol{\omega},\boldsymbol{\omega}^I,\boldsymbol{\theta},\boldsymbol{\theta}^I))},\quad i \in \mathcal{A}.\label{eq:mnl-gaiunet-prob}
\end{align}
\end{linenomath}
We find the parameters by maximizing the following objective function $O(D;\boldsymbol{\omega},\boldsymbol{\omega}^I,\boldsymbol{\theta},\boldsymbol{\theta}^I)$:
\begin{linenomath}
\begin{align}
&\boldsymbol{\omega},\boldsymbol{\omega}^I,\boldsymbol{\theta},\boldsymbol{\theta}^I = \argmax_{\boldsymbol{\omega},\boldsymbol{\omega}^I,\boldsymbol{\theta},\boldsymbol{\theta}^I} O(D;\boldsymbol{\omega},\boldsymbol{\omega}^I,\boldsymbol{\theta},\boldsymbol{\theta}^I), \label{eq:mnl-gaiunet-argmax}\\
    &O (D;\boldsymbol{\omega},\boldsymbol{\omega}^I,\boldsymbol{\theta},\boldsymbol{\theta}^I) \coloneqq \mathrm{LL}(D;\boldsymbol{\omega},\boldsymbol{\omega}^I,\boldsymbol{\theta},\boldsymbol{\theta}^I) + 
     \alpha L_1(D;\boldsymbol{\omega})+ \alpha^I L_1^I(D;\boldsymbol{\omega}^I)+ \beta^I \Omega(D;\boldsymbol{\omega},\boldsymbol{\omega}^I,\boldsymbol{\theta},\boldsymbol{\theta}^I),\\    
    &\mathrm{LL}(D;\boldsymbol{\omega},\boldsymbol{\omega}^I,\boldsymbol{\theta},\boldsymbol{\theta}^I) \coloneqq \sum_{(y_k,\mathbf{x}_k)\in D}\log P(a_k|\mathbf{x}_k;\boldsymbol{\omega},\boldsymbol{\omega}^I,\boldsymbol{\theta},\boldsymbol{\theta}^I),\label{eq:mnl-gaiunet-loglikelihood}\\
    &L_1^I(D;\boldsymbol{\omega}^I) \coloneqq \sum_{i\in \mathcal{A}}\sum_{(j,k) \in \mathcal{S}_i^I}|\omega_{ijk}^I|,\label{eq:mnl-gaiunet-l1-I}\\
    &\Omega(D;\boldsymbol{\omega},\boldsymbol{\omega}^I,\boldsymbol{\theta},\boldsymbol{\theta}^I) \coloneqq \sum_{x_i \in D} \left|\frac{1}{|\mathcal{S}_i^I|}\sum_{(j,k)\in \mathcal{S}_i^I}v_{ij}(x_{ij};\theta_{ij})v_{ijk}(x_{ij},x_{ik};\theta_{ijk}^I)\right|, \label{eq:mnl-gaiunet-Omega}
\end{align}
\end{linenomath}
where $L^I_1$ denotes the $L_1$ norm for the weight $\omega_{ijk}^I$ and $\alpha^I (\leq 0)$ is the regularization parameter for the norm. $\Omega(\cdot)$ is a marginal clarity constraint representing the degree of non-orthogonality between each main effect $v_{ij}$ and pairwise interactions $v_{ijk}$ \cite{yang2021gami}. The constraint is necessary because $v_{ij}$ would easily be absorbed by $v_{ijk}$. We optimize the problem using stochastic gradient descent methods such as the Adam optimizer.

The importance score of the interactions is defined as follows: 
\begin{linenomath}\begin{align}
\textrm{Imp}(x_{ij},x_{ik};\boldsymbol{\omega},
\boldsymbol{\theta}) \coloneqq \frac{\int_{\min{(x_{ij})}}^{\max{(x_{ij})}}\int_{\min{(x_{ik})}}^{\max{(x_{ik})}}\omega_{ijk}^Iv_{ijk}(x_{ij},x_{ik};\theta_{ijk}^I)}{(\max{(x_{ij})}-\min{(x_{ij})})(\max{(x_{ik})}-\min{(x_{ik})})},\quad i \in \mathcal{A},( j,k)\in \mathcal{S}_i^I \label{eq:mnl-gaiunet-importance}
\end{align}\end{linenomath}
We calculate the importance score using evenly partitioned values in the range of $\min(x_{ij})$ to $\max(x_{ij})$ and $\min(x_{ik})$ to $\max(x_{ik})$  because it would be difficult to calculate the integral directly. 

Algorithm~\ref{alg:mnl-gaiunet} shows the procedure for finding the parameters with MNL-GAIUNet. First, the parameters for the main effects are optimized based on Eq.~\eqref{eq:mnl-gaunet} (Line~\ref{line:mnl-gaunet}). Then, we find the explanatory variables with greater importance score than the threshold $\mathrm{Th}_{\mathrm{Imp}}$ and create the set of their variable pairs because the number of combinations of all variables is often too large for computation (Line~\ref{line:calc-importance}-\ref{line:select-pairs}). The assumption is premised on the fact that the importance score of variable pairs with smaller scores would rarely be large. The parameters and weights for the pairwise interactions in MNL-GAIUNet are learned by fixing the parameters and weights of the main effect based on Eq.~\eqref{eq:mnl-gaiunet-argmax} (Line~\ref{line:mnl-gaiunet}). Finally, we fine-tune all parameters and weights based on Eq.~\eqref{eq:mnl-gaiunet-argmax} (Line~\ref{line:finetuning}).

\begin{algorithm}[t] 
\small
\centering
\caption{\small Learning MNL-GAIUNet} \label{alg:mnl-gaiunet}
  \begin{algorithmic}[1]
  \State $\mathcal{S}_i \gets \{1,\dots,M\}$ 
  \State $\boldsymbol{\tilde{\omega}},\boldsymbol{\tilde{\theta}} \gets \argmax_{\boldsymbol{\omega},\boldsymbol{\theta}} \left(\textrm{LL}(D;\boldsymbol{\omega},\boldsymbol{\theta}) + \alpha L_1(D;\boldsymbol{\omega})\right)$ \label{line:mnl-gaunet}
  \State $\tilde{\mathcal{S}}_i \gets \{i | i \in \mathcal{S}_i, \mathrm{Imp}(x_{ij};\boldsymbol{\tilde{\omega}},\boldsymbol{\tilde{\theta}})\geq \mathrm{Th}_{\mathrm{Imp}}\}$ \label{line:calc-importance}
  \State $\mathcal{S}_i^I \gets \{(j,k) | j\in \tilde{S}_i , k \in \tilde{S}_i, j \neq k\} $ \label{line:select-pairs}
  \State $\boldsymbol{\omega}^I,\boldsymbol{\theta}^I = \argmax_{\boldsymbol{\omega}^I,\boldsymbol{\theta}^I} \left(\textrm{LL}(D;\boldsymbol{\tilde{\omega}},\boldsymbol{\omega}^I,\boldsymbol{\theta},\boldsymbol{\theta}^I) +\alpha L_1(D;\boldsymbol{\omega})+ \alpha^I L_1^I(D;\boldsymbol{\omega}^I)+ \beta^I \Omega(D;\boldsymbol{\omega},\boldsymbol{\omega}^I,\boldsymbol{\theta},\boldsymbol{\theta}^I)\right)$ \label{line:mnl-gaiunet}
  \State $\boldsymbol{\omega},\boldsymbol{\omega}^I,\boldsymbol{\theta},\boldsymbol{\theta}^I = \argmax_{\boldsymbol{\omega},\boldsymbol{\omega}^I,\boldsymbol{\theta},\boldsymbol{\theta}^I} \left(\mathrm{LL}(D;\boldsymbol{\omega},\boldsymbol{\omega}^I,\boldsymbol{\theta},\boldsymbol{\theta}^I)  +\alpha L_1(D;\boldsymbol{\omega})+ \alpha^I L_1^I(D;\boldsymbol{\omega}^I)+ \beta^I \Omega(D;\boldsymbol{\omega},\boldsymbol{\omega}^I,\boldsymbol{\theta},\boldsymbol{\theta}^I)\right)$\label{line:finetuning}
  \end{algorithmic}
\end{algorithm}

\section{Experiments for Synthetic Data} \label{sec:experiment-synthetic-data}

In this section, we evaluate our models using synthetic data that simulate the mode choice of transportation between buses and cars with unobservable constraints.

\subsection{Experimental Settings}

We assumed that the transportation modes were chosen based on MNL. We designed the utility $U$ for each transportation mode to represent the unobservable constraints as follows:
\begin{linenomath}\begin{align}
    V^B &\coloneqq
    \begin{cases}
        -0.5x^B_c - 0.2 x^B_t -0.3 x^B_{a} - 0.4x^B_{e}& \mathrm{if}~ x^B_a \leq 4\\
        -\infty & \mathrm{otherwise}
    \end{cases},\\
    U^B &\coloneqq  V^B+ \mathrm{Gumbel}(0,1),\\
    V^C &\coloneqq 
    \begin{cases}
        -0.5x^C_c - 0.2 x^C_t -0.3 x^C_{a} - 0.4x^C_{e} & \mathrm{if}~ x^C_c \leq 20\\
        -\infty & \mathrm{otherwise}
    \end{cases},\\
    U^C &\coloneqq V^C + \mathrm{Gumbel} (0,1),
\end{align}\end{linenomath}
where $V$ is the deterministic utility. Let superscripts $B$  and $C$ be Bus and Taxi, respectively. $x_c$, $x_t$, $x_a$, and $x_e$ denote the explanatory variables for travel cost, travel time, access time, and egress time, respectively. The random utility was assumed to be $\mathrm{Gumbel}(0,1)$.  That is, the choice probability of each model is calculated as follows:
\begin{linenomath}\begin{align}
    P^B =\frac{\exp{(V^B)}}{\sum_{m\in\{B,C\}}{\exp{(V^{m})}}},\\
    P^C =\frac{\exp{(V^C)}}{\sum_{m\in\{B,C\}}{\exp{(V^{m})}}},
\end{align}\end{linenomath}
In the above model, the bus is not selected if the access time exceeds four minutes, whereas the car is not selected if the cost exceeds \$20. However, we assumed that the constraints were not met. We generated 10,000 synthetic data points, which are the set of cost, travel time, access time, egress time for each mode and the result of mode selection based on the MNL, and recorded 6,201 sets that satisfied at least one of the constraints in the dataset. Each variable was drawn with the uniform distributions described in Tab.~\ref{tab:param_exp1}. 
\begin{table}
\rowcolors{2}{}{gray!15}
    \begin{center}
    \small
        \caption{Explanatory variable range for each transportation mode}
    \label{tab:param_exp1}
    \begin{tabular}{ccc} 
    \toprule
        \textbf{Explanatory Variable}&  \textbf{Bus}& \textbf{Taxi}\\ 
    \midrule         
         Cost [dollars]&  [2, 6]& [10, 35]\\          
         Travel Time [minutes]&  [15, 60]& [10, 20]\\
         Access Time [minutes]&  [2, 10]& [0, 2]\\ 
         Egress Time [minutes]&  [2, 10]& [0, 2]\\ 
    \bottomrule
    \end{tabular}
    \end{center}
\end{table}

\subsection{Experimental Results}

\subsubsection{Prediction accuracy and interpretability}

We evaluated three methods: MNL-Linear, ASU-DNN, and MNL-GAUNet. For ASU-DNN and MNL-GAUNet, we compared the performances of Tanh and LeakyReLU as the activation functions. For MNL-Linear and MNL-GAUNet, we also compared the cases where the model parameters were shared between the two transportation modes and where they were not. We evaluated the performance with the log-likelihood and accuracy, which is calculated as the agreement ratio between the predicted and selected modes. 

We present the prediction accuracy results in Tab.~\ref{tab:log-likelihood-syntheticdata}. In the table, -SW denotes the shared parameter model and Tanh/LeakyReLU denotes the activation function of the model. All methods achieved accuracies better than 0.9. For MNL-GAUNet, LeakyReLU outperformed Tanh, whereas the opposite was true for ASU-DNN. Both ASU-DNN and MNL-GAUNet achieved slightly better performance than MNL-Linear in terms of both log-likelihood and accuracy, while MNL-GAUNet showed comparable performance to ASU-DNN. 

\begin{table}[t]
    \centering
    \rowcolors{2}{}{gray!15}
    \small
    \caption{Log-likelihood for the model without shared weights }\label{tab:log-likelihood-syntheticdata}
    \begin{tabular}{lrr}
    \toprule
 \textbf{Methods}& \textbf{Log-Likelihood}& \textbf{Accuracy}\\
    \midrule
    MNL-Linear & -130& 0.95\\
 MNL-Linear-SW& -171&0.94\\
    \midrule
    ASU-DNN (Tanh)& -260& 0.92\\
    ASU-DNN (LeakyReLU)& -10& 1.0\\
    \midrule
    MNL-GAUNet (Tanh)& -13& 0.99\\
    MNL-GAUNet (LeakyReLU)& -49& 0.98\\
 MNL-GAUNet-SW (Tanh)& -17&0.99\\
 MNL-GAUNet-SW (LeakyReLU)& -54&0.98\\
 \bottomrule
    \end{tabular}
\end{table}

\begin{figure}[t]
    \centering
    \begin{minipage}[b]{0.32\linewidth}
    \centering
    \includegraphics[width=1.0\hsize]{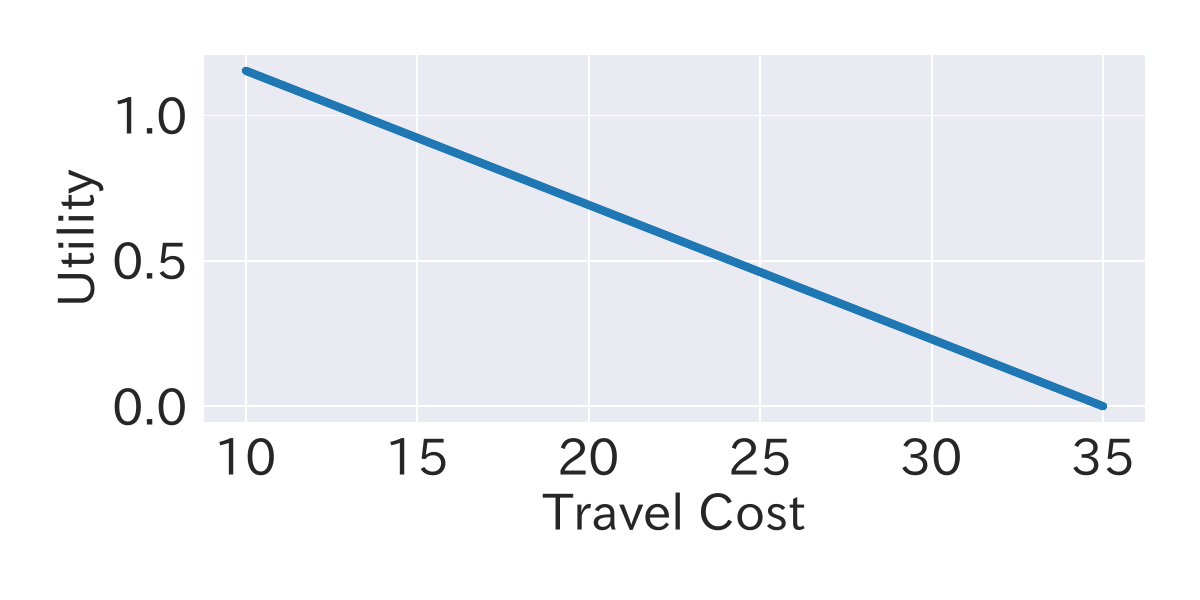}
    \subcaption{MNL-Linear}\label{fig:synthetic-mnl-linear-travel-cost}
    \end{minipage}
    \begin{minipage}[b]{0.32\linewidth}
    \centering
    \includegraphics[width=1.0\hsize]{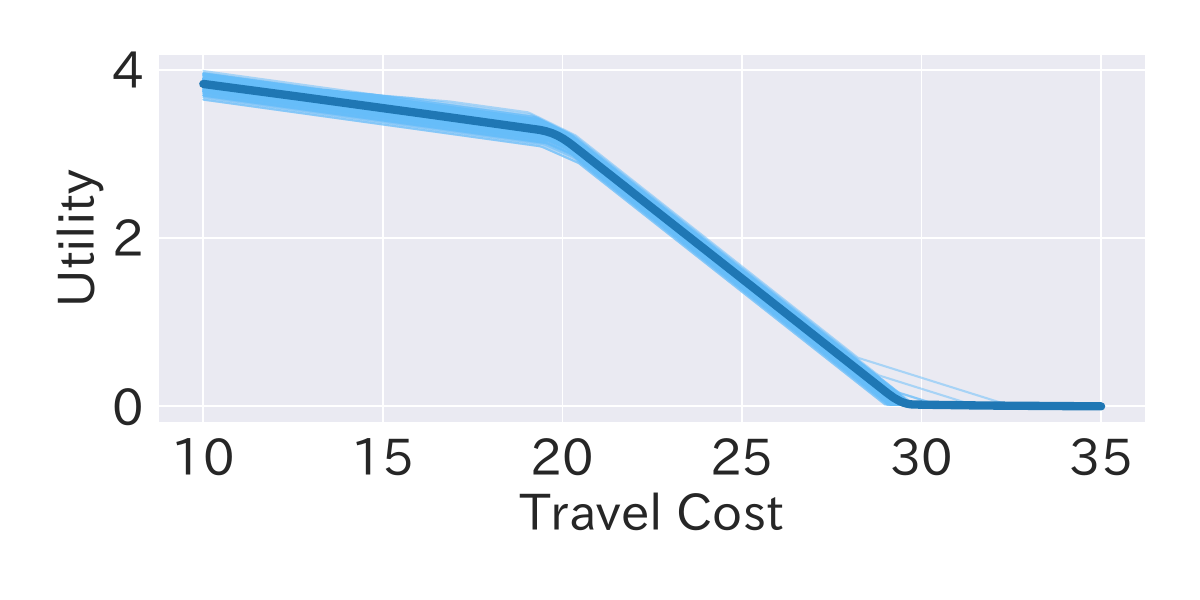}
    \subcaption{ASUDNN(LeakyReLU)}\label{fig:synthetic-asudnn-travel-cost}
    \end{minipage}
    \begin{minipage}[b]{0.32\linewidth}
    \centering
    \includegraphics[width=1.0\hsize]{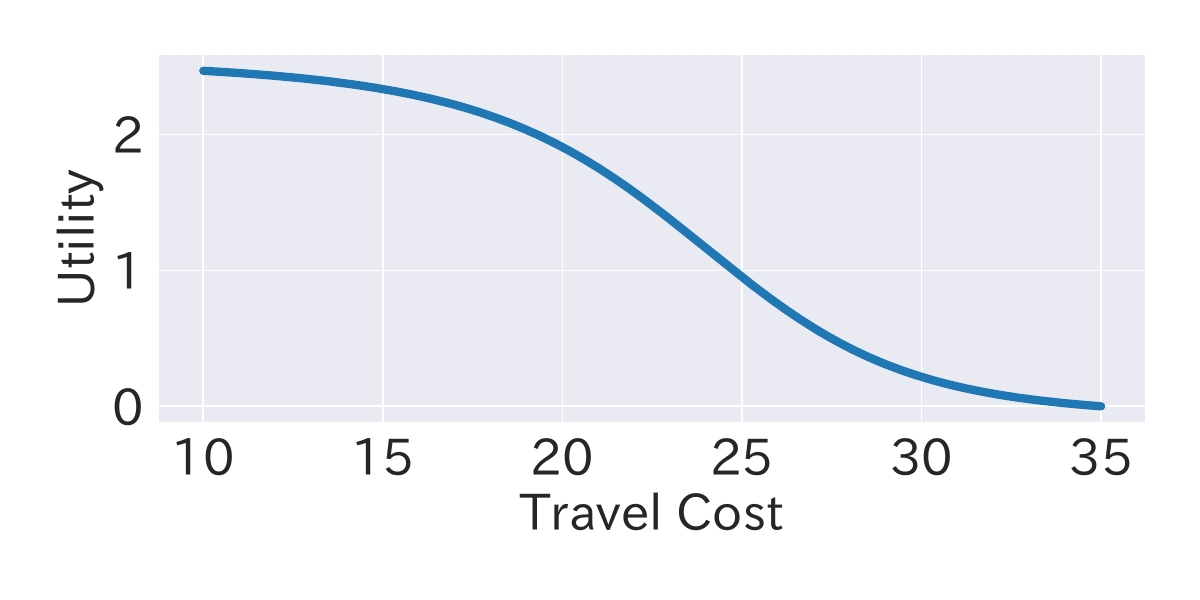}
    \subcaption{MNL-GAIUNet(Tanh)}\label{fig:synthetic-GAUNet-travel-cost}
    \end{minipage}
    \caption{The effects for taxi travel cost of buses based on (a) MNL-Linear, (b) ASU-DNN(LeakyReLU) and (c) MNL-GAUNet(Tanh). In (b), the thin lines were computed with the values included in the dataset as the values of the other explanatory variables. The bold line is the median.}
    \label{fig:synthetic-travel-cost}
\end{figure}

\begin{figure}[t]
    \centering
    \begin{minipage}[b]{0.32\linewidth}
    \centering
    \includegraphics[width=1.0\hsize]{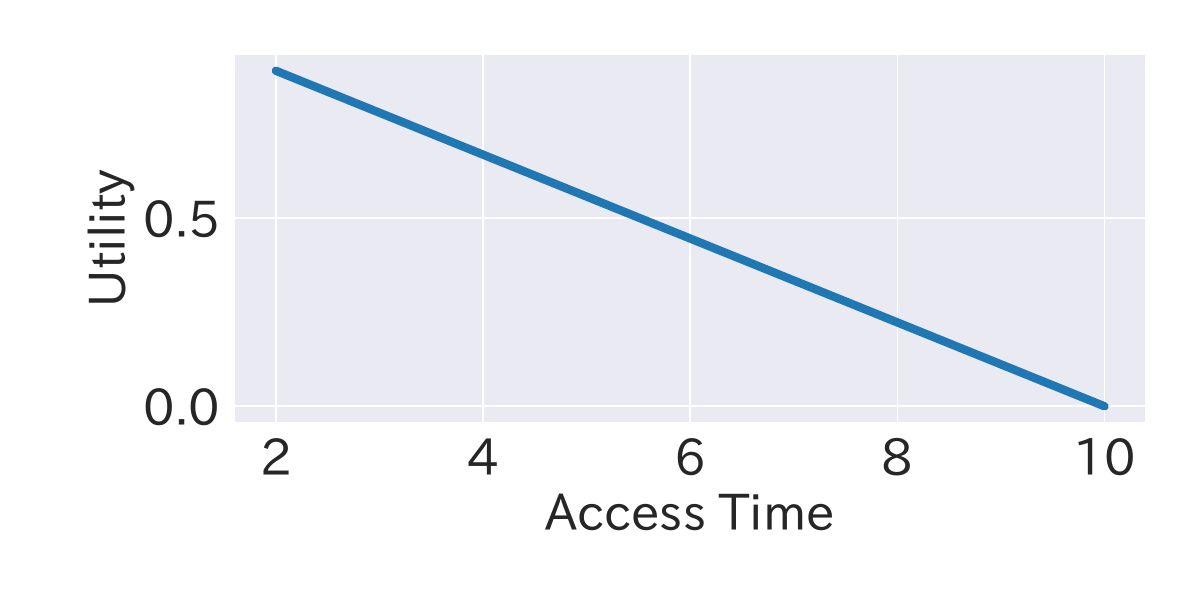}
    \subcaption{MNL-Linear}\label{fig:synthetic-mnl-linear-access-time}
    \end{minipage}
    \begin{minipage}[b]{0.32\linewidth}
    \centering
    \includegraphics[width=1.0\hsize]{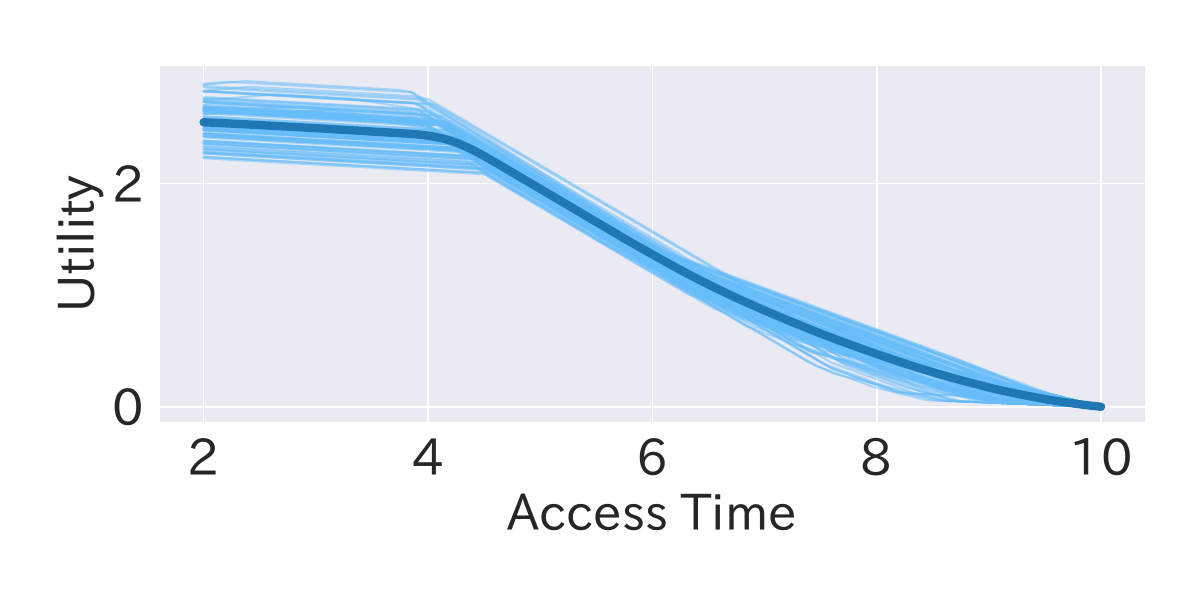}
    \subcaption{ASUDNN(LeakyReLU)}\label{fig:synthetic-asudnn-access-time}
    \end{minipage}
    \begin{minipage}[b]{0.32\linewidth}
    \centering
    \includegraphics[width=1.0\hsize]{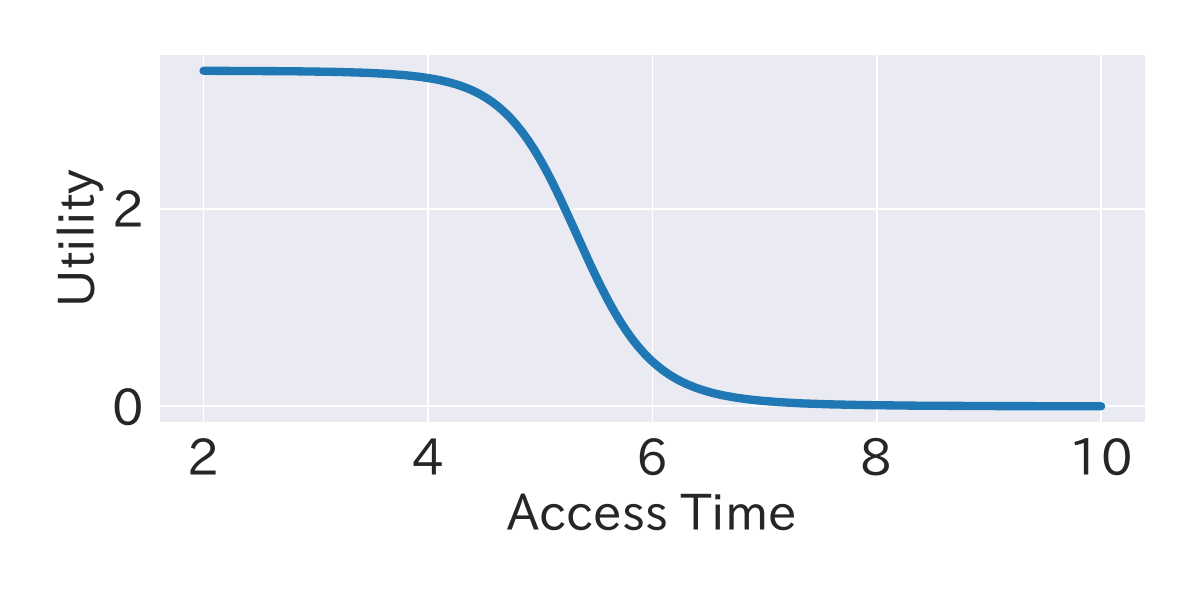}
    \subcaption{MNL-GAIUNet(Tanh)}\label{fig:synthetic-GAUNet-access-time}
    \end{minipage}
    \caption{The effects of (a) MNL-Linear, (b) ASU-DNN(LeakyReLU) and (c) MNL-GAUNet(Tanh) on the bus access time. In (b), the thin lines were computed with the values included in the dataset as the values of the other explanatory variables. The bold line is the median.}
    \label{fig:synthetic-access-time}
\end{figure}

The learned utilities of each variable help in understanding the decision-making of users. We show the effects of taxi travel cost and bus access time for the models with the largest log-likelihood for each method (i.e., MNL-Linear, ASU-DNN (LeakyReLU), and MNL-GAUNet (Tanh)) in Fig.~\ref{fig:synthetic-travel-cost} and \ref{fig:synthetic-access-time}.  In calculating the utility of each explanatory variable in ASU-DNN, the values of other explanatory variables included in the dataset were used because the utility also depends on the values of the other variables. In Fig.~\ref{fig:synthetic-travel-cost}~(b) and \ref{fig:synthetic-access-time}~(b), the thin lines represent the utility computed with the values. The bold lines are the medians. On the other hand, we draw single bold lines in MNL-Linear and MNL-GAUNet~(Tanh) because these models are independent of the values of the other explanatory variables. 

Figure~\ref{fig:synthetic-travel-cost} shows that the effects of all models monotonically decrease with increasing travel cost. The direction of the change is reasonable because higher travel costs suppress the usage of the mode. However, the change in the effect at approximately 20 dollars differed between the MNL-Linear model and the other neural network-based models. For ASU-DNN (LeakyReLU) and MNL-GAUNet (Tanh), the effect decayed drastically at approximately 20 dollars, whereas for MNL-Linear, it did not. The results suggest that the neural network-based models can capture unobservable constraints on not selecting taxis that cost more than 20 dollars. 

The results shown in Fig.~\ref{fig:synthetic-access-time} also indicate that while all models captured the broad decreasing trend with respect to the bus access time, only the neural network-based models captured the significant decay around the access time constraints~(i.e., four minutes), which was the same as the effect of taxi travel costs. The shapes of the travel costs and the access time are slightly different between ASU-DNN (LeakyReLU) and MNL-GAUNet (Tanh) because of the activation functions used, but we found that the difference has little effect on the prediction accuracy~(Tab.~\ref{tab:log-likelihood-syntheticdata}).

\begin{figure}[t]
    \centering
    \begin{minipage}[b]{0.32\linewidth}
    \centering
    \includegraphics[width=1.0\hsize]{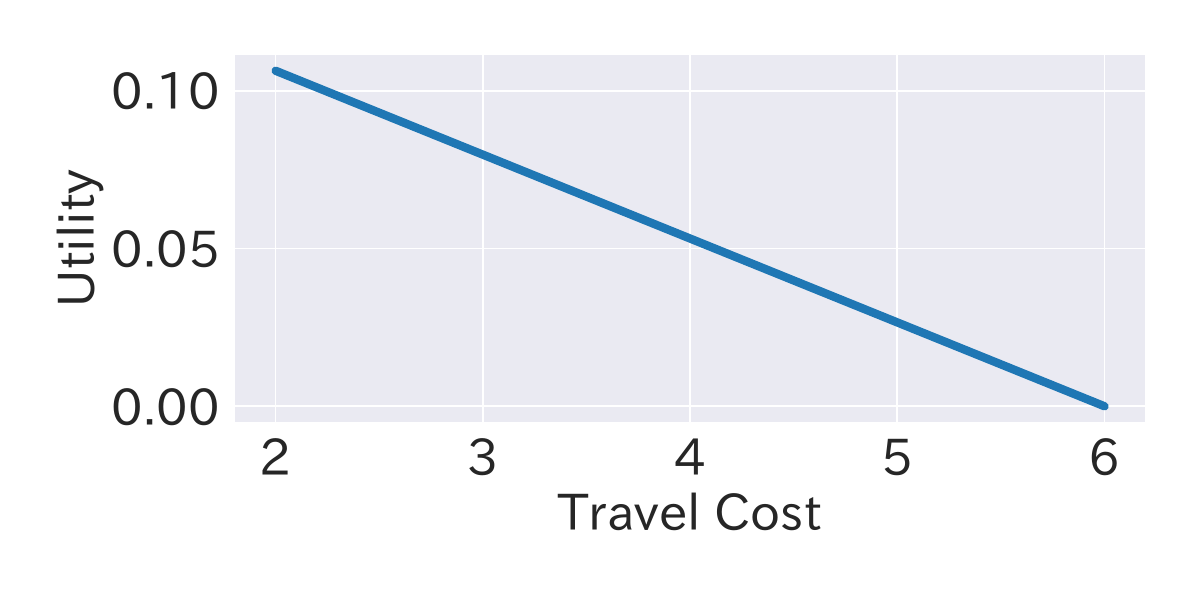}
    \subcaption{MNL-Linear: Travel Cost}\label{fig:synthetic-mnllinear-bus-travelcost2}
    \end{minipage}
    \begin{minipage}[b]{0.32\linewidth}
    \centering
    \includegraphics[width=1.0\hsize]{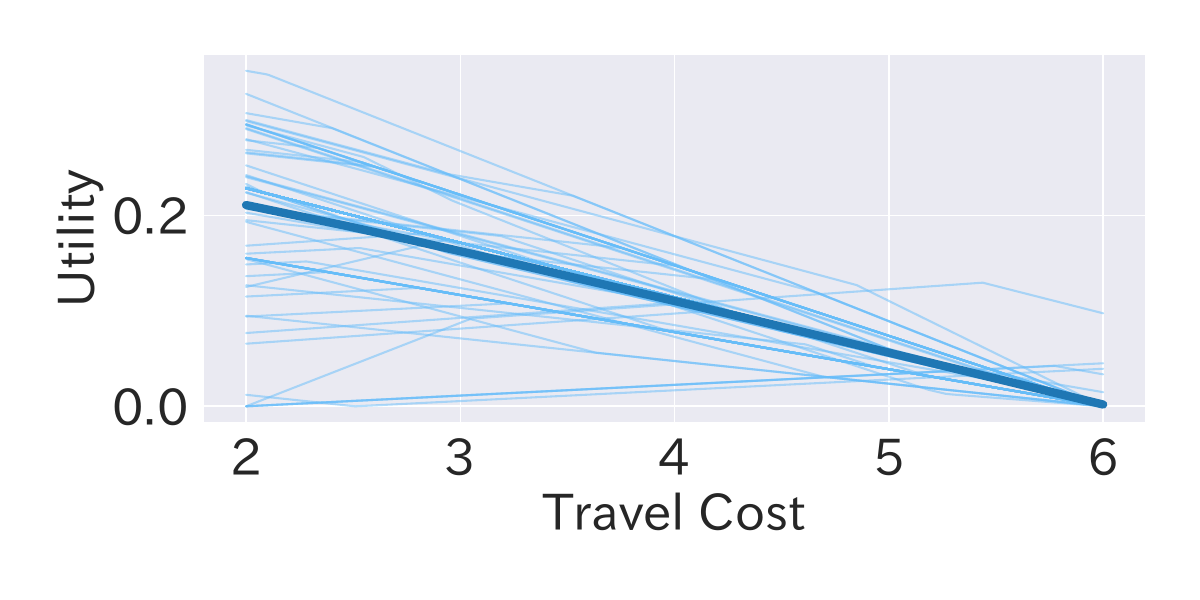}
    \subcaption{ASU-DNN: Travel Cost}\label{fig:synthetic-asudnn-bus-travelcost2}
    \end{minipage}
    \begin{minipage}[b]{0.32\linewidth}
    \centering
    \includegraphics[width=1.0\hsize]{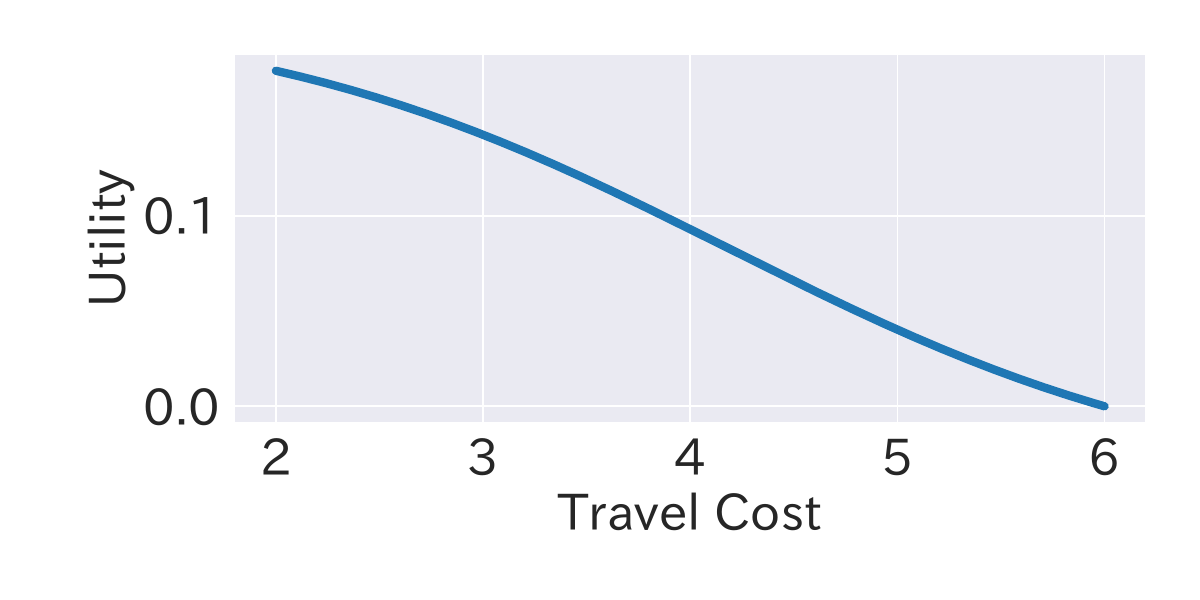}
    \subcaption{MNL-GAUNet: Travel Cost}\label{fig:synthetic-gaunet-bus-travelcost2}
    \end{minipage}
        \begin{minipage}[b]{0.32\linewidth}
    \centering
    \includegraphics[width=1.0\hsize]{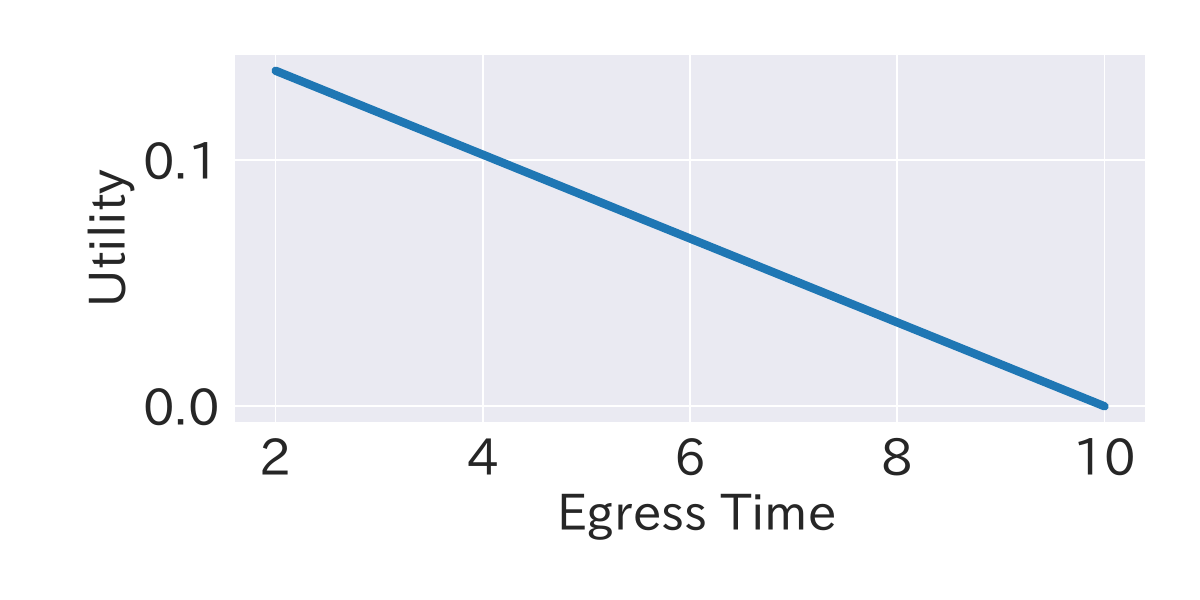}
    \subcaption{MNL-Linear: Egress Time}\label{fig:synthetic-mnllinear-bus-egresstime2}
    \end{minipage}
            \begin{minipage}[b]{0.32\linewidth}
    \centering
    \includegraphics[width=1.0\hsize]{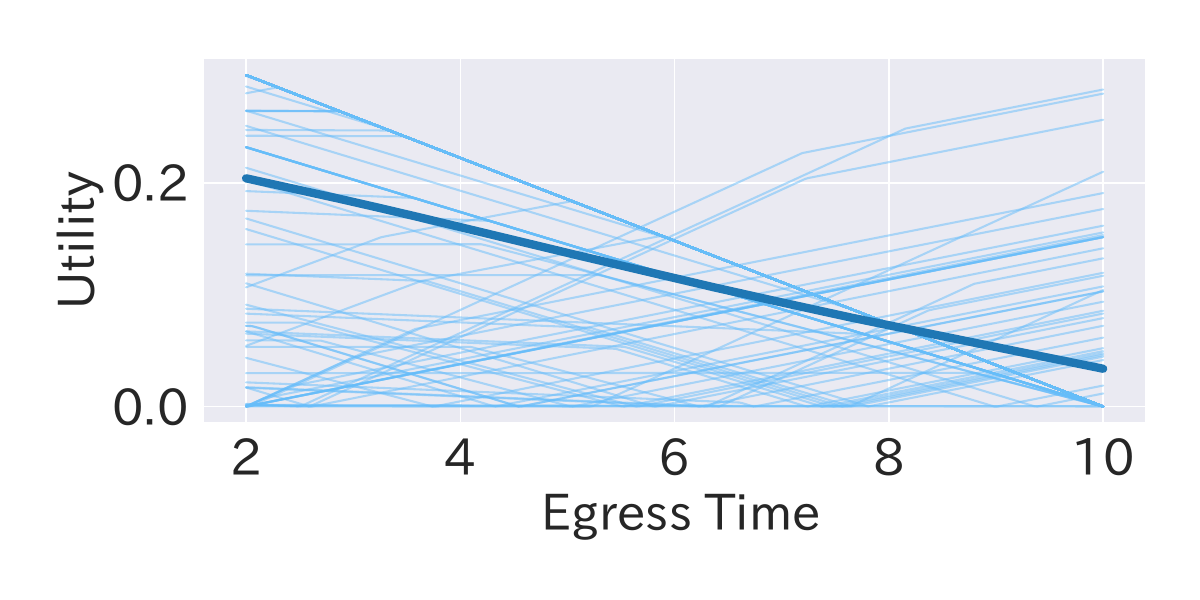}
    \subcaption{ASU-DNN: Egress Time}\label{fig:synthetic-asudnn-bus-egresstime2}
    \end{minipage}
                \begin{minipage}[b]{0.32\linewidth}
    \centering
    \includegraphics[width=1.0\hsize]{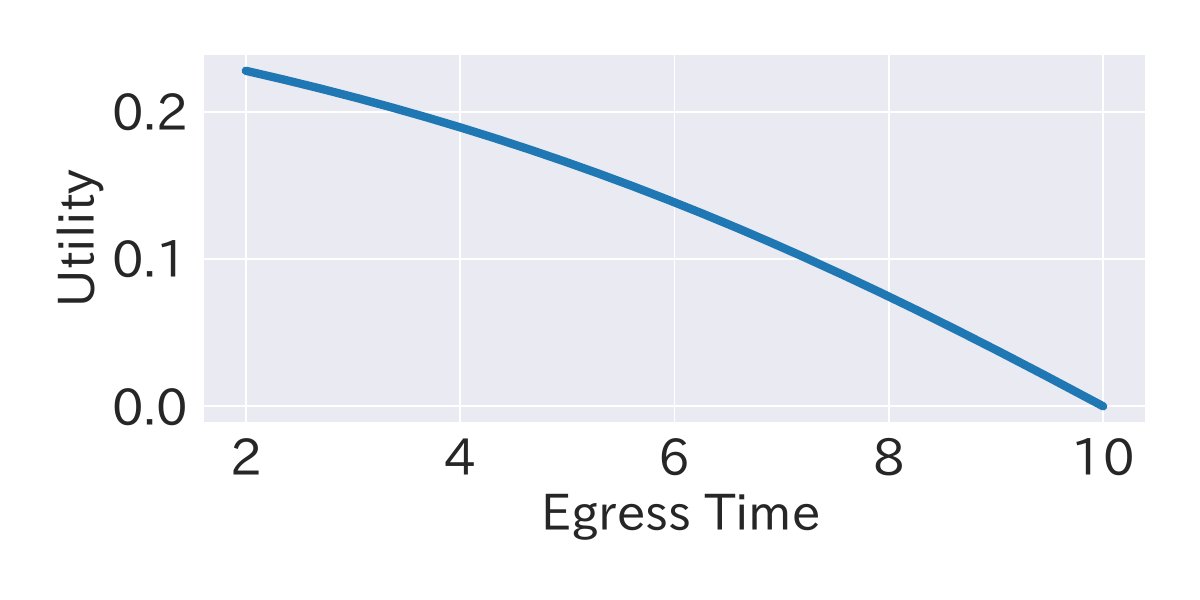}
    \subcaption{MNL-GAUNet: Egress Time}\label{fig:synthetic-gaunet-bus-egresstime2}
    \end{minipage}
    \begin{minipage}[b]{0.32\linewidth}
    \centering
    \includegraphics[width=1.0\hsize]{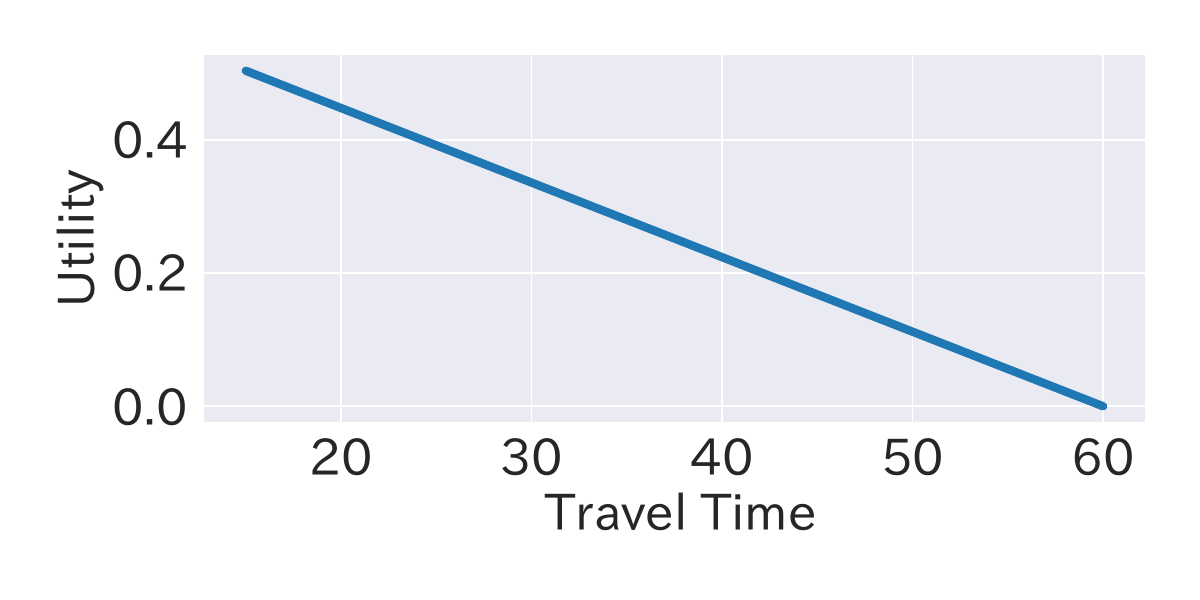}
    \subcaption{MNL-Linear: Travel Time}\label{fig:synthetic-mnllinear-bus-traveltime2}
    \end{minipage}
    \begin{minipage}[b]{0.32\linewidth}
    \centering
    \includegraphics[width=1.0\hsize]{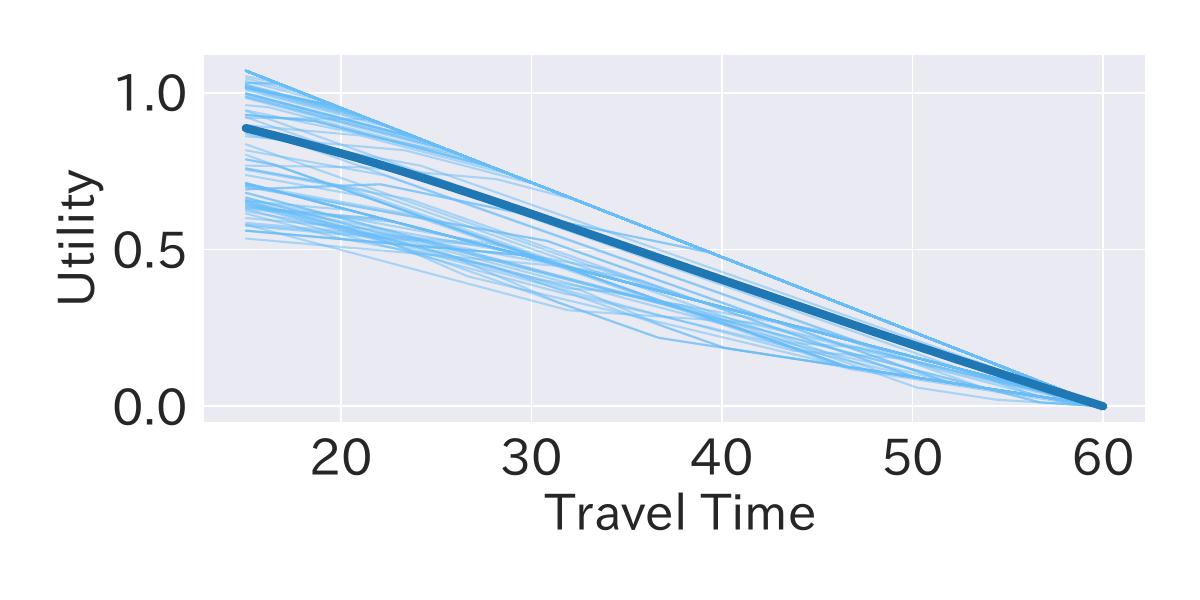}
    \subcaption{ASU-DNN: Travel Time}\label{fig:synthetic-asudnn-bus-traveltime2}
    \end{minipage}
    \begin{minipage}[b]{0.32\linewidth}
    \centering
    \includegraphics[width=1.0\hsize]{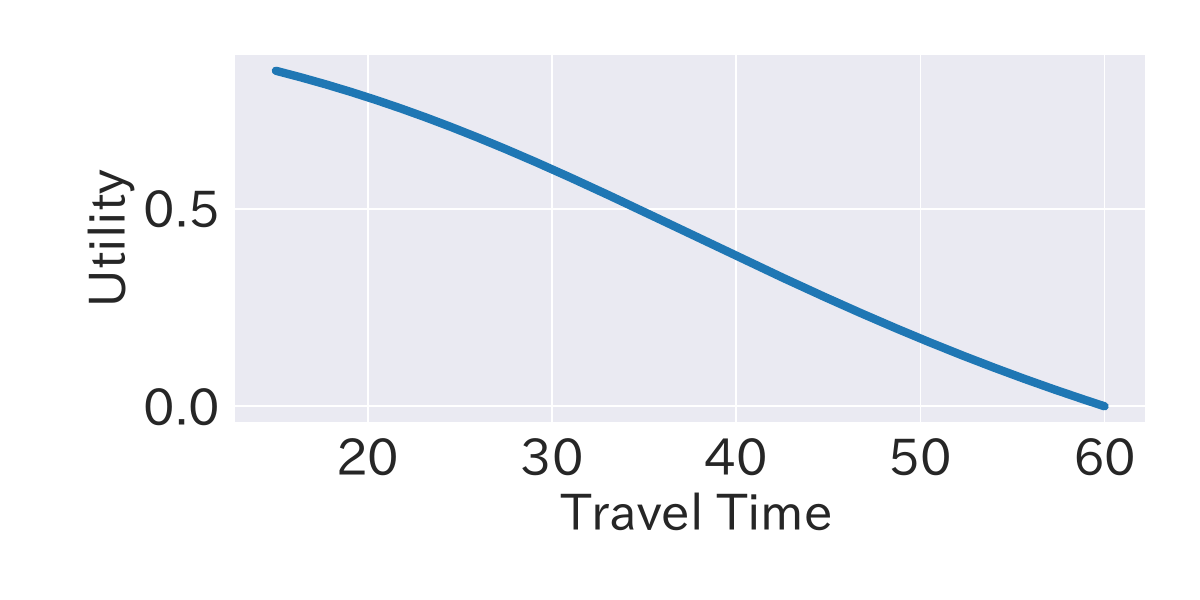}
    \subcaption{MNL-GAUNet: Travel Time}\label{fig:synthetic-gaunet-bus-traveltime2}
    \end{minipage}
        
    \caption{Estimated utility of the time-related explanatory variables (i.e., travel cost (a, b, c), egress time (d, e, f), and travel time (g, h, i) ) for buses via MNL-Linear (a, d, g), ASU-DNN (b, e, h), and MNL-GAUNet (c, f, i). The thin lines were computed with the values included in the dataset as the values of the other explanatory variables. The bold line is the median.}
    \label{fig:synthetic-other-utility-bus}
\end{figure}

We observed the other bus utilities learned by each method to deepen our understanding of their interpretability. We also show the estimated utility of the other variables for the buses with each method in Fig.~\ref{fig:synthetic-other-utility-bus}. We found that the utilities estimated by ASU-DNN can have significant variance depending on the values of other variables. In particular, for the bus egress time, although the utility should decrease monotonically with respect to the increase in the egress time, the utilities increased in some cases with respect to the increase in egress time. The results indicate that observing the prediction accuracy and/or the median of the utilities is not sufficient to understand the behavior of the utilities learned by ASU-DNN, making it difficult to interpret the model with ASU-DNN. On the other hand, because the utilities learned by MNL-GAUNet or MNL-Linear are independent of the other variables, it is easier to interpret the results and to determine whether the results are more reasonable than those of ASU-DNN.

\subsubsection{Policy evaluation}

\begin{figure}[t]
    \centering
    \begin{minipage}[b]{0.45\linewidth}
    \centering
    \includegraphics[width=1.0\hsize]{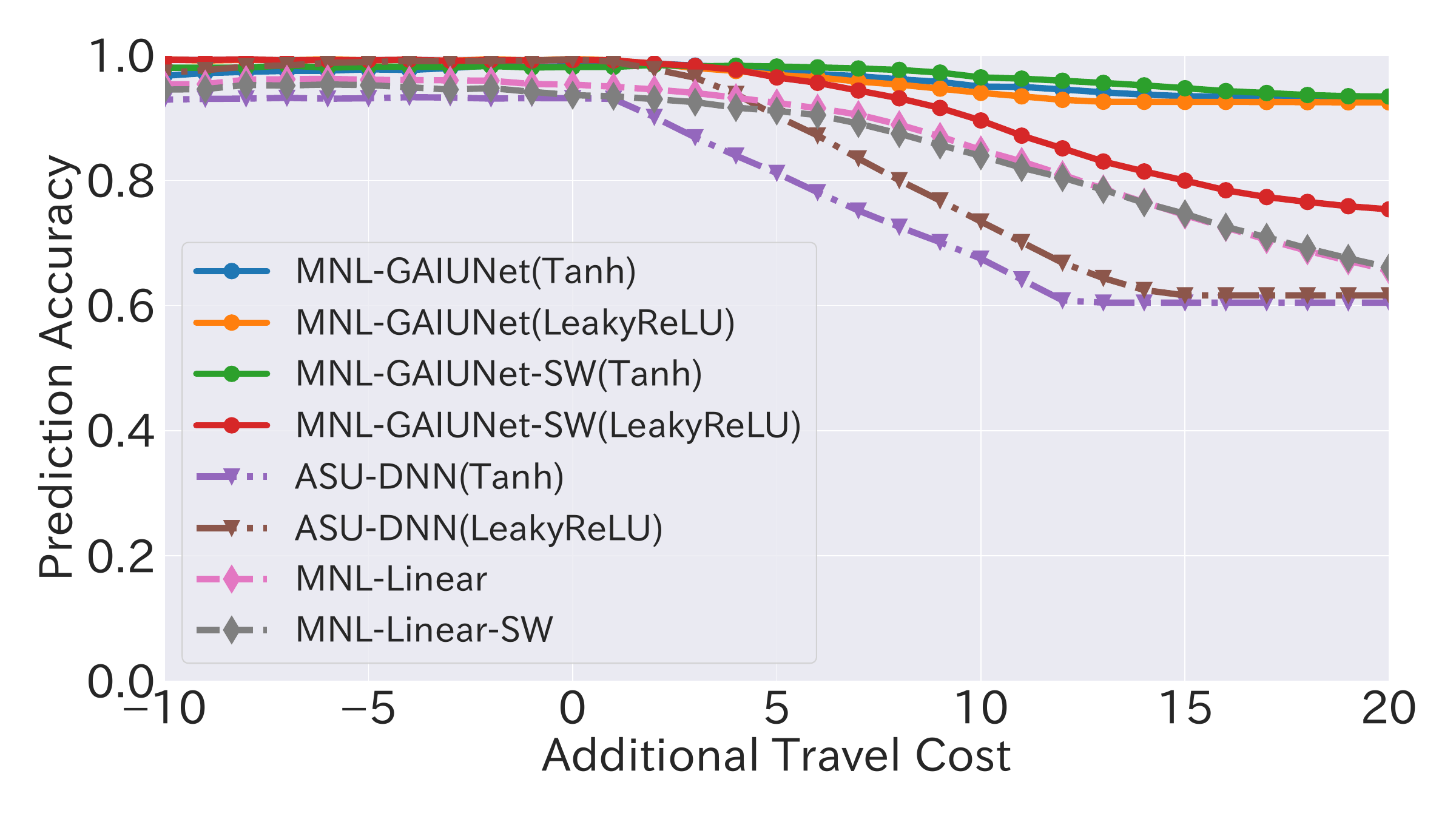}
    \subcaption{Additional Taxi Cost}\label{fig:additional_cost}
    \end{minipage}
    \begin{minipage}[b]{0.45\linewidth}
    \centering
    \includegraphics[width=1.0\hsize]{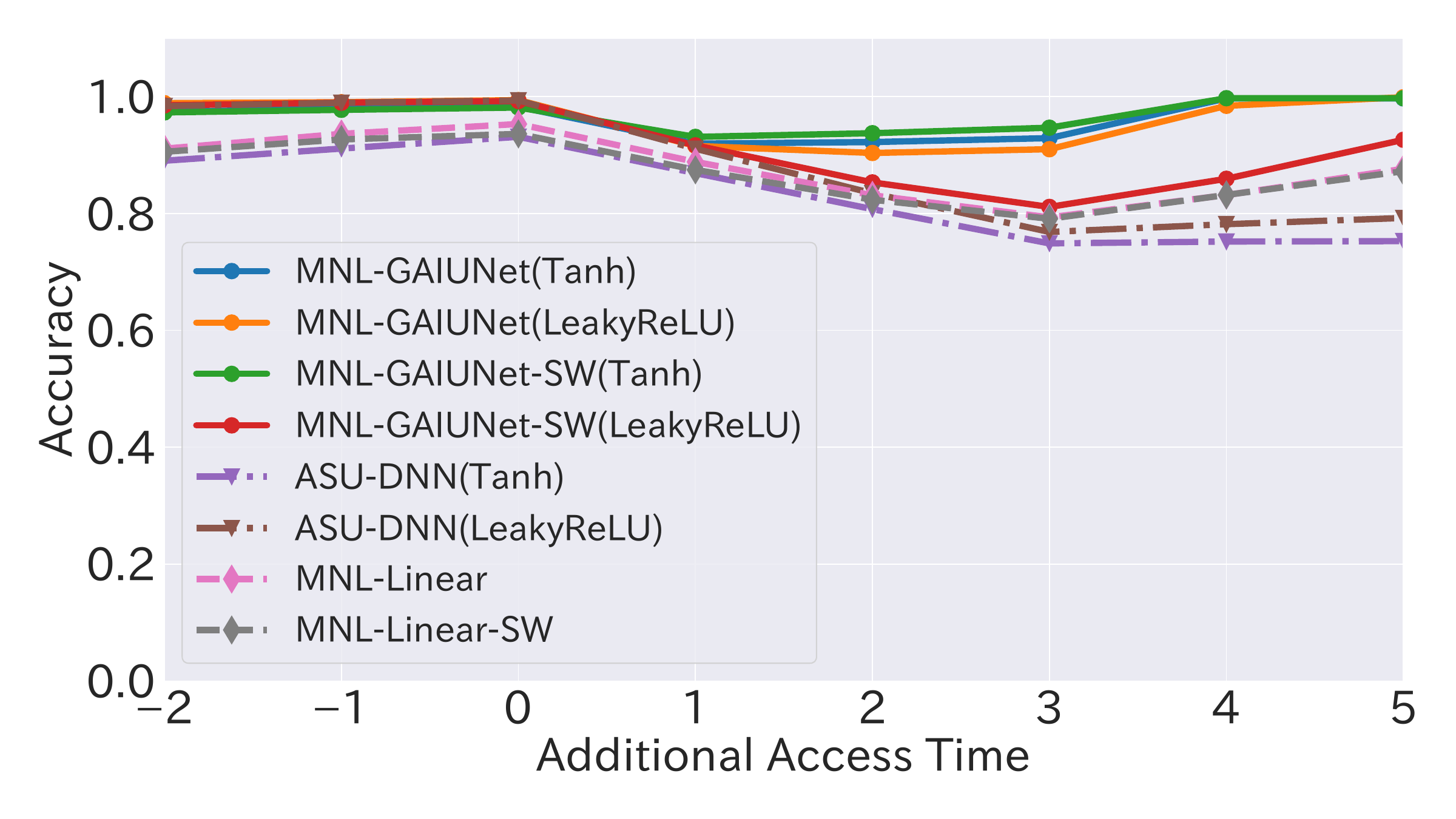}
    \subcaption{Additional Bus Access Time}\label{fig:additional_acc_time}
    \end{minipage}
    \caption{Policy evaluation performance with the additional taxi travel costs (a) and the additional bus access time (b). Circle $\circ$, inverted triangle $\triangledown$, and diamond $\diamond$ represent MNL-GAUNet, ASU-DNN, and MNL-Linear, respectively.}
    \label{fig:policy_eval}
\end{figure}

We also compared the policy evaluation performance of each method to answer the simple question of whether ASU-DNN and MNL-GAIUNet, which have comparable prediction accuracies, are comparable in terms of policy evaluation performance. We simulated two policies: changes in the taxi travel costs and changes in the bus access time. In the policy of changes in taxi travel costs, we measured accuracy when travel costs were increased by an additional -10 to 20 dollars. When the additional travel costs are positive, the buses are more likely to be selected because users select no taxis when the travel costs exceed 20 dollars because of their constraints. Figure~\ref{fig:additional_cost} shows the accuracy of each method. We found that the methods of ASU-DNN and MNL-Linear decreased the accuracy when the travel costs additionally increased. The results indicate that these models could not learn the mode choice considering the unobservable constraints from the data. On the other hand, our MNL-GAUNet based models, except for MNL-GAUNet-SW(LeakyReLU), maintain accuracy even if the travel cost is additionally increased. The results revealed that while the prediction accuracy of our models was comparable to that of the ASU-DNN-based models, our method could capture the constraints from the data.

We also measured the accuracy of each method if the access time of buses was increased by an additional -2 to 5 minutes. Because users select no buses when the access time exceeds four minutes due to their constraints, taxis are more likely to be selected when the additional access time is positive. Figure~\ref{fig:additional_acc_time} shows the accuracy of each method. The results exhibit almost the same trend as the results of the above simulation for additional travel taxi costs. We found that the methods of ASU-DNN and MNL-Linear worsened the accuracy when the additional access time of the buses was positive. The results indicate that these conventional models could not learn the mode choice considering the unobservable constraints from the data. On the other hand, our MNL-GAUNet based models, except for MNL-GAUNet-SW(LeakyReLU), maintained accuracy even if the additional access time was positive. Therefore, the results revealed that while the prediction accuracy of our models was comparable to that of the ASU-DNN based models, our method could capture the constraints from the data.

In summary, we have revealed that the prediction accuracy of the collected data, even when cross-validated, does not necessarily reflect the accuracy of the policy evaluation. Although the prediction accuracies of ASU-DNN (LeakyReLU) and MNL-GAUNet (Tanh) were comparable in terms of collected data, MNL-GAUNet (Tanh) outperformed ASU-DNN (LeakyReLU) in policy evaluation. One of the reasons could be the difference between the distribution of the data collected a priori and the distribution of the data obtained in the policy evaluation. That is, the results could be related to the fact that the estimation by ASU-DNN has a large variance owing to the interaction between the variables, as shown in Fig.~\ref{fig:synthetic-other-utility-bus}.

\section{Experiments with Trip Survey Data in Tokyo} \label{sec:experiment-toyosu}

\subsection{Experimental settings}
This section introduces the experimental setup and model parameters using trip survey data in Tokyo, Japan. 

\subsubsection{Data Summary}
We evaluated our models with trip survey data collected in Tokyo, Japan, for several months per year from 2018 through 2021. The subjects lived or worked in a waterfront district in Tokyo. The number of subjects was approximately 200 in 2018 and 300 in other years. The subjects recorded the travel destinations/objectives~(e.g., commuting to work/school) and transport modes~(e.g., trains) of the trips via their smartphones. The number of trips was 12,000 in 2018, 30,000 in 2019, and 35,000 in other years. The trip destinations/objectives were \textit{Home}, \textit{Office/School}, \textit{Club-Activity}, \textit{Business}, \textit{Meal}, \textit{Leisure}, \textit{Sightseeing}, \textit{Pickup}, \textit{Waiting}, and others. The transport modes were \textit{Train, Walking, Subway, Bicycle, Bus, Shared-Bike, Taxi, and Others~(\textit{Freight-Truck}, \textit{Monorail}, \textit{Motorcycle}, \textit{Tram}, and \textit{velotaxi})}. 

We focused on the following destinations/objectives for which decision-making behavior would differ:\textit{ Office/School, Meal, Leisure, Hospital, and Sightseeing}. We also focused on the following major transport modes as alternatives: \textit{Train}, \textit{Walking}, \textit{Subway}, \textit{POV~(Privately-Owned-Vehicle)}, \textit{POB~(Privately-Owned-Bike)}, \textit{Bus}, \textit{Shared-Bike}, and  \textit{Taxi}. We equated Train and Subway to \textit{Train}, POB and Shared-Bike to \textit{Bike}, and POV and Taxi to \textit{Car}.

\begin{figure}[ht]
    \centering
    \begin{minipage}[b]{0.45\linewidth}
    \centering
    \includegraphics[width=1.0\hsize]{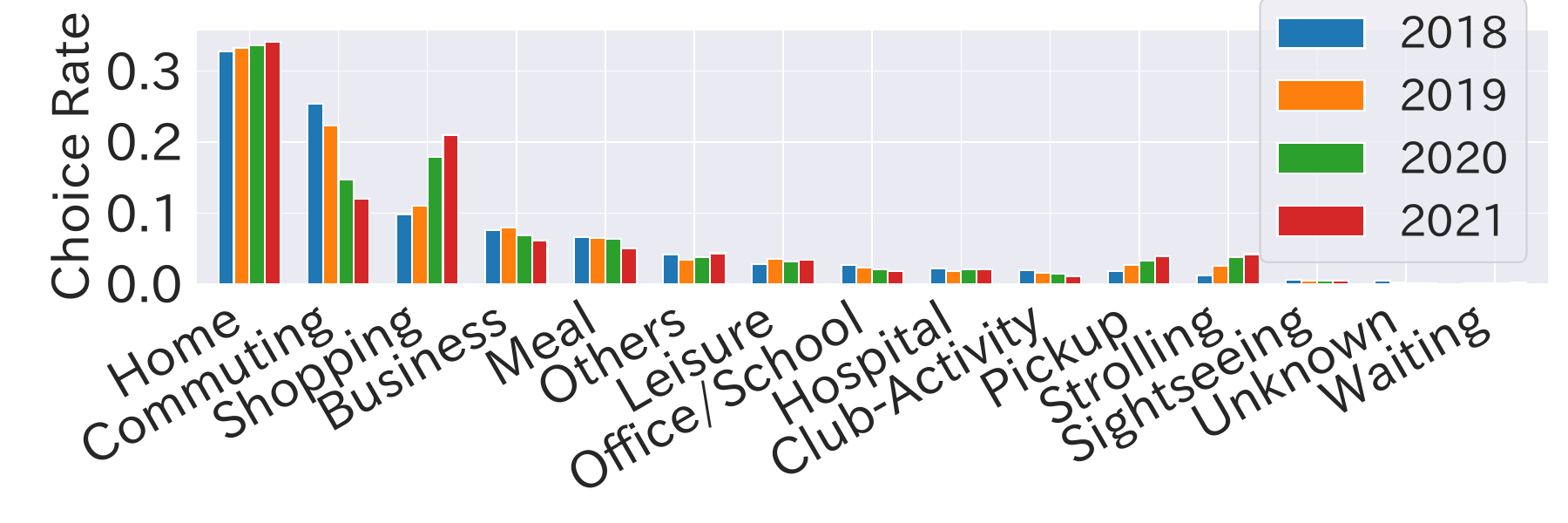}
    \subcaption{All Data}\label{fig:toyosu-objective_org}
    \end{minipage}
    \begin{minipage}[b]{0.45\linewidth}
    \centering
    \includegraphics[width=1.0\hsize]{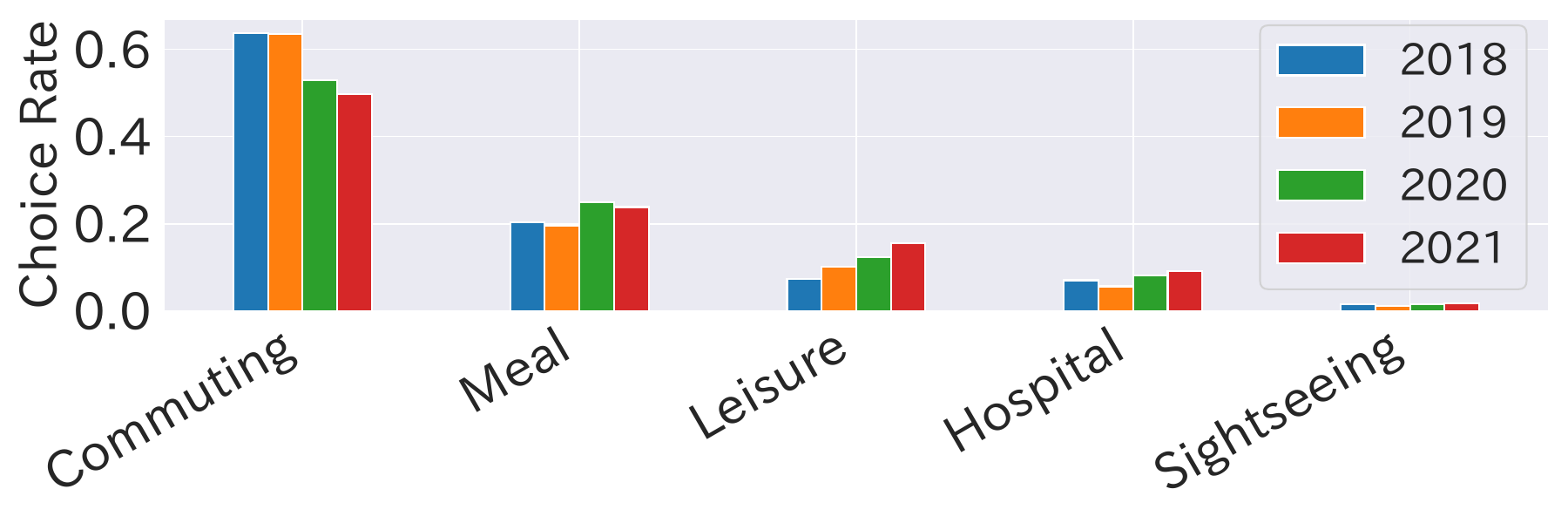}
    \subcaption{Selected Objective}\label{fig:toyosu-objective}
    \end{minipage}
    \caption{Histogram for travel objective in all data~\subref{fig:toyosu-objective_org} and in the data used for the evaluation~\subref{fig:toyosu-objective}}
    \label{fig:toyosu-stats-objective}
\end{figure}

\begin{figure}[ht]
    \centering
    \begin{minipage}[b]{0.45\linewidth}
    \centering
    \includegraphics[width=1.0\hsize]{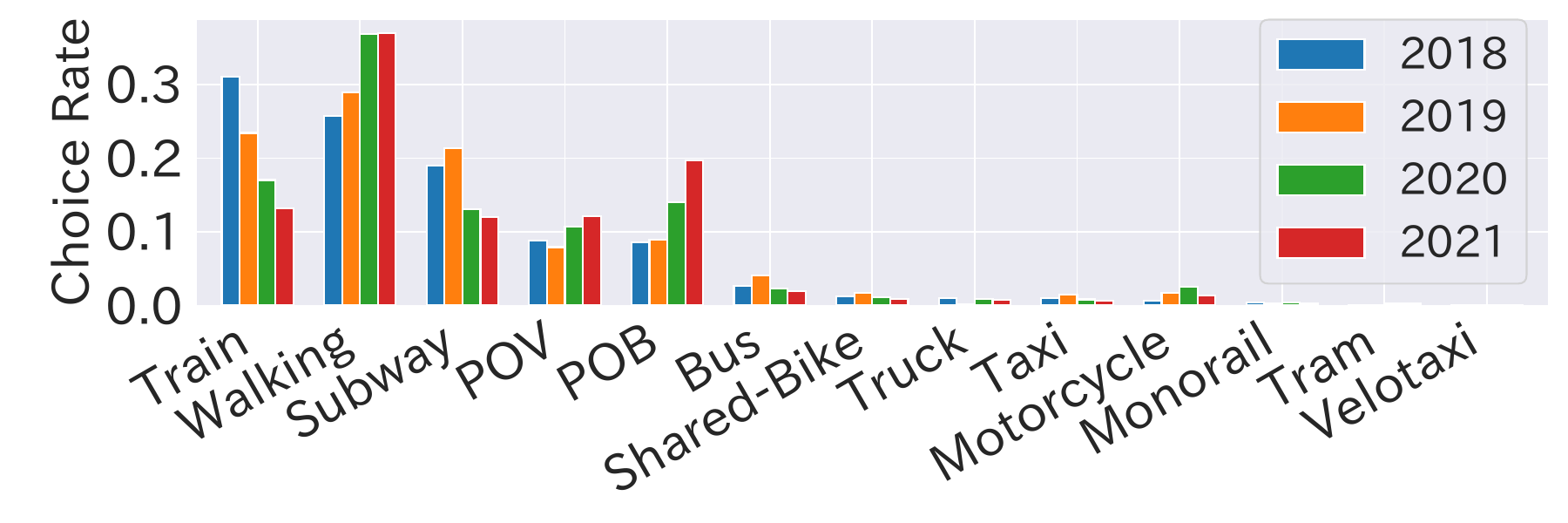}
    \subcaption{All Data}\label{fig:toyosu-mode_org}
    \end{minipage}
    \begin{minipage}[b]{0.45\linewidth}
    \centering
    \includegraphics[width=1.0\hsize]{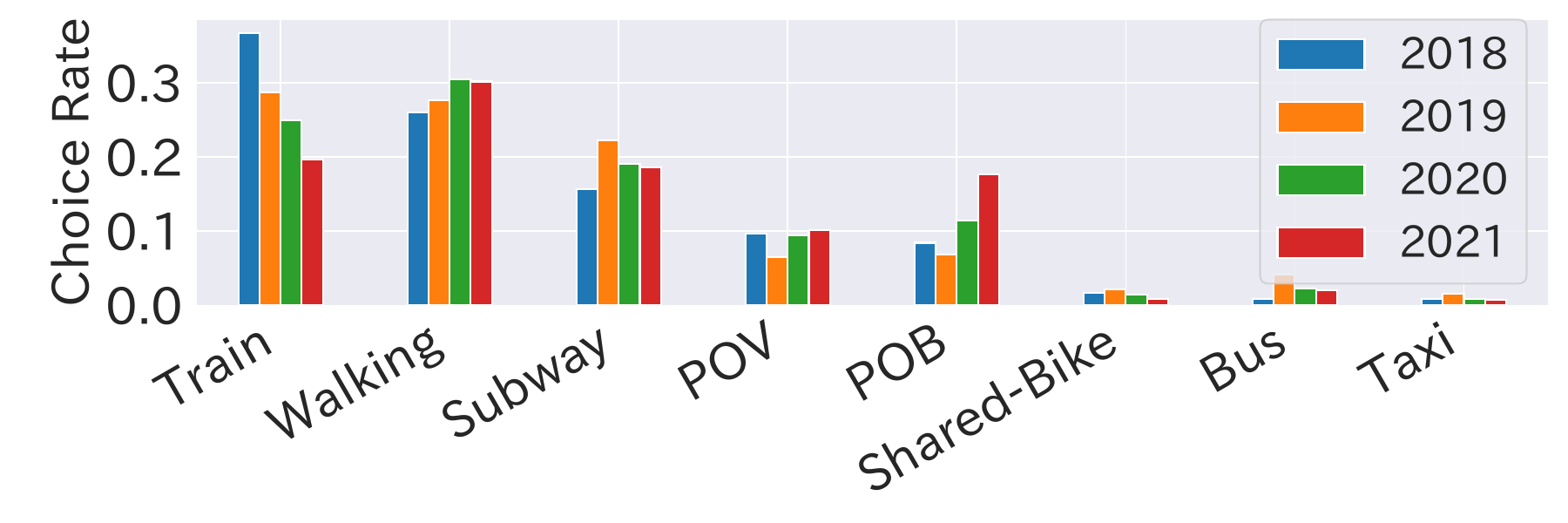}
    \subcaption{Selected travel modes}\label{fig:toyosu-mode}
    \end{minipage}
    \caption{Histogram for travel modes}
    \label{fig:toyosu-stats-mode}
\end{figure}

\begin{table}[ht]
    \centering
    \rowcolors{2}{}{gray!15}
    \small
    \caption{Overview of Trip Survey Data in Tokyo}
    \label{tab:toyosu-pp-data}
    \begin{tabular}{lcccc}
        \toprule
        \textbf{Year} & \textbf{2018} & \textbf{2019} & \textbf{2020} & \textbf{2021} \\
        \midrule
        Subjects    &  \multicolumn{4}{c}{people live or work at waterfront district in Tokyo}\\
        Destination/Purpose    &  \multicolumn{4}{p{12cm}}{\centering Home, \textbf{Office/School}, Shopping, Business, \textbf{Meal}, \textbf{Leisure}, \textbf{Hospital}, \textbf{Club-Activity}, Pickup, \textbf{Sightseeing}, Waiting, and Others}\\
        Transport Modes    &  \multicolumn{4}{p{12cm}}{\textbf{Train}, \textbf{Privately-Owned-Vehicles}, \textbf{Walking}, \textbf{Subway}, \textbf{Privately-Owned-Bike}, \textbf{Bus}, \textbf{Shared-Bike}, \textbf{Taxi}, Freight-truck, Monorail, Motorcycle, Tram, and Velotaxi}\\
        Period & Aug./27-Dec/31 & Jul./08-2020/Mar./13 & Sep./30-Dec./23 & Jul./08-Nov./30 \\
        \#Trips & 12321 & 30585 & 35924 & 38490 \\
        \#Trips$_{\mathrm selected}$ & 3383 & 8266 & 8005 & 7380 \\
        \#Subjects & 218 & 316 & 326 & 325 \\
        \#Subjects$_{\mathrm selected}$ & 174 & 297 & 309 & 312 \\
        \bottomrule
    \end{tabular}
\end{table}
Table~\ref{tab:toyosu-pp-data} shows the data overview in detail. We present a histogram showing the ratio of trip destinations/objectives recorded and those selected from the data in Fig.~\ref{fig:toyosu-stats-objective}. The travel objective shares were almost the same in all years except for some objectives. The commuting ratios in 2018 and 2019 were greater than those in 2020 and 2021, while the ratios of shopping/meal/leisure/pickup/strolling ratios in 2018 and 2019 were lower than those in 2020 and 2021. This change could have been caused by the COVID-19 pandemic.  Figure~\ref{fig:toyosu-stats-mode} shows the ratio of transport modes. The mode shares were also almost the same in all years except for some modes. The ratio of public transportation, such as trains, subways, and buses, decreased after 2020 whereas the ratio of walking, POV, and POB increased because of the pandemic. 

\begin{figure}[ht]
    \centering
    \begin{minipage}[b]{0.45\linewidth}
    \centering
    \includegraphics[width=1.0\hsize]{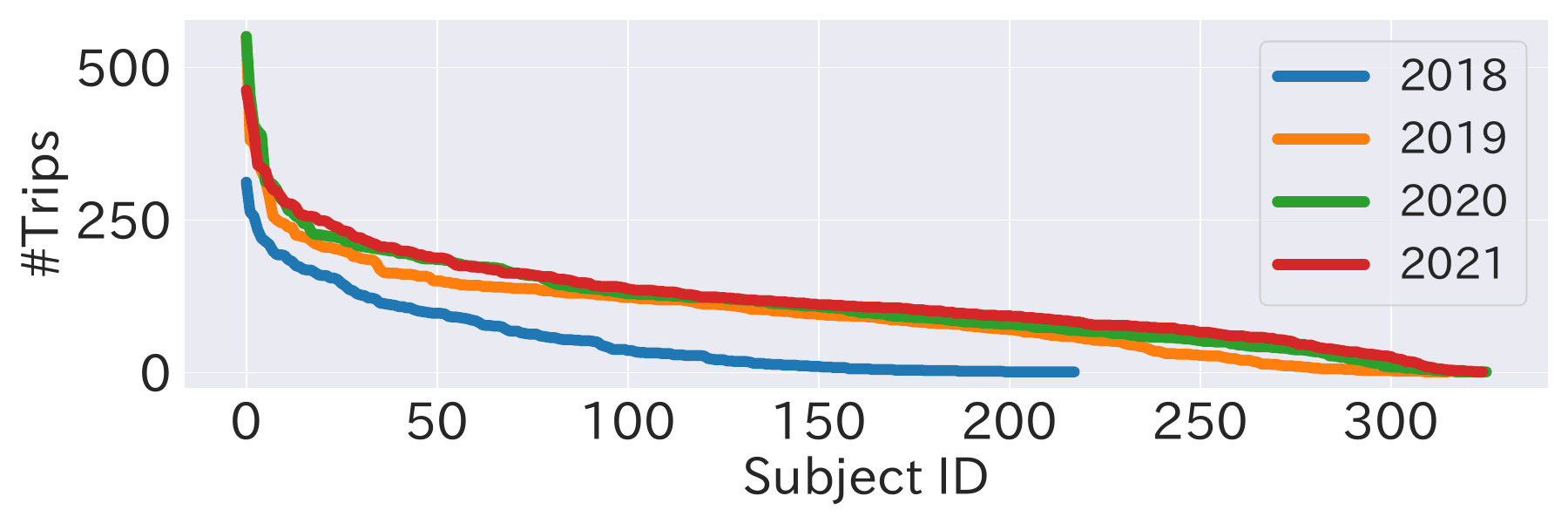}
    \subcaption{All Data}\label{fig:toyosu-num_trips_org}
    \end{minipage}
    \begin{minipage}[b]{0.45\linewidth}
    \centering
    \includegraphics[width=1.0\hsize]{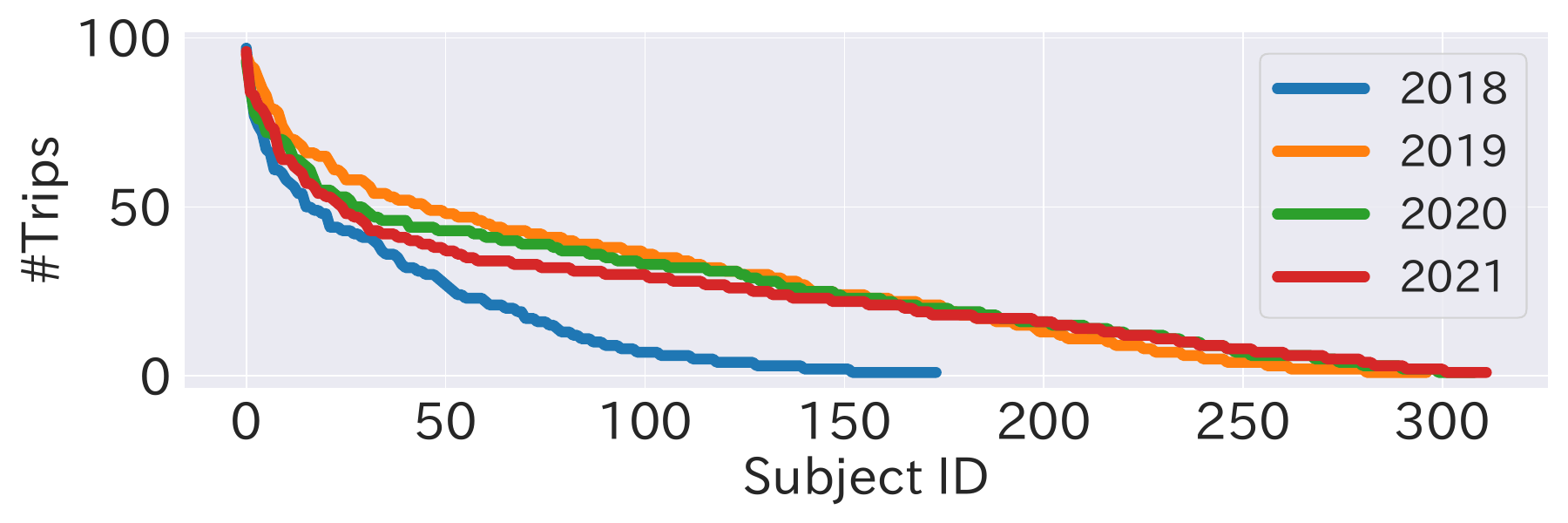}
    \subcaption{Selected Data}\label{fig:toyosu-num_trips}
    \end{minipage}
    \caption{Total number of trips in all data~\subref{fig:toyosu-num_trips_org} and in the recorded data for the evaluation~\subref{fig:toyosu-num_trips}}
    \label{fig:toyosu-stats-num_trips}
\end{figure}

Figure~\ref{fig:toyosu-stats-num_trips}  shows the number of trips recorded for each subject. The number did not change before and after the pandemic, although the number in 2018 was the lowest because it had the lowest number of subjects.  The number of trips was much greater for subjects below 10\% than for other subjects. 

\begin{figure}[t]
    \centering
    \begin{minipage}[b]{0.45\linewidth}
    \centering
    \includegraphics[width=1.0\hsize]{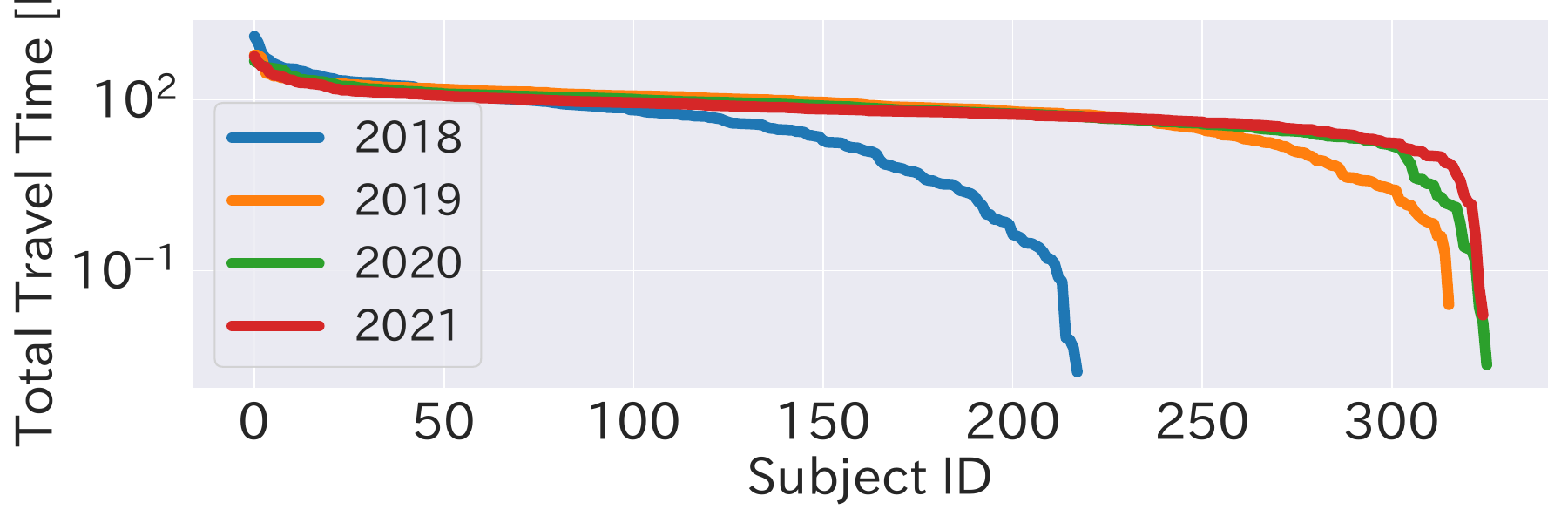}
    \subcaption{All Data}\label{fig:toyosu-hour_trips_org}
    \end{minipage}
    \begin{minipage}[b]{0.45\linewidth}
    \centering
    \includegraphics[width=1.0\hsize]{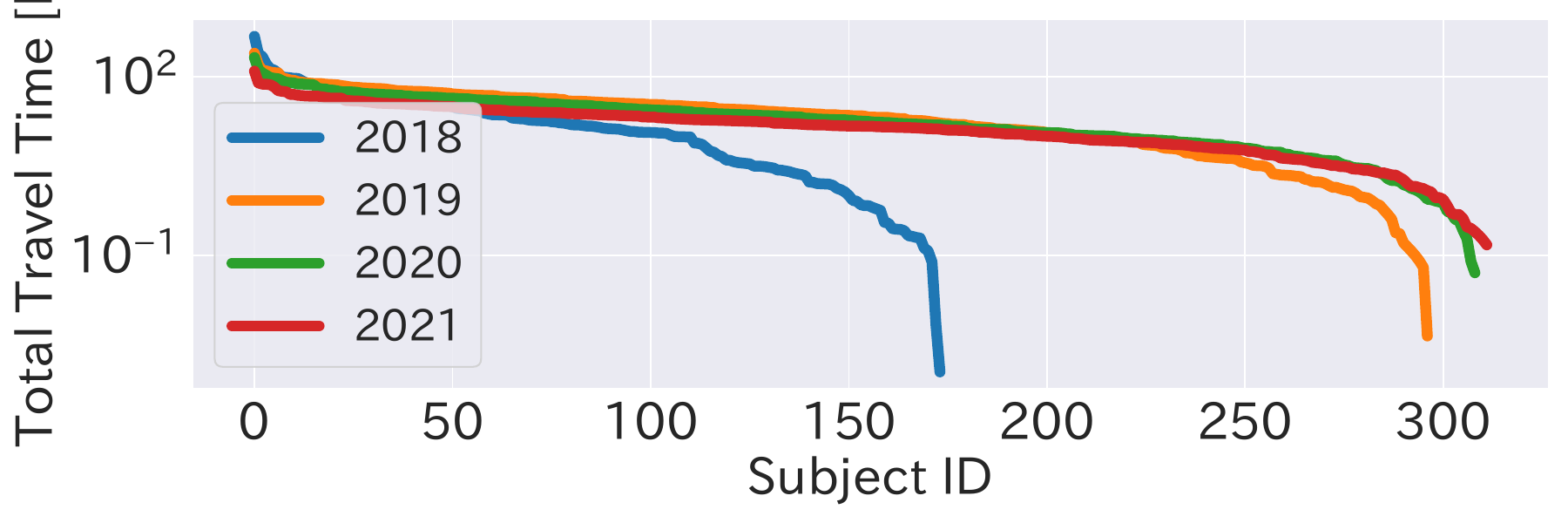}
    \subcaption{Selected Data}\label{fig:toyosu-hour_trips}
    \end{minipage}
    \caption{Total travel time for all data~\subref{fig:toyosu-hour_trips_org} and for the data recorded for the evaluation~\subref{fig:toyosu-hour_trips}}
    \label{fig:toyosu-stats-hour_trips}
\end{figure}

We show the total travel time of each subject in Fig.~\ref{fig:toyosu-stats-hour_trips}. The total travel time did not change much before and after the pandemic. A small fraction of the subjects had very short travel times compared with the other subjects.

\subsubsection{Explanatory variables}

In this section, we present the explanatory variables selected for the experiments. Table~\ref{tab:features} shows the list of variables recorded in the data. The travel destination/objective, time, and distance were recorded for all travel modes. The other variables, cost, number of transfers, access distance, access time, egress distance, and egress time, were recorded for the train and the bus only.

\begin{table}[t]
\centering
\rowcolors{2}{}{gray!15}
\small
\caption{Recorded variables and selected explanatory variables(\Square:recorded, \CheckedBox:selected, -:not recorded)}
\label{tab:features}

\begin{tabular}{lccccc}
\toprule
       & \textbf{Train} & \textbf{Bus} & \textbf{Car} & \textbf{Bike} & \textbf{Walking} \\ \midrule
Travel Destination/Objective     & \CheckedBox   & \CheckedBox   &  \CheckedBox   &  \CheckedBox   &  \CheckedBox  \\
Travel Time   &  \CheckedBox  &  \CheckedBox  &  \CheckedBox   &  \CheckedBox   &  \CheckedBox  \\
Travel Dist.     & \Square   &  \Square  &  \Square   &  \Square   &  \Square  \\
Cost     & \Square   &  \Square  &  -   &  -   &  -  \\
\#Transfers   &  \CheckedBox  &   \CheckedBox &   -  &   -  &  -  \\
Access Dist. &  \Square  &  \Square  &  -   &  -   &  -  \\
Access Time &  \CheckedBox  &  \CheckedBox  &  -   &  -   &   - \\
Egress Dist. &  \Square  &  \Square  &  -   &   -  &   - \\
Egress Time &  \CheckedBox  &  \CheckedBox  &   -  &  -   &  -  \\ \bottomrule
\end{tabular}
\end{table}

We calculate the variance inflation factor~(VIF), a measure that represents how much the variable can be estimated using the other variables for each explanatory variable candidate to analyze the magnitude of multicollinearity. The VIF is calculated as follows:
\begin{align}
    \tilde{x}_i &\coloneqq \alpha^i_0 + \sum_{j\neq i}\alpha^i_j x^i_j + \varepsilon^i, \forall i \in \{1,\dots,N\},\\
    R^2_i &\coloneqq 1 - \frac{\sum (x_i - \tilde{x}_i)^2}{\sum (x_i - \bar{x}_i)^2}, \forall i \in \{1,\dots,N\},\\
    \mathrm{VIF}_i &\coloneqq \frac{1}{1-R^2_i}, \forall i \in \{1,\dots,N\},
\end{align}
where $N$ is the number of explanatory variable candidates, and $i$ is its index. $\tilde{x}_i$ is the value estimated by linear regression using the other variables $\{1,\dots,N\}\backslash i$  and $\bar{x}$ is the mean value based on collected data $\mathcal{D}$. $R^2_i$ and $\mathrm{VIF}_i$ are the coefficient of determination and VIF for $x_i$, respectively. A high $\mathrm{VIF}_i$ indicates that the variable set has multicollinearity because the variable $x_i$ can be accurately estimated using other variables. We selected a combination of variables so that the VIFs would be less than 10. 

We selected Travel Destination/Objective, Travel Time, \#Transfers, Access Time, and Egress Time as the explanatory variables shown in Table~\ref{tab:features}. We confirmed that the VIFs for these variables were less than 10 as shown in Table~\ref{tab:vif}.

\begin{table}[t]
\centering
\rowcolors{2}{}{gray!15}
\small
\caption{Variance inflation factors for selected explanatory variables}
\label{tab:vif}
\begin{tabular}{lrrrrrrrr}
\toprule
 & \multicolumn{2}{c}{\textbf{2018}}& \multicolumn{2}{c}{\textbf{2019}} & \multicolumn{2}{c}{\textbf{2020}} &\multicolumn{2}{c}{\textbf{2021}}\\
 \cmidrule{2-9}
 & \textbf{Train} & \textbf{Bus} & \textbf{Train} & \textbf{Bus} & \textbf{Train} & \textbf{Bus} & \textbf{Train} & \textbf{Bus} \\
\midrule
Travel Time & 4.39 & 9.23 & 4.32 & 6.55 & 3.85 & 5.94 & 3.47 & 5.60 \\
\#Transfers & 2.78 & 8.81 & 2.57 & 6.43 & 2.27 & 5.28 & 2.02 & 4.84 \\
Access Time & 2.05 & 1.51 & 1.85 & 1.36 & 1.94 & 1.45 & 1.91 & 1.30 \\
Egress Time & 1.94 & 1.84 & 1.89 & 1.63 & 1.73 & 1.64 & 1.87 & 1.67 \\
\bottomrule
\end{tabular}
\end{table}

\subsubsection{Parameters}

We list the parameters for each model in Table~\ref{tab:models}. The number of layers and nodes was set to one for MNL-Linear; two layers and five nodes in each layer were set for the other models. We evaluated the performance with two different activation functions: hyperbolic tangent~(Tanh) and LeakyReLU~\cite{maas2013rectifier}. We set the regularization parameters, $\alpha$, $\alpha^I$ and $\beta^I$, to \num{1e-3}. The models with \textit{-SW} represent the NN weights for the travel time that are pooled across all modes~(i.e., Train, Bus, Car, Bike, and Walking). We set the batch size, and the learning rate for all models to 200 and \num{1e-3} respectively. For MNL-GAIUNet, we use 0.1 as $\mathrm{Th}_\mathrm{Imp}$.

\begin{table}[t]
    \centering
    \rowcolors{2}{}{gray!15}
    \small
    \caption{Parameter settings in each model: MNL-Linear, MNL-Linear-SW, ASU-DNN with Tanh/LeakyReLU, MNL-GAIUNet with Tanh/LeakyReLU, and MNL-GAIUNet-SW with Tanh/LeakyReLU }\label{tab:models}
    \begin{tabular}{lcccc}
    \toprule
     & \textbf{\#Nodes} & \textbf{Activation Func.} & \textbf{Regularization Param.} &  \textbf{Param. Sharing}  \\
    \midrule
    MNL-Linear & 1 & - & - & -\\
    MNL-Linear-SW & 1 & - & - & Travel Time[T,B,C,Bk,W]$^1$\\
    \midrule
    ASU-DNN(Tanh) & [5,5] & Tanh & - & -\\
    ASU-DNN(LeakyReLU) & [5,5] & LeakyReLU & - & -\\
    \midrule
    MNL-GAUNet(Tanh) & [5,5] & Tanh & -\num{1e-3} & -\\
    MNL-GAUNet(LeakyReLU) & [5,5] & LeakyReLU & -\num{1e-3} & -\\
    MNL-GAUNet-SW(Tanh) & [5,5] & Tanh & -\num{1e-3}  & Travel Time[T,B,C,Bk,W]\\
    MNL-GAUNet-SW(LeakyReLU) & [5,5] & LeakyReLU & -\num{1e-3} & Travel Time[T,B,C,Bk,W]\\
    \midrule
    MNL-GAIUNet(Tanh) & [5,5] & Tanh & -\num{1e-3} & -\\
    MNL-GAIUNet(LeakyReLU) & [5,5] & LeakyReLU & -\num{1e-3} & -\\
    MNL-GAIUNet-SW(Tanh) & [5,5] & Tanh & -\num{1e-3}  & Travel Time[T,B,C,Bk,W]\\
    MNL-GAIUNet-SW(LeakyReLU) & [5,5] & LeakyReLU & -\num{1e-3} & Travel Time[T,B,C,Bk,W]\\
    \bottomrule
    \end{tabular}
     \footnotesize{\rightline{$^1$ T:Train, B:Bus, C:Car, Bk:Bike, W:Walking}}
\end{table}
We evaluated the performance using five-fold cross-validation. In this validation method, the dataset was randomly partitioned into five equal (or nearly equal) folds. The models were trained on four folds and tested on one remaining fold. The process was performed five times, each time using a different fold. Performance was measured as the average of the log-likelihoods over the five evaluations.

\subsection{Experimental Results}\label{subsec:experiment-toyosu-result}

In this section, we evaluate our method in two aspects: prediction accuracy, measured by log-likelihood, and interpretability.

\subsubsection{Prediction Accuracy}
We show the log-likelihood for the models without shared network weights in Table~\ref{tab:log-likelihood}. We observed that all models were learned properly because the log-likelihood of the models was much smaller than the initial log-likelihood and the difference in log-likelihood for each model between \textit{test} and \textit{train} was small. 

We compared the performance of Tanh and LeakyReLU as activation functions in ASU-DNN, MNL-GAUNet, and MNL-GAIUNet. The results showed that Tanh was better than LeakyReLU in MNL-GAUNet and MNL-GAIUNet, while LeakyReLU was better than Tanh in ASU-DNN. This could be because the utility function approximated by ASU-DNN, which has connections between explanatory variables, was more complex than those of MNL-GAUNet and MNL-GAIUNet, which have no or limited connections between explanatory variables. The result was consistent with those of previous studies in which the ReLU family~(e.g., ReLU and LeakyReLU) achieved good performance in the approximation of complex functions~(e.g., deep and large network) while classical activation functions work well in learning simple functions~(e.g., shallow and small networks)~\cite{glorot2011deep}.  

ASU-DNN with LeakyReLU, MNL-GAUNet with Tanh, and MNL-GAIUNet with Tanh achieved better log-likelihood than MNL-Linear. The results indicated that choice policy had a nonlinear relationship with respect to the explanatory variables. MNL-GAUNet was comparable to ASU-DNN and MNL-GAIUNet. The results indicated that the interaction effect between the explanatory variables had little impact on mode choices in the experiments.

\begin{table}[t]
    \centering
    \rowcolors{2}{}{gray!15}
    \small
    \caption{Log-likelihood for the model without shared weights }\label{tab:log-likelihood}
    \begin{tabular}{lrrrrrrrr}
    \toprule
    & \multicolumn{2}{c}{\textbf{2018}}& \multicolumn{2}{c}{\textbf{2019}} & \multicolumn{2}{c}{\textbf{2020}} &\multicolumn{2}{c}{\textbf{2021}}\\
 \cmidrule{2-9}
     & \textbf{train} & \textbf{test} & \textbf{train} & \textbf{test} & \textbf{train} & \textbf{test} & \textbf{train} & \textbf{test} \\
    \midrule
    Initial LL & -1013 & -1013 & -2492 & -2492 & -2397 & -2397 & -2150 & -2150 \\
    \midrule
    MNL-Linear & \textbf{-447} & \textbf{-457} & \textbf{-1010} & \textbf{-1028} & \textbf{-1159} & \textbf{-1168} & \textbf{-1075} & \textbf{-1084} \\
    \midrule
    ASU-DNN(Tanh) & -525 & -532 & -1195 & -1203 & -1315 & -1325 & -1267 & -1277 \\
    ASU-DNN(LeakyReLU) & \textbf{-395} & \textbf{-413} & \textbf{-899} & \textbf{-914} & \textbf{-1066} & \textbf{-1089} & \textbf{-1004} & \textbf{-1027} \\
    \midrule
    MNL-GAUNet(Tanh) & \textbf{-418} & \textbf{-427} & \textbf{-906} & \textbf{-913} & \textbf{-1095} & \textbf{-1106} & \textbf{-1021} & \textbf{-1030} \\
    MNL-GAUNet(LeakyReLU) & -442 & -454 & -1007 & -1017 & -1190 & -1205 & -1157 & -1169 \\
    \midrule
    MNL-GAIUNet(Tanh) & \textbf{-415} & \textbf{-423} & \textbf{-902} & \textbf{-909} & \textbf{-1094} & \textbf{-1104} & \textbf{-1020} & \textbf{-1031} \\
    MNL-GAIUNet(LeakyReLU) & -439 & -449 & -995 & -990 & -1175 & -1178 & -1156 & -1168 \\
    \bottomrule
    \end{tabular}
\end{table}

We show the log-likelihood for the model with the shared network weights of the travel time in Table~\ref{tab:log-likelihood-sw}. The results with ASU-DNN were not included because ASU-DNN cannot pool weights across alternatives owing to the neural-network architectures. We observed that all models were learned properly because the log-likelihood of the models was much smaller than the initial log-likelihood, and the difference in log-likelihood for each model between \textit{test} and \textit{train} was small. In the shared weights scenarios, Tanh achieved better log-likelihood than LeakyReLU as the activation function of our models in the test. 

The log-likelihood for our models with shared weights was inferior to that of models with independent weights but was comparable to MNL-Linear-SW. These findings suggest that while the universal effect of travel time can be captured by linear functions, mode-specific effects may necessitate nonlinear functions.

\begin{table}[t]
    \centering
    \rowcolors{2}{}{gray!15}
    \small
    \caption{Log-Likelihood for the model with the shared weights}\label{tab:log-likelihood-sw}
    \begin{tabular}{lrrrrrrrr}
    \toprule
    & \multicolumn{2}{c}{\textbf{2018}}& \multicolumn{2}{c}{\textbf{2019}} & \multicolumn{2}{c}{\textbf{2020}} &\multicolumn{2}{c}{\textbf{2021}}\\
 \cmidrule{2-9}
     & \textbf{train} & \textbf{test} & \textbf{train} & \textbf{test} & \textbf{train} & \textbf{test} & \textbf{train} & \textbf{test} \\
    \midrule
    Initial LL                                         & -1013 & -1013 & -2492 & -2492 & -2397 & -2397 & -2150 & -2150 \\
    \midrule
    MNL-Linear-SW                            & -452 & -463 & -1023 & -1039 & -1170 & -1178 & -1104 & -1112 \\
    \midrule
    ASU-DNN(Tanh) & - & - & - & - & - & - & - & - \\
    ASU-DNN(LeakyReLU) & - & - & - & - & - & - & - & - \\
    \midrule
    MNL-GAUNet-SW(Tanh)             & -443 & -450 & -978 & -986 & -1174 & -1187 & -1112 & -1120 \\
    MNL-GAUNet-SW(LeakyReLU) & -435 & -445 & -966 & -977 & -1189 & -1207 & -1121 & -1133 \\
    \midrule
    MNL-GAIUNet-SW(Tanh)             & -441 & -448 & -928 & -923 & -1112 & -1108 & -1053 & -1038 \\
    MNL-GAIUNet-SW(LeakyReLU) & -435 & -445 & -921 & -915 & -1178 & -1200 & -1109 & -1115 \\
    \bottomrule
    \end{tabular}
\end{table}

\subsubsection{Interpretability}

\begin{figure}[t]
    \centering
    \begin{minipage}[b]{0.32\linewidth}
    \centering
    \includegraphics[width=1.0\hsize]{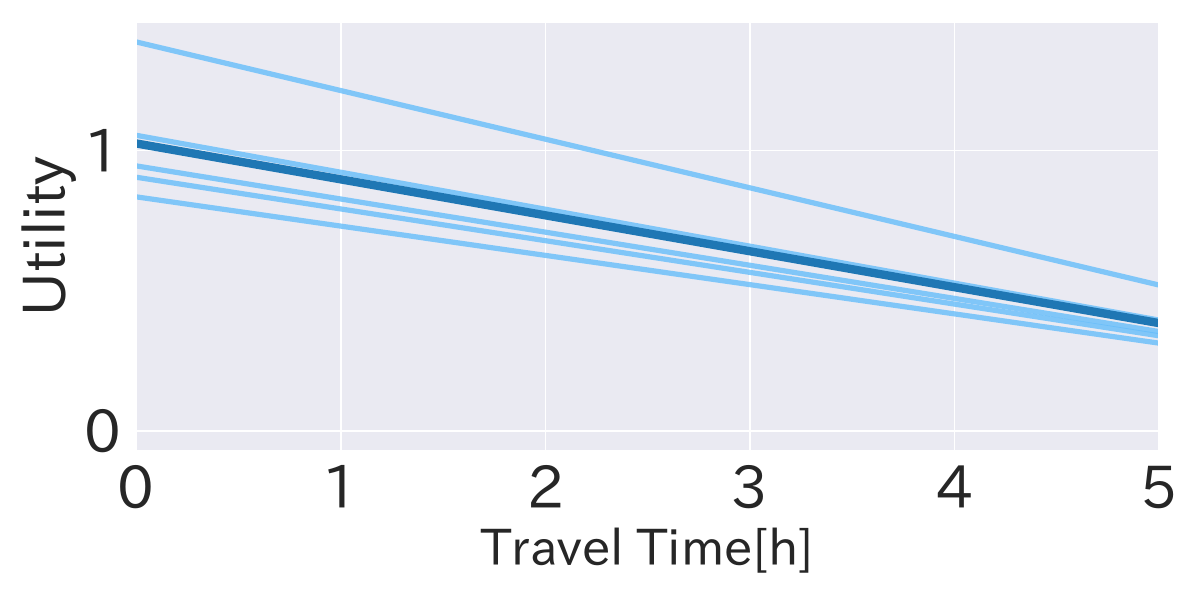}
    \subcaption{MNL-Linear: Travel Time}\label{fig:toyosu2020-mnl-linear-bus-travel_time}
    \end{minipage}
    \begin{minipage}[b]{0.32\linewidth}
    \centering
    \includegraphics[width=1.0\hsize]{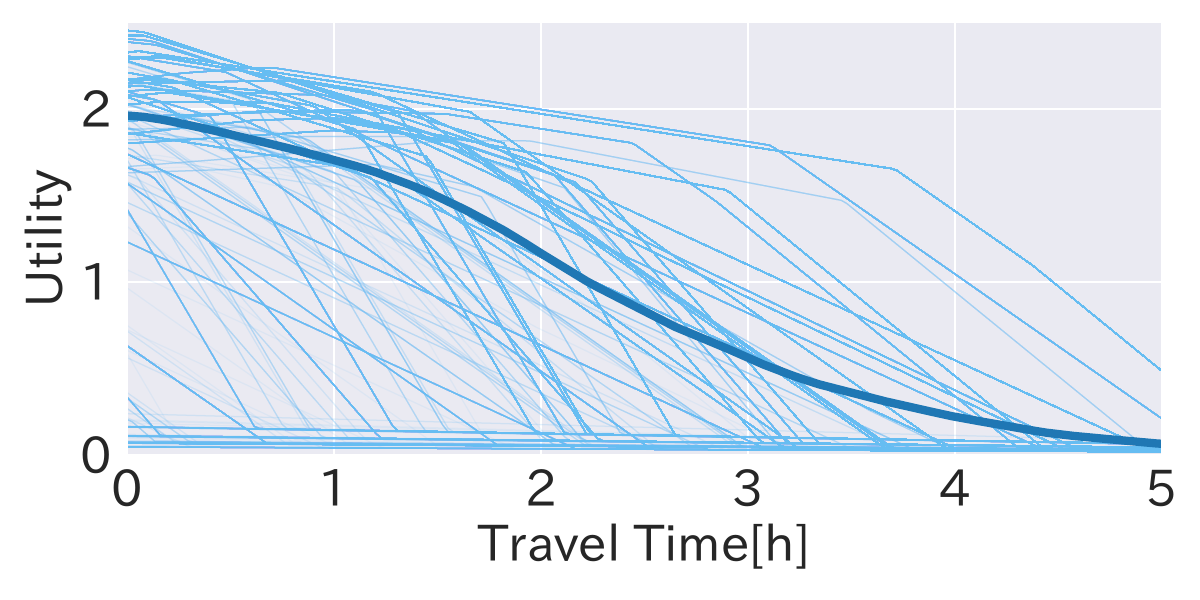}
    \subcaption{ASU-DNN: Travel Time}\label{fig:toyosu2020-asudnn-bus-travel_time}
    \end{minipage}
    \begin{minipage}[b]{0.32\linewidth}
    \centering
    \includegraphics[width=1.0\hsize]{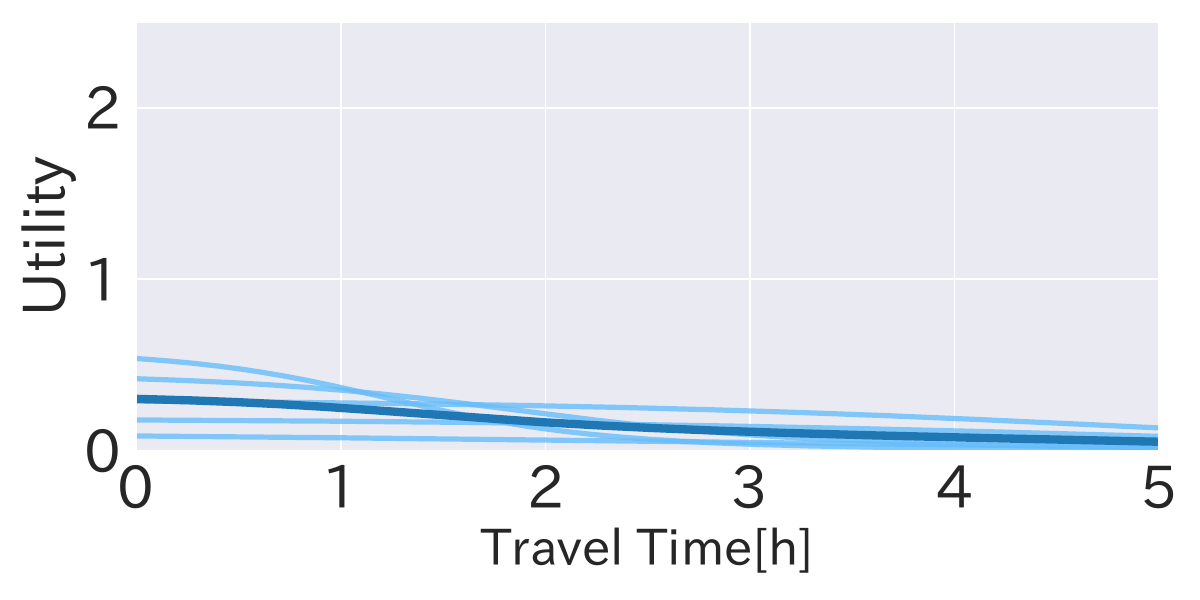}
    \subcaption{MNL-GAIUNet: Travel Time}\label{fig:toyosu2020-GAUNet-bus-travel_time}
    \end{minipage}
        \begin{minipage}[b]{0.32\linewidth}
    \centering
    \includegraphics[width=1.0\hsize]{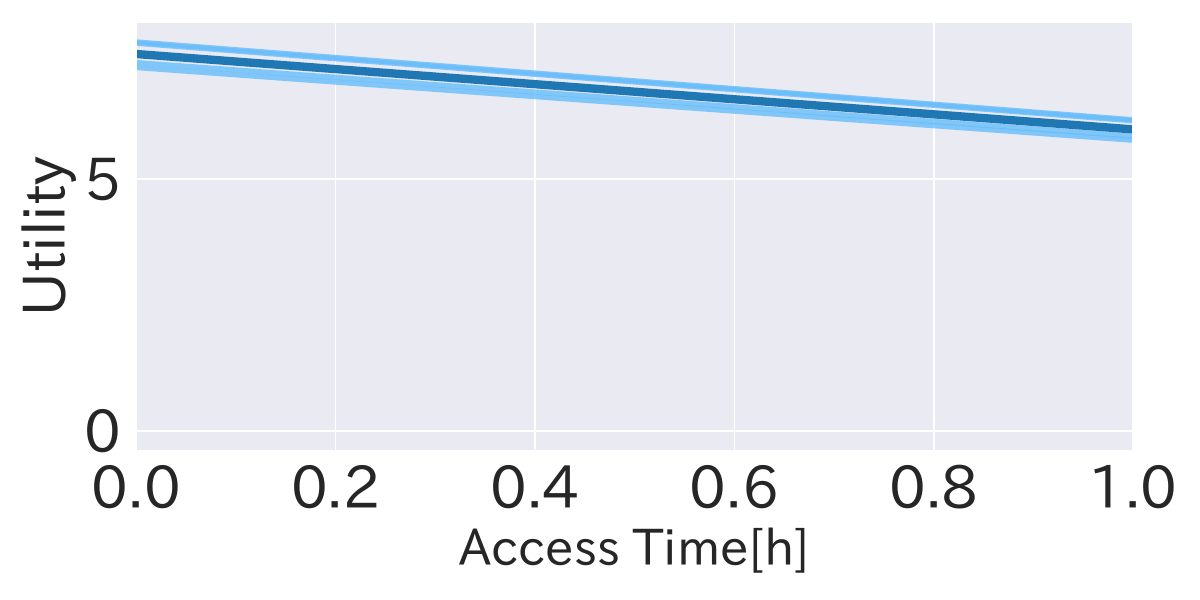}
    \subcaption{MNL-Linear: Access Time}\label{fig:toyosu2020-mnl-linear-bus-access_time}
    \end{minipage}
    \begin{minipage}[b]{0.32\linewidth}
    \centering
    \includegraphics[width=1.0\hsize]{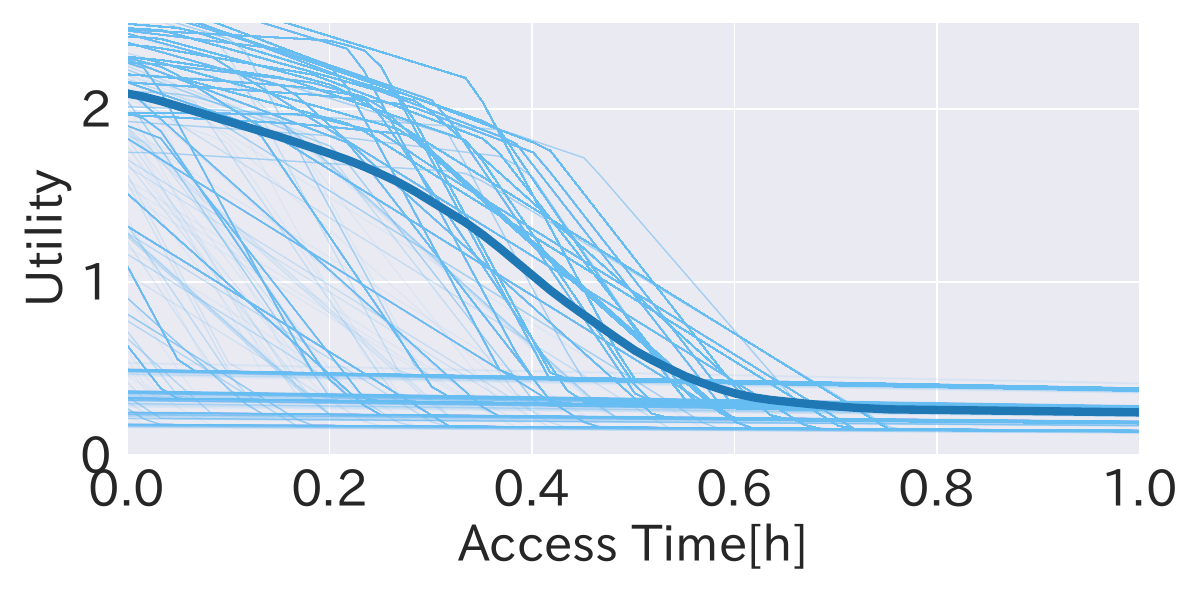}
    \subcaption{ASU-DNN: Access Time}\label{fig:toyosu2020-asudnn-bus-access_time}
    \end{minipage}
    \begin{minipage}[b]{0.32\linewidth}
    \centering
    \includegraphics[width=1.0\hsize]{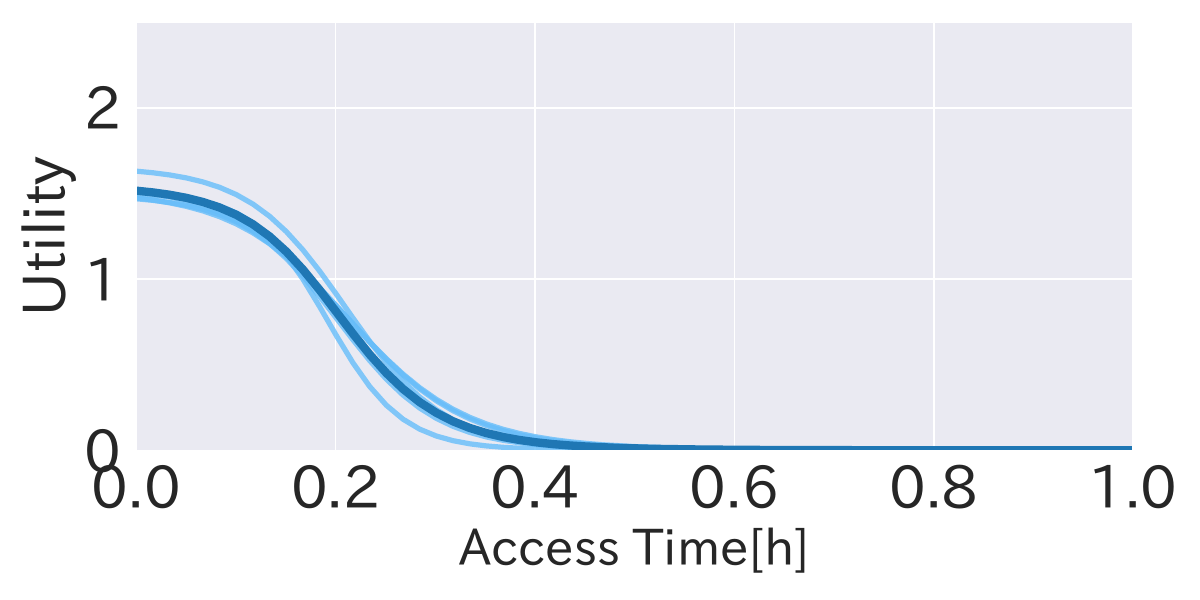}
    \subcaption{MNL-GAIUNet: Access Time}\label{fig:toyosu2020-GAUNet-bus-access_time}
    \end{minipage}
        \begin{minipage}[b]{0.32\linewidth}
    \centering
    \includegraphics[width=1.0\hsize]{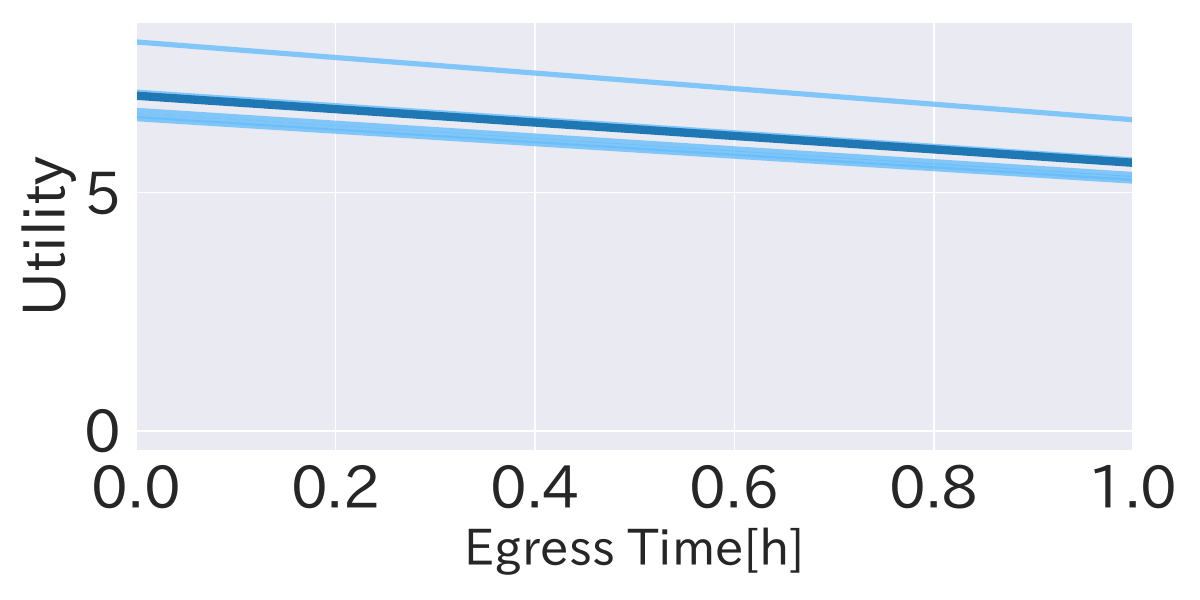}
    \subcaption{MNL-Linear: Egress Time}\label{fig:toyosu2020-MNL-Linear-bus-egress_time}
    \end{minipage}
    \begin{minipage}[b]{0.32\linewidth}
    \centering
    \includegraphics[width=1.0\hsize]{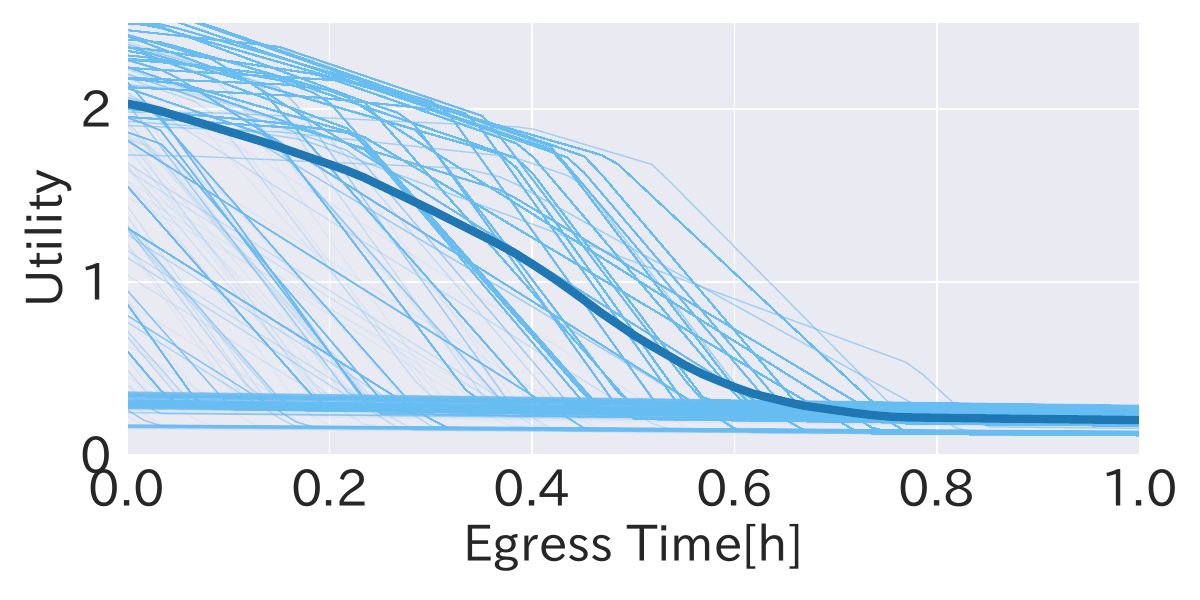}
    \subcaption{ASU-DNN: Egress Time}\label{fig:toyosu2020-asudnn-bus-egress_time}
    \end{minipage}
    \begin{minipage}[b]{0.32\linewidth}
    \centering
    \includegraphics[width=1.0\hsize]{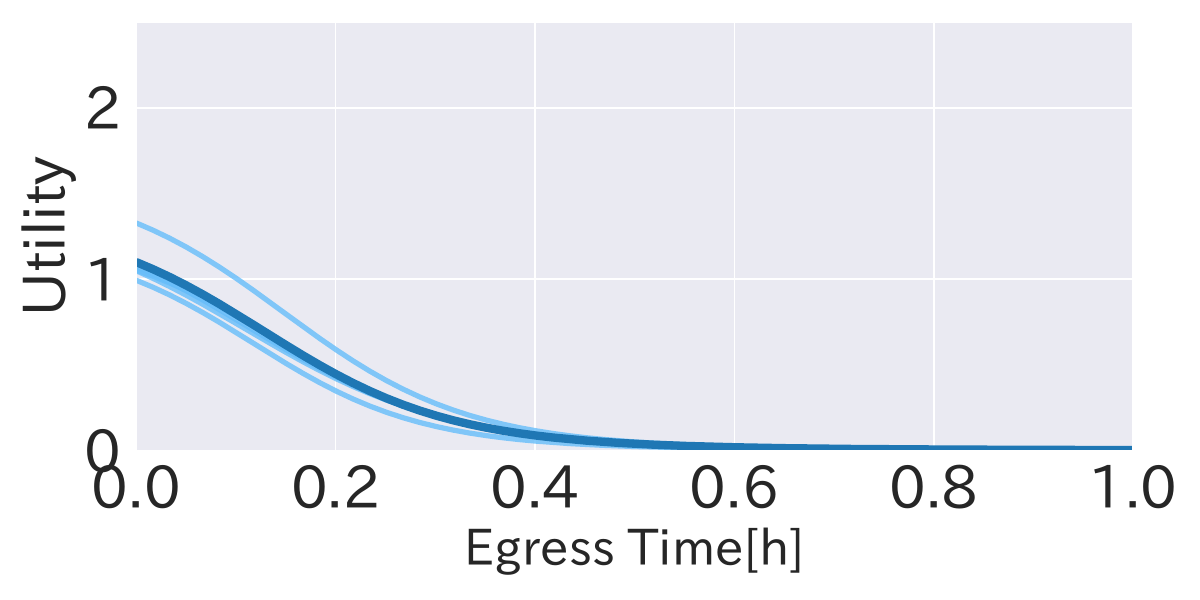}
    \subcaption{MNL-GAIUNet: Egress Time}\label{fig:toyosu2020-GAUNet-bus-egress_time}
    \end{minipage}
    \begin{minipage}[b]{0.32\linewidth}
    \centering
    \includegraphics[width=1.0\hsize]{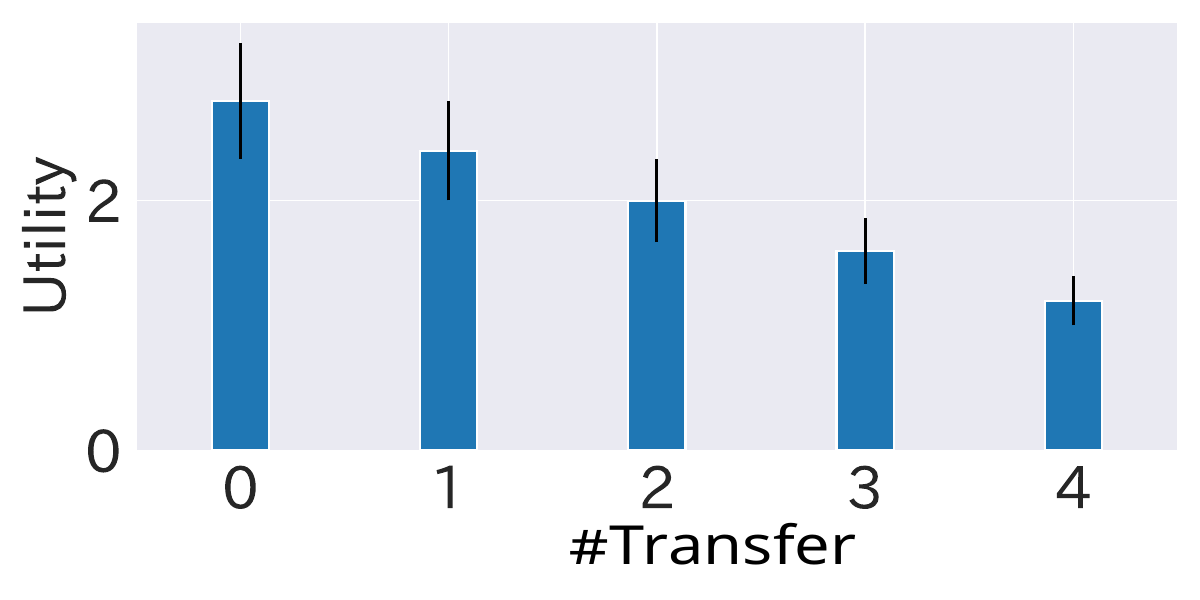}
    \subcaption{MNL-Linear: \#Transfers}\label{fig:toyosu2020-MNL-Linear-bus-num_transit}
    \end{minipage}    
    \begin{minipage}[b]{0.32\linewidth}
    \centering
    \includegraphics[width=1.0\hsize]{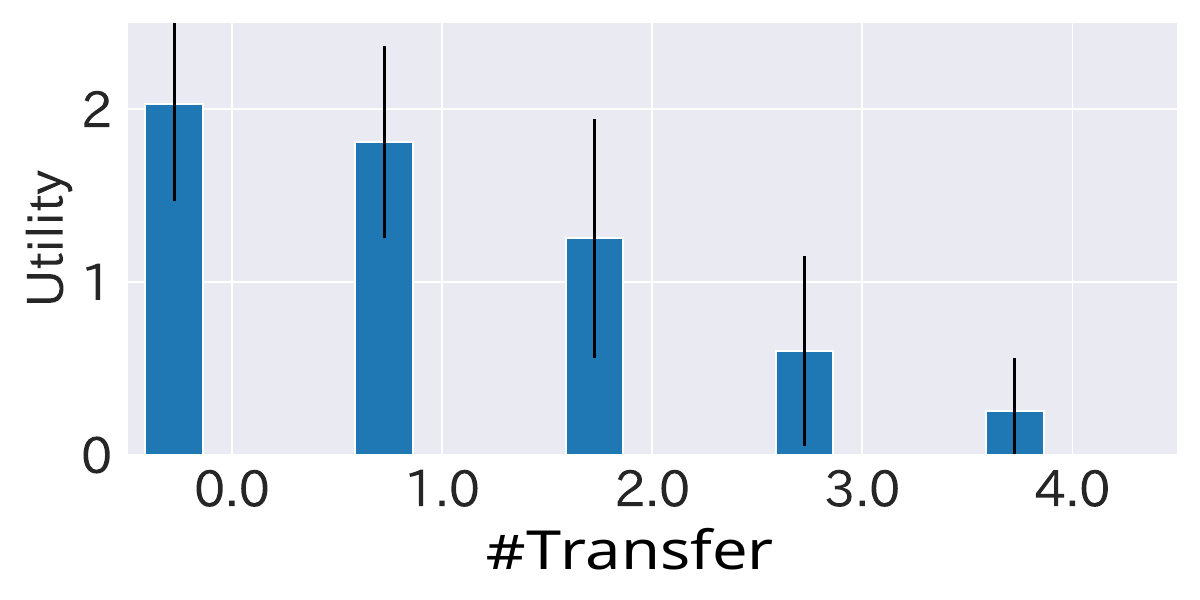}
    \subcaption{ASU-DNN: \#Transfers}\label{fig:toyosu2020-asudnn-bus-num_transit}
    \end{minipage}
\begin{minipage}[b]{0.32\linewidth}
    \centering
    \includegraphics[width=1.0\hsize]{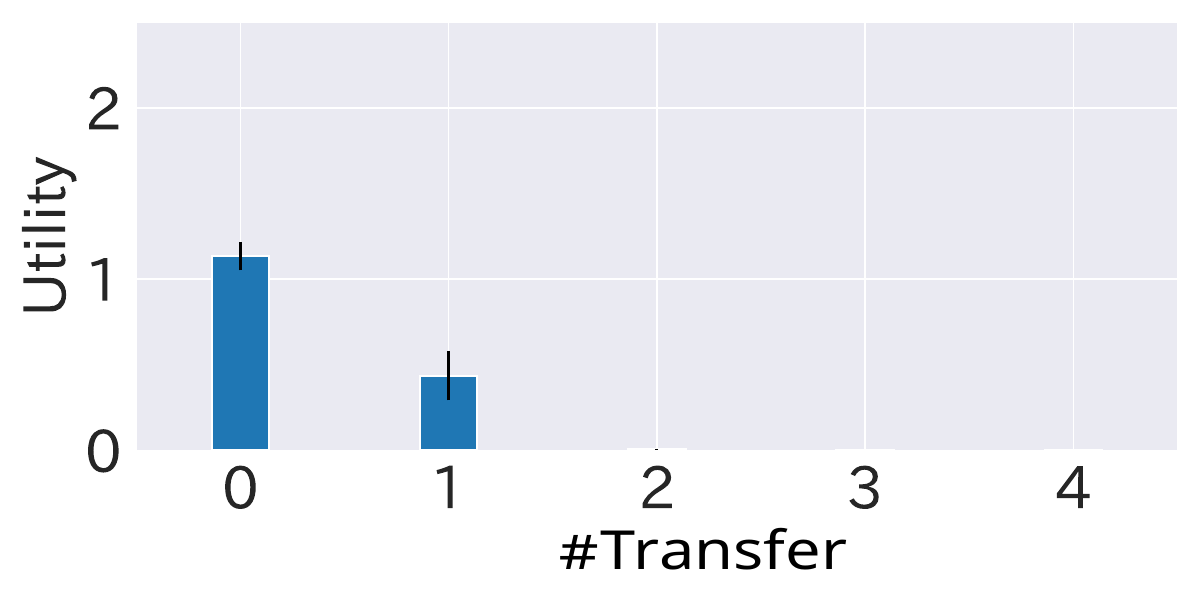}
    \subcaption{MNL-GAIUNet: \#Transfers}\label{fig:toyosu2020-GAUNet-bus-num_transit}
    \end{minipage}    
    \caption{Utility with respect to explanatory variables for buses based on MNL-Linear~(a, d, g, j), ASU-DNN~(b, e, h, k) and MNL-GAUNet~(c, f, i, l). The thin lines were
computed with the values included in the dataset as the values of the other explanatory variables. The bold lines and bars are the
medians.}
    \label{fig:toyosu2020-bus}
\end{figure}

We show the utility of each explanatory variable for the buses in MNL-Linear, ASU-DNN~(LeakyReLU), and MNL-GAIUNet~(Tanh) learned via the 2020 dataset in Fig.~\ref{fig:toyosu2020-bus} as typical examples. The thin lines are the utility of each variable calculated with the five learned models with each fold. In calculating the utility of each explanatory variable in ASU-DNN, the values of other explanatory variables included in each fold are used, and the results are shown with thin lines because the utility also depends on the values of the other variables. The bold lines are the medians. 

The results indicate that even if the log-likelihoods were comparable, the shape of the utility estimated by each model could be significantly different. First, when observing the bold lines of time-related variables (i.e., travel time, access time, and egress time), MNL-GAIUNet and ASU-DNN decrease to zero or close to zero around specific values or higher, and the utility becomes flat; MNL-Linear does not. This was due to the difference in the utility functions for each model. That is, MNL-GAIUNet and ASU-DNN can learn non-linear utility, but MNL-Linear can learn linear utility. Next, we also found that the shape of the utility functions estimated by the ASU-DNN for each variable often had high variance because it depended on the values of the other variables, although the shapes with medians were similar to those estimated by MNL-GAIUNet. The variance could make the result interpretation and the policy evaluation difficult. Finally, the small variance of the utility function estimated by MNL-GAIUNet between five folds indicates that MNL-GAIUNet was robust for the dataset. This robustness makes interpretation of the results and consistent policy evaluation easy.

\begin{figure}[t]
    \centering
        \begin{minipage}[b]{0.32\linewidth}
    \centering
    \includegraphics[width=1.0\hsize]{Toyosu2020-MNL-GAIUNet-s-bus-access_time.pdf}
    \subcaption{Bus Access Time}\label{fig:toyosu2020-GAIUNet-bus-accesstime}
    \end{minipage}
                \begin{minipage}[b]{0.32\linewidth}
    \centering
    \includegraphics[width=1.0\hsize]{Toyosu2020-MNL-GAIUNet-s-bus-egress_time.pdf}
    \subcaption{Bus Egress Time}\label{fig:toyosu2020-GAIUNet-bus-egresstime}
    \end{minipage}
        \begin{minipage}[b]{0.32\linewidth}
    \centering
    \includegraphics[width=1.0\hsize]{Toyosu2020-MNL-GAIUNet-s-bus-num_transit.pdf}
    \subcaption{Bus \#Transfer}\label{fig:toyosu2020-GAIUNet-bus-transfer}
    \end{minipage}
    \begin{minipage}[b]{0.32\linewidth}
    \centering
    \includegraphics[width=1.0\hsize]{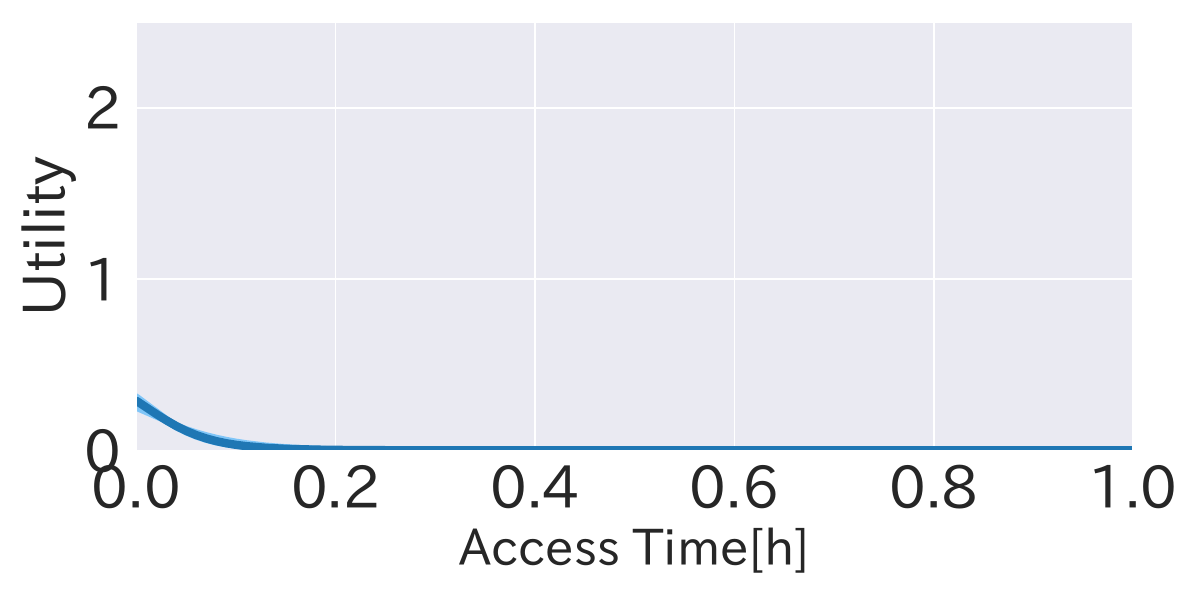}
    \subcaption{Train Access Time}\label{fig:toyosu2020-GAIUNet-train-accesstime}
    \end{minipage}
    \begin{minipage}[b]{0.32\linewidth}
    \centering
    \includegraphics[width=1.0\hsize]{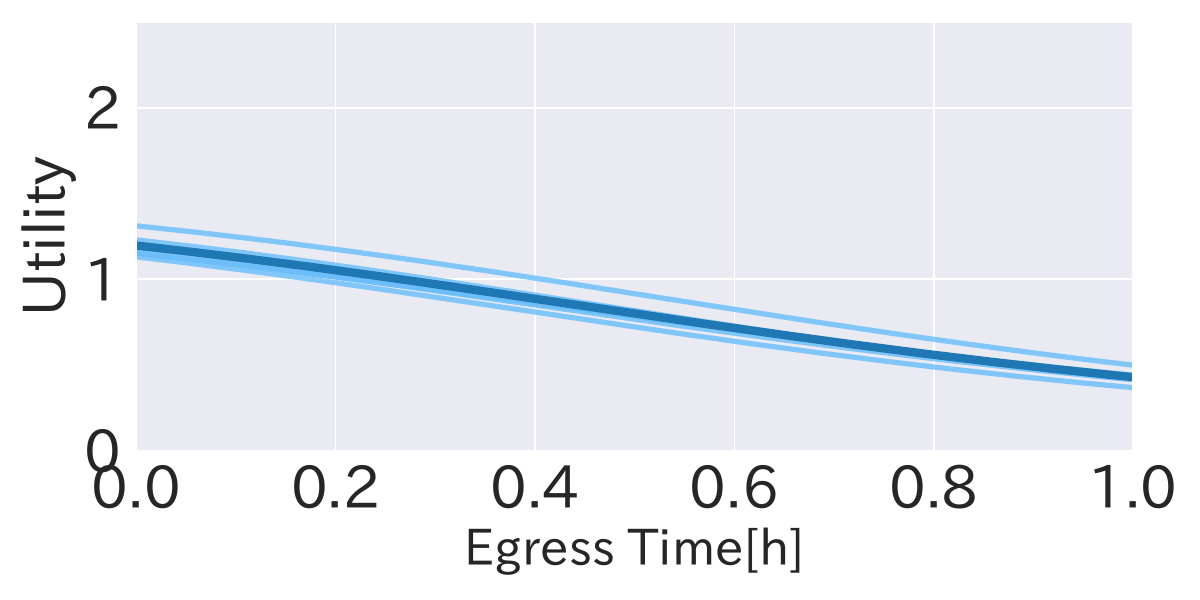}
    \subcaption{Train Egress Time}\label{fig:toyosu2020-GAIUNet-train-egresstime}
    \end{minipage}
    \begin{minipage}[b]{0.32\linewidth}
    \centering
    \includegraphics[width=1.0\hsize]{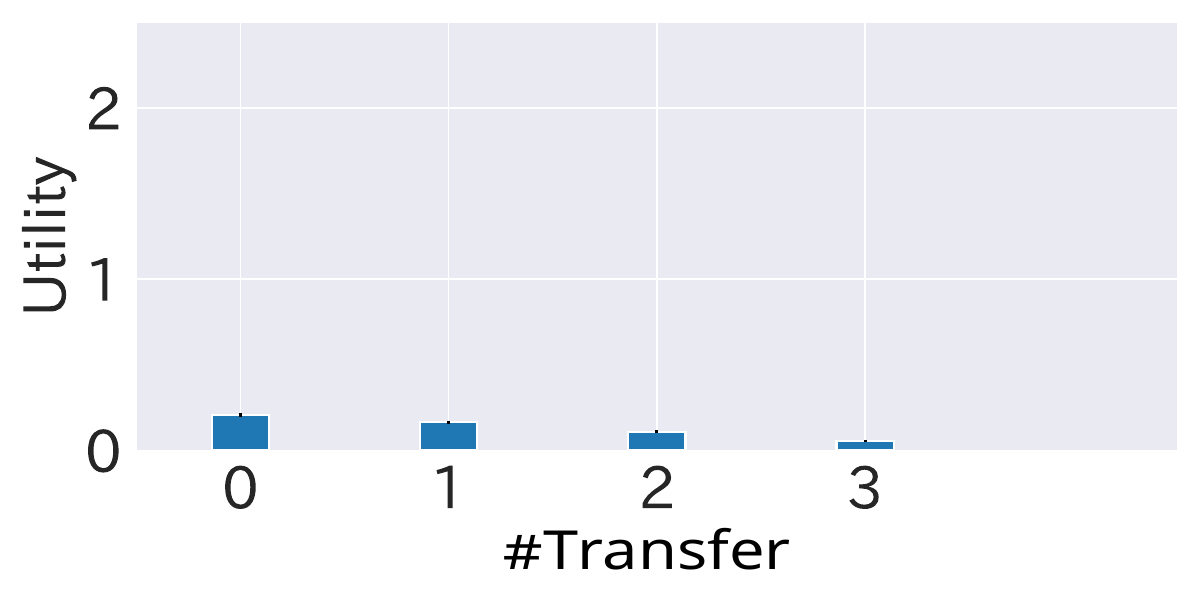}
    \subcaption{Train \#Transfer}\label{fig:toyosu2020-GAIUNet-train-transfer}
    \end{minipage}
    \caption{Utility for the explanatory variables (access time (a) (d), egress time (b) (e), and \#transfer (c) (f)) of buses (a) (b) (c) and trains (d) (e) (f) based on MNL-GAUNet in the 2020 dataset. The thin lines represent the utility in each fold. The bold lines are the medians. }
    \label{fig:toyosu2020-utility-GAIUNet-bus-train}
\end{figure}

Figure~\ref{fig:toyosu2020-utility-GAIUNet-bus-train}  shows the utilities with respect to three variables, access time, egress time, and the number of transfers, for the buses and trains estimated by MNL-GAIUNet in the 2020 dataset. We revealed that these utilities were different between transport modes, although shared weights for these variables have often been applied in previous analyses. The utility for the bus access time rapidly decayed from approximately 0.2 to 0.4 hours, while that of the trains decayed before 0.2, but the magnitude was smaller than that of the buses in Fig.~\ref{fig:toyosu2020-GAIUNet-bus-accesstime} and \ref{fig:toyosu2020-GAIUNet-train-accesstime}. The results indicated that the users were more concerned with the access time of the buses than the trains. We also found that the users could accept longer egress time on the trains than on the buses (Fig.~\ref{fig:toyosu2020-GAIUNet-bus-egresstime} and \ref{fig:toyosu2020-GAIUNet-bus-egresstime}). Moreover, we observed that the number of transfers that the users preferred for the buses could be less than two in Fig.~\ref{fig:toyosu2020-GAIUNet-bus-transfer}.  

\begin{figure}[t]
    \centering
    \begin{minipage}[b]{0.32\linewidth}
    \centering
    \includegraphics[width=1.0\hsize]{Toyosu2020-MNL-GAIUNet-s-bus-travel_time.pdf}
    \subcaption{Bus Travel Time}\label{fig:toyosu2020-GAIUNet-bus-traveltime}
    \end{minipage}
    \begin{minipage}[b]{0.32\linewidth}
    \centering
    \includegraphics[width=1.0\hsize]{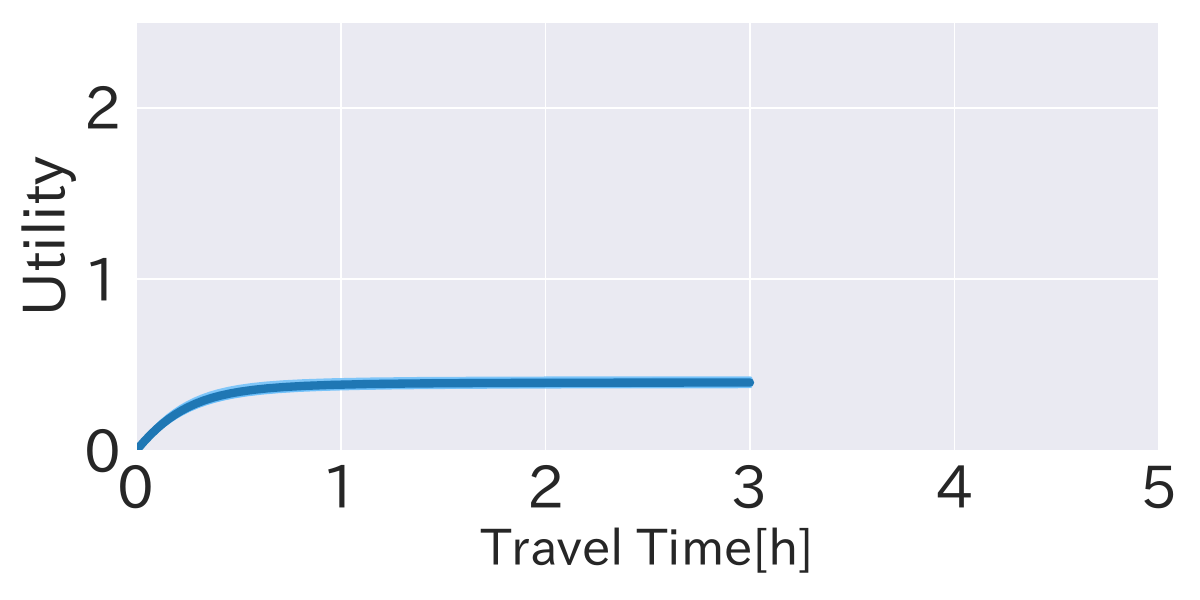}
    \subcaption{Train Travel Time}\label{fig:toyosu2020-GAIUNet-train-traveltime}
    \end{minipage}
    \begin{minipage}[b]{0.32\linewidth}
    \centering
    \includegraphics[width=1.0\hsize]{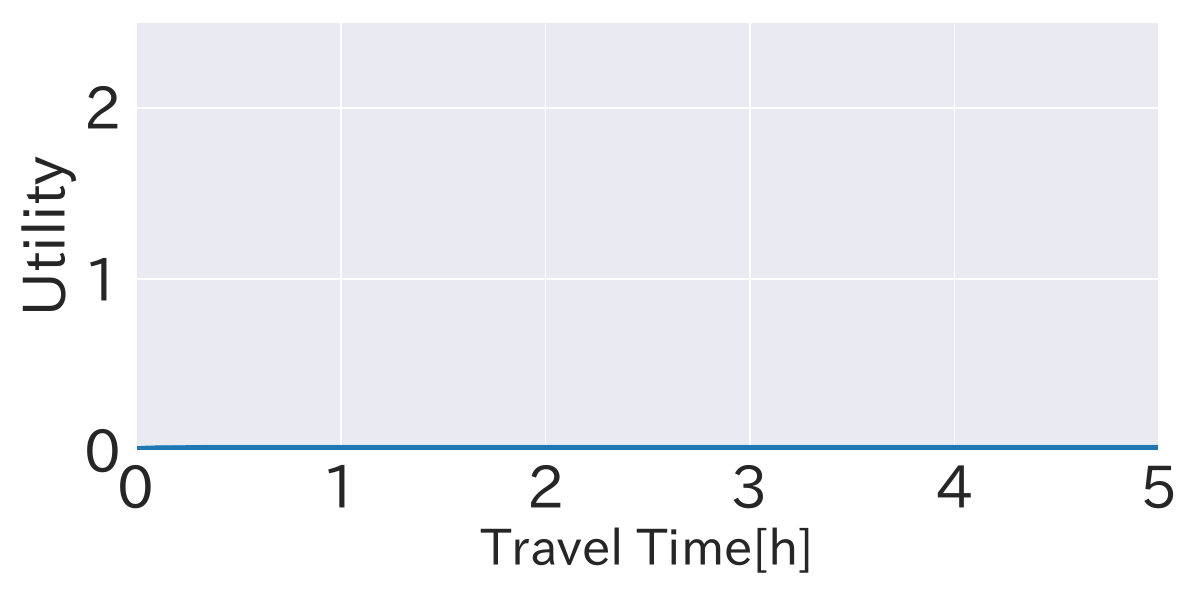}
    \subcaption{Car Travel Time}\label{fig:toyosu2020-GAIUNet-car-traveltime}
    \end{minipage}
        \begin{minipage}[b]{0.32\linewidth}
    \centering
    \includegraphics[width=1.0\hsize]{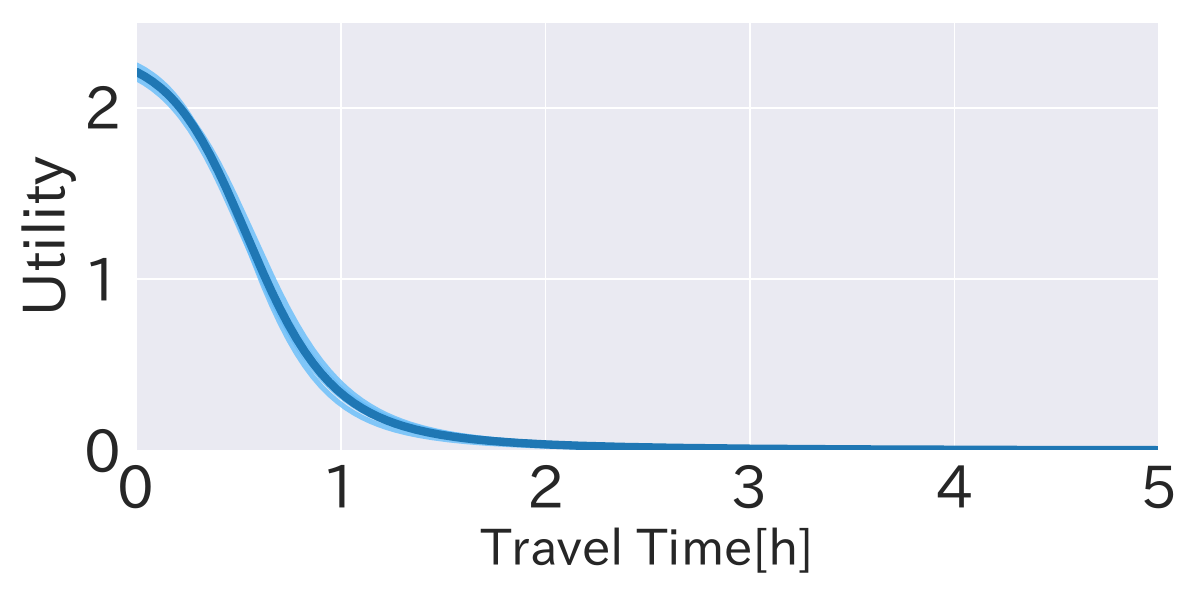}
    \subcaption{Bike Travel Time}\label{fig:toyosu2020-GAIUNet-bike-traveltime}
    \end{minipage}
    \begin{minipage}[b]{0.32\linewidth}
    \centering
    \includegraphics[width=1.0\hsize]{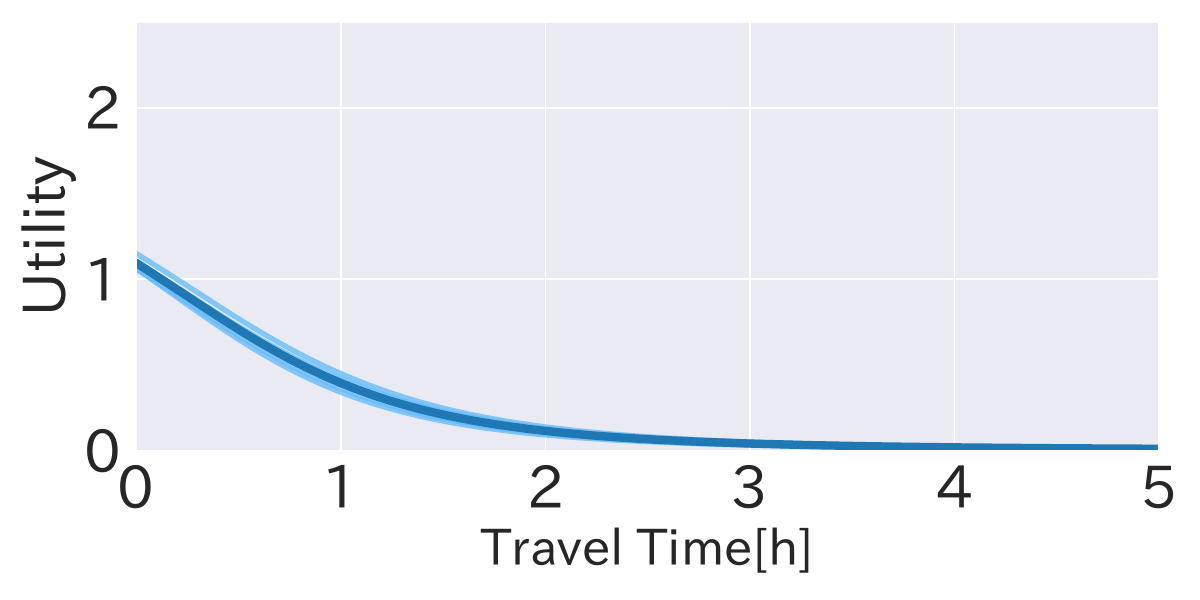}
    \subcaption{Walking Travel Time}\label{fig:toyosu2020-GAIUNet-walking-traveltime}
    \end{minipage}
    \caption{Utility for travel time of buses~(a), trains~(b), cars~(c), bikes~(d), and walking~(e) estimated with MNL-GAUNet in the 2020 dataset. The thin lines represent the utility in each fold. The bold lines are the medians. }
    \label{fig:toyosu2020-utility-GAIUNet-travel-time}
\end{figure}

Figure~\ref{fig:toyosu2020-utility-GAIUNet-travel-time} shows the utilities of travel time for all transportation modes, buses, trains, cars, bikes, and walking, in the 2020 dataset. The results revealed that users have little regard for the travel time of the buses and the cars, but accept the travel time of bicycles or walking within 0.2 h. Moreover, trains tended to be selected when the train travel time was longer than 0.5 h.  It would have been difficult to reveal these differences in time preference for these modes of transport without MNL-GAIUNet, which makes it easy to understand the results intuitively.

\section{Discussions and Conclusions}\label{sec:conclusion}

Discrete choice models have provided valuable insights for policymakers and businesses for many years. Interpretability of the analysis could often be a higher priority than prediction accuracy because their objective is to make better policy or business based on the analysis. We developed GAUNet, which is the utility function of DCM using a novel NN architecture based on GAMs. MNL-GAUNet, which is the multinomial logit model that uses GAUNet as the utility function, is the natural generalization of the standard MNL with linear utility functions. We also extended GAUNet to GAIUNet, adding interaction terms between two explanatory variables. 

We evaluated MNL-GAUNet with synthetic data of transport mode choice, including user unobservable constraints for the taxi travel cost and bus access time. We compared MNL-GAUNet to ASU-DNN and MNL-Linear in two aspects: prediction accuracy and policy evaluation. In terms of prediction accuracy, we showed that MNL-GAUNet outperformed MNL-Linear and was comparable to ASU-DNN. We also found that the utilities estimated by ASU-DNN had high variance with respect to the values of the explanatory variables other than the target variables while the estimation with MNL-GAIUNet was robust. In addition, the utility provided by MNL-GAIUNet compared with that of MNL-Linear provided intuitive interpretations of utility changes. In the policy evaluation, we found that while ASU-DNN and MNL-Linear decreased the prediction accuracy when additional taxi travel costs or bus access time was simulated as policies, MNL-GAUNet maintained the accuracy.  The results indicate that MNL-GAUNet can capture the unobservable constraints from data without prior knowledge. Moreover, the gap in the prediction accuracy between the test data and policy evaluation revealed that the accuracy with the test data could not justify the validity of the policy evaluation with the models. Therefore, we must validate the model from various perspectives. The intuitive interpretation provided by MNL-GAIUNet could be a reassuring tool for such validation.

We evaluated MNL-GAUNet and MNL-GAIUNet with transport mode prove-person data collected in Tokyo, Japan, from 2018 -- 2021. The results showed that our models outperformed MNL-Linear and were comparable to ASU-DNN in terms of prediction accuracy, as well as the evaluation of the above synthetic data. We also compared MNL-GAIUNet, ASU-DNN, and MNL-Linear by visualizing the learned utility with respect to each explanatory variable. We revealed that the utility of each variable learned with ASU-DNN had greater variance than that with MNL-GAIUNet and MNL-Linear because the utility for an explanatory variable estimated with ASU-DNN depends on the values of the other explanatory variables in the datasets, as well as the synthetic data. The variance could make understanding user preference and policy evaluation difficult. In addition, we compared the utilities of time-related variables and the number of transfers in each transport mode learned with MNL-GAIUNet. We confirmed that our approach could intuitively interpret the estimated user preference.  For instance, the results indicated that the users were more concerned with the access time of the buses than the trains. We believe that the users could accept longer egress time on the trains than the buses and that the number of transfers that the users preferred for the buses could be less than two.

In this study, we applied our models to transport mode choice. We will apply the model to broader choice behavior modeling tasks as the next steps.  In addition, our GAUNet and GAIUNet can naturally be applied to any variation of logit model families, such as (cross-)nested-logit models and latent-class models, by using our NN architecture as their utility functions. We will extend our model to these logit models in future work.

\section*{Acknowledgments}

This work was performed as a project of Intelligent Mobility Society Design, Social Cooperation Program~(Next Generation Artificial Intelligence Research Center, the University of Tokyo with Toyota Central R\&D Labs., Inc.)

\section*{Declaration of competing interest}
The authors declare that they have no known competing financial interests or personal relationships that could have appeared to influence the work reported in this paper.

\printcredits

\bibliographystyle{cas-model2-names}


\appendix

\bio{}
\endbio

\bio{}
\endbio

\end{document}